\documentclass{mindlab}

\usepackage{microtype}
\usepackage{amsmath}
\usepackage{amssymb}
\usepackage{array}
\usepackage{booktabs}
\usepackage{graphicx}
\usepackage{hyperref}
\usepackage{url}
\usepackage{xcolor}
\usepackage{enumitem}
\usepackage{float}
\usepackage{placeins}
\usepackage{listings}
\usepackage{tikz}
\usepackage{pgfplots}
\pgfplotsset{compat=1.18}
\usetikzlibrary{positioning,arrows.meta,calc,fit,backgrounds}
\usepackage{tabularx}

\newcolumntype{L}[1]{>{\raggedright\arraybackslash}p{#1}}
\newcolumntype{Y}{>{\raggedright\arraybackslash}X}

\DeclareUnicodeCharacter{1F624}{\faAngry}
\DeclareUnicodeCharacter{1F355}{\faPizzaSlice}
\DeclareUnicodeCharacter{1F62D}{\faSadCry}
\DeclareUnicodeCharacter{1F6CF}{\faBed}
\DeclareUnicodeCharacter{1F60C}{\faSmileBeam}
\DeclareUnicodeCharacter{FE0F}{}

\newcommand{\macaron}{Macaron-V1}
\newcommand{\venti}{Macaron-V1-Venti}
\newcommand{\tall}{Macaron-V1-Tall}
\newcommand{\codingventi}{Macaron-V1-Coding-Venti}
\newcommand{\emoji}[1]{\raisebox{-0.2ex}{\includegraphics[height=1.9ex]{figures/emoji/#1.png}}}
\newcommand{\chatbench}{Macaron ChatBench}
\newcommand{\livingbench}{Macaron LivingBench}
\newcommand{\mol}{Mixture of LoRA}
\newcommand{\molshort}{MoL}
\newcommand{\uifora}{UI4A}
\newcommand{\hcp}{Harness Context Protocol}
\newcommand{\longstraw}{\textsf{LongStraw}}
\newcommand{\mint}{\textsf{MinT}}
\newcommand{\mindforge}{\textsf{MindForge}}

\newenvironment{dialog}[2]{%
  \begin{tcolorbox}[arc=4pt,boxrule=0.4pt,colback=mindlabbg,colframe=mindlabline,
    left=6pt,right=6pt,top=5pt,bottom=5pt,before skip=6pt,after skip=6pt]
  \setlength{\parindent}{0pt}\footnotesize\raggedright
  \def\dialogleftname{#1}\def\dialogrightname{#2}%
}{\end{tcolorbox}}
\newcommand{\userturn}[1]{{\color{mindlabmuted}\textbf{User:}\ #1}\par\medskip}
\newcommand{\turnpair}[2]{%
  \begin{minipage}[t]{0.48\textwidth}%
    {\bfseries\color{mindlabmuted}\small \dialogleftname}\par\vspace{1pt}\hrule height 0.3pt\vspace{3pt}%
    {\color{mindlabink}#1}%
  \end{minipage}\hfill
  \begin{minipage}[t]{0.48\textwidth}%
    {\bfseries\color{mindlabblue}\small \dialogrightname}\par\vspace{1pt}\hrule height 0.3pt\vspace{3pt}%
    {\color{mindlabink}#2}%
  \end{minipage}\par\medskip
}

\title{Macaron-V1: Towards Open Continual Learning with Self-Improvement and Mixture-of-LoRA}
\author{Mind Lab}
\metadata[Models]{\url{https://huggingface.co/collections/mindlab-research/macaron-v1}}
\metadata[Correspondence]{\texttt{\textcolor{mindlabblue}{contact@mindlab.ltd}}}
\metadata[Date]{August 2026}

\abstract{
\macaron{} is an open agent-model family for experiential intelligence: learning
from experience in real environments and continuing to learn after deployment.
It is organized around two system goals. \emph{Adaptation} is pursued through
recursive improvement of versioned model-harness pairs, where
experience from one configuration is evaluated under an external contract and
used to construct its successor. \emph{Collaboration} is pursued via the
Mixture-of-LoRA (\molshort{}) architecture that freezes a base model, composes
specialist LoRA adapters, and selects one LoRA per user turn. The flagship \venti{} (748B) combines a 744B GLM-5.2 base with four
LoRAs for chat, agent, coding, and GenUI; the Qwen3.6-35B-based \tall{} (50B)
uses the same design for local deployment.

This report presents \macaron{} as a co-designed system spanning architecture,
algorithms, and infrastructure. The \molshort{} architecture supports continual 
learning through extensible LoRA specialists. The algorithm combines Model-Harness Co-design and 
recursive self-improvement loop, including the \uifora{} component-native GenUI harness, a stateful action substrate,
versioned \hcp{} contract, and the agentic RL framework \mindforge{}. The supporting infrastructure
includes the post-training platform \mint{}, the long-context RL
method \longstraw{}, and stability techniques for sparse MoE and DSA base models. We evaluate
\macaron{} on Personal Intelligence, GenUI, and general capability benchmarks
against frontier baselines. Our results validate the current system, while
compounding gains from continual learning and collective intelligence remain open questions.
}

\begin{document}
\maketitle

\vspace{-0.55cm}
\begin{figure}[H]
    \centering
    \includegraphics[width=0.78\textwidth]{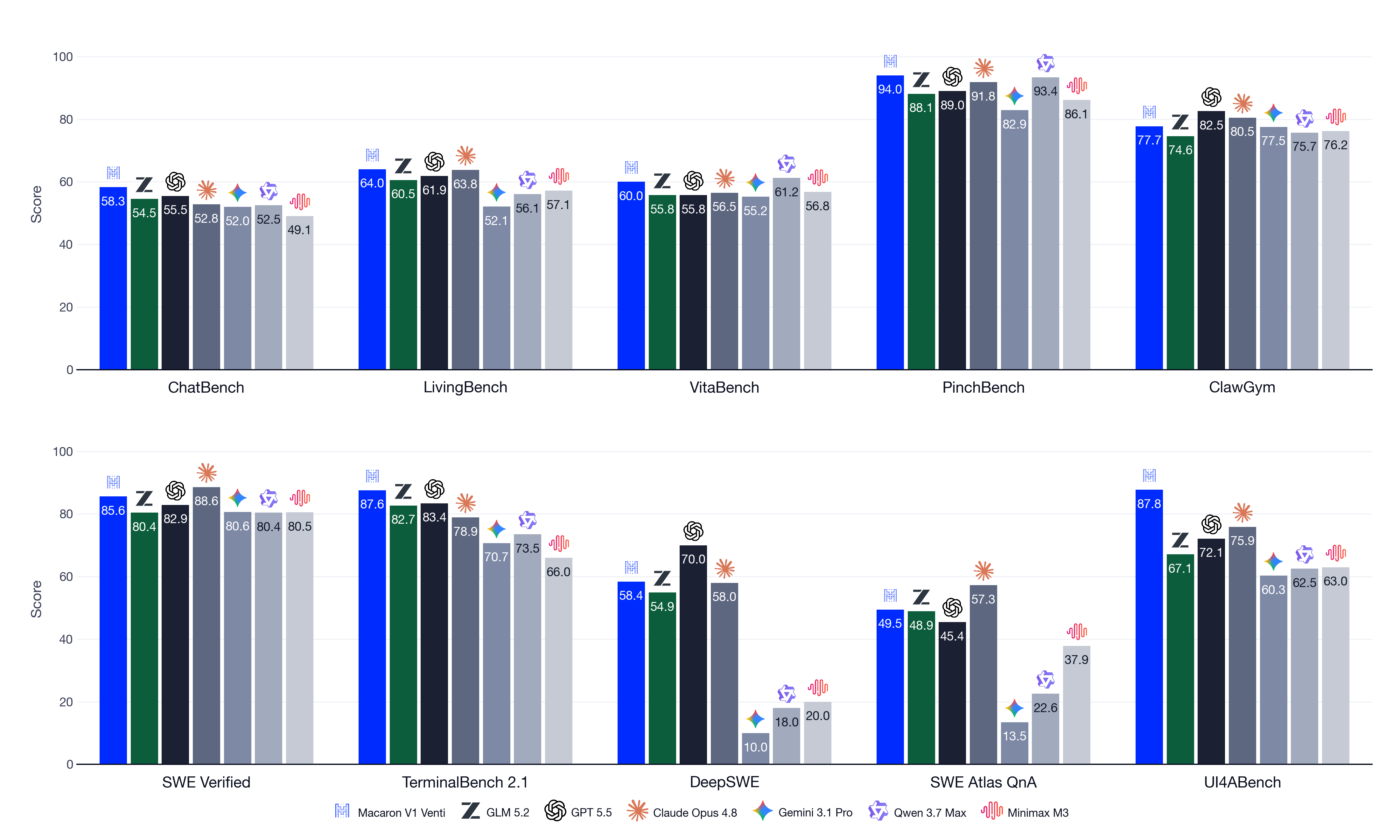}
    \captionsetup{justification=centering}
    \vspace{-0.2cm}
    \caption{\venti{} main results.}
    \label{fig:eval_bar}
\end{figure}

\setcounter{tocdepth}{2}
\tableofcontents
\newpage

\section{Introduction}
\label{sec:introduction}

As pre-trained models mature, an increasing share of advances in capabilities
such as agentic tool use and coding has come from post-training
\citep{openai_gpt55_2026,anthropic_opus48_2026,
kimi_k2_2025,glm5_2026}. The effectiveness of post-training, however, is
closely coupled to the environment in which a model is trained, evaluated, and
deployed
\citep{yao_second_half_2025,meta_harness_2026,
continual_harness_2026}. The data a model learns from, the tools it can use,
the users it serves, and the interactions it must support jointly shape the
behavior required of a deployed agent
\citep{when_cl_requires_learning_2026,continual_harness_2026,
meta_harness_2026}. These conditions vary across deployments and continue to
evolve over time, as new knowledge becomes available, new domains emerge, and
interaction state accumulates across episodes
\citep{tic_lm_2025,when_cl_requires_learning_2026,exg_2026,delta_mem_2026}.

A common practice is centralized post-training, in which a model is optimized
over a bounded set of tasks and environments, aligned with the current harness,
and consolidated into a checkpoint for release. Training on a broad task
mixture can improve aggregate performance across tasks, but heterogeneous
objectives can also create cross-task interference when they compete through
shared parameters
\citep{badit_2026}. More fundamentally, the resulting system remains tied to
the task and environment distribution available during training, while new
knowledge, tools, user needs, and interactions continue to arrive after release
\citep{tic_lm_2025,when_cl_requires_learning_2026,
continual_harness_2026,exg_2026}. We see this as approximating a static optimum
rather than constituting genuine intelligence. What we consider genuine
intelligence is \emph{experiential intelligence}: the ability to learn from
experience accumulated in a real environment and to keep learning after
deployment
\citep{silver_sutton_experience_2025,yao_second_half_2025}.

Mind Lab is built for experiential intelligence. We use \emph{Personal
Intelligence} for the user-facing competence required when an agent works over
time for a particular person: using tools reliably, remembering constraints
across sessions, deciding when to ask a clarifying question and when to act,
and staying honest with the person on the other end of the conversation. These
behaviors describe the product target; experiential intelligence adds a
temporal requirement that deployment experience feeds a later model--harness
revision. Current agent-model training largely treats the target behaviors as
capabilities to develop before release and then hold fixed: the model is tuned
against a snapshot of the environment and shipped. Once deployed, the
resulting system has limited mechanisms for turning execution experience into
systematic improvement over time~\citep{exg_2026,continual_harness_2026}.
\macaron{} is our attempt to build that revision loop.

\paragraph{Adaptation and collaboration.}
Experiential intelligence, as we operationalize it, has two complementary
dimensions, and \macaron{} is built around both.

\emph{Adaptation} is pursued through recursive improvement over explicitly
versioned model--harness pairs: each version names a fixed model snapshot and
runtime configuration.
Recent approaches differ in where and how they make experience persistent.
Parametric continual-learning methods update model weights across changing
tasks or distributions
\citep{continual_learning_llm_survey_2026,tic_lm_2025,
when_cl_requires_learning_2026}. Memory- and state-based methods carry
trajectories, retrieved experience, or compact online states forward without
immediately modifying model parameters
\citep{jitrl_2026,delta_mem_2026,livemem_2026,
metis_2026,exg_2026}. Context- and harness-based methods instead update the
instructions, tools, skills, memory procedures, orchestration, or execution
structure surrounding the model
\citep{ace_2026,meta_harness_2026,continual_harness_2026,
recursive_harness_2026}.

\macaron{} connects harness adaptation with modular parametric adaptation: it
builds harder tasks from seeds, audits trajectories in the production harness,
searches the context configurations that shape behavior, and selects
trajectories for subsequent updates to specialist LoRA adapters. Adaptation is
therefore a property we seek across successive model--harness revisions rather
than in any single checkpoint. We use \emph{recursive improvement} in a bounded
and auditable sense: experience generated by one versioned model--harness
configuration is evaluated under an external contract and used to construct a
successor revision
\citep{rsi_survey_2026}. Repeating this process across successive revisions,
while measuring retention, transfer, and cumulative improvement, is the
longer-term continual-learning objective
\citep{continual_learning_llm_survey_2026,
when_cl_requires_learning_2026}.

\emph{Collaboration} is pursued through composition. Joint post-training over
heterogeneous tasks within a shared parameter space can create cross-task
interference, motivating a representation in which specialists remain
separable
\citep{badit_2026}. Existing approaches compose specialists at different
granularities: mixture-of-adapter systems select or combine LoRA specialists at
the task, sequence, or token level
\citep{lorauter_2026,molora_2026}, while multi-agent systems coordinate
complete model instances through routing, communication, discussion, or
response aggregation
\citep{multiagent_collaboration_survey_2025,rethinking_moa_2025}.
Mixture-of-LoRA (\molshort{}) occupies a different operating point: it keeps a
large base model frozen, layers specialist LoRA adapters on top
\citep{lora2022,lora_without_regret_2025}, and selects one specialist per user
turn through a shared runtime
\citep{mol_harness2026}.

\molshort{} demonstrates \emph{modular collaboration}. We reserve
\emph{collective intelligence} for the stronger setting in which independently
trained specialists contribute complementary capabilities, enabling the
composed system either to outperform its strongest constituent or to solve
tasks that no constituent can solve alone under a matched evaluation budget
\citep{rethinking_moa_2025,multiagentbench_2025,superminds_2026}.
Extending \molshort{} to include specialists trained by different teams or for
different users is the broader collective-intelligence objective.

Adaptation and collaboration are complementary dimensions of an evolving agent
system. Collaboration determines how distinct capabilities are represented,
selected, and composed, while adaptation determines how those capabilities and
the surrounding harness change in response to experience. Extending this
process across multiple generations, so that specialists, routing policies,
and harness configurations improve together while retaining prior capabilities,
remains a longer-term research direction
\citep{evolverouter_2026,evochamber_2026}.

\begin{figure}[t]
    \centering
    \includegraphics[width=0.95\linewidth]{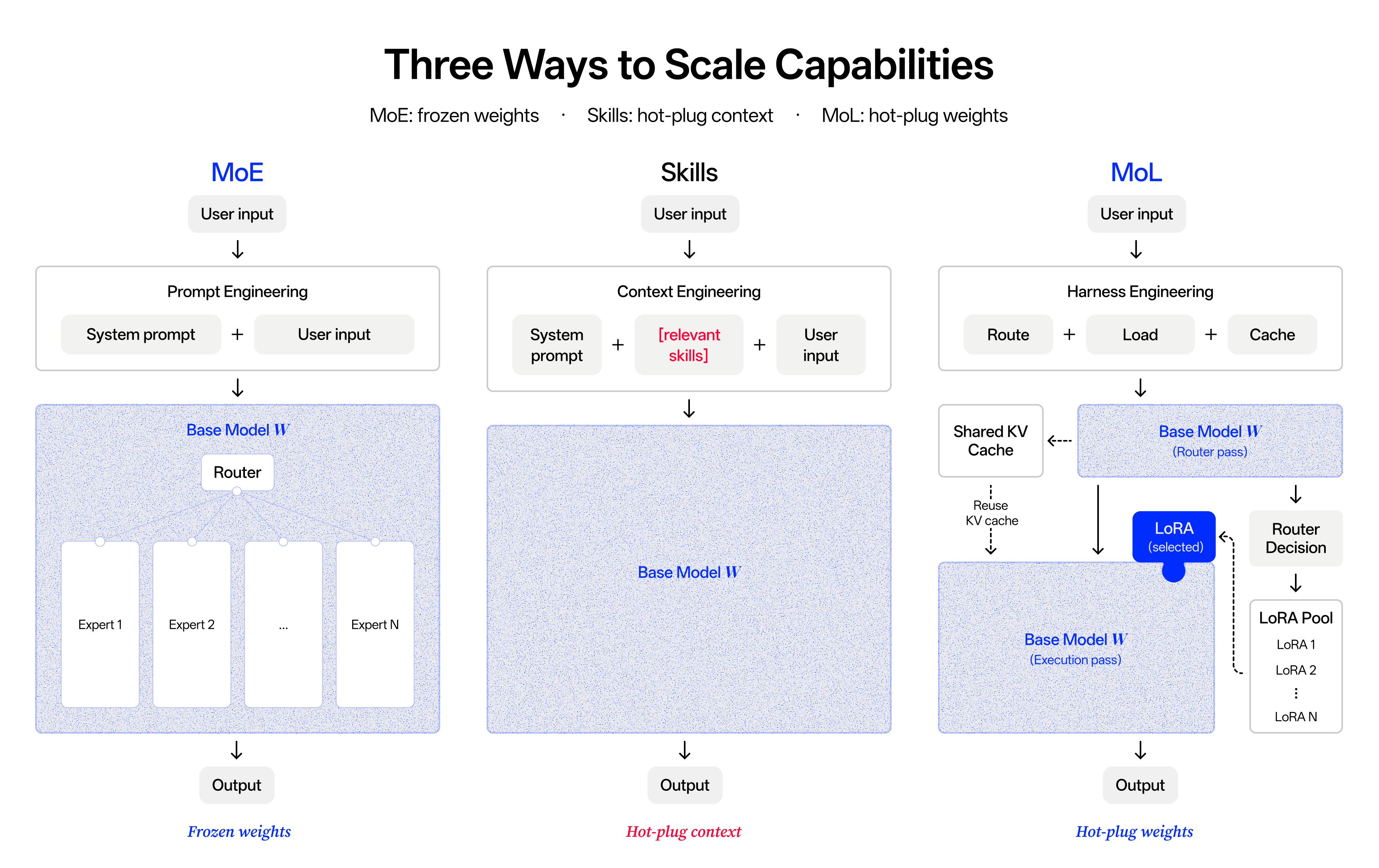}
    \caption{Three ways to scale model capability. Mixture-of-Experts (MoE) scales the base itself; skills scale scaffolding around a fixed model; \molshort{} (\mol) keeps the base frozen and composes specialist LoRA adapters through a Proxy-mediated routing loop. This separation is an implementation substrate for our continual-learning and collective-intelligence goals; the diagram is not a controlled comparison of the three alternatives.}
    \label{fig:mol_three_ways}
\end{figure}

\paragraph{Model family.}
\macaron{} is a family of open models rather than a single checkpoint.
\begin{itemize}[leftmargin=*]
  \item \venti{}: the flagship release, conventionally labeled 748B,
  consisting of a 744B GLM-5.2 base
  \citep{glm52_2026,glm5_2026} and four LoRA specialists. The label is
  release-facing; Section~\ref{sec:mol_specialists} gives the public
  stored-tensor count. It is
  post-trained with \mint{}
  \citep{lu2026announcing} and served through the \molshort{} harness
  \citep{mol_harness2026}.

  \item \tall{}: a 50B model post-trained from Qwen3.6
  \citep{qwen36_2026,qwen3_2025}, targeted at local deployment and
  lower-latency serving. It ships the same four-adapter \molshort{} design as
  \venti{}, with adapter placement and ranks chosen for Qwen3.6's MoE expert
  layout. Its public adapter configurations use a different rank and expert
  target set from \venti{}; the resulting aggregate is therefore reported as a
  rounded release footprint rather than inferred from base size alone.

  \item \codingventi{}: a coding-specialized build of \venti{} released
  by merging the coding LoRA directly into the base rather than serving it
  as a \molshort{} adapter.
\end{itemize}

The two \molshort{}-served variants share the same architectural pattern: a
frozen base, a small set of LoRA specialists, a Proxy-mediated route hop whose
label is emitted by L0, and a harness that stays close to production during both
training and serving. \codingventi{} is the single-specialist exception.

\paragraph{Contributions.}
This report describes \macaron{} as a co-designed system. Concretely, we
contribute:

\begin{itemize}[leftmargin=*]
  \item \textbf{\mol{} architecture.}
  A Proxy-mediated, per-turn composition scheme on top of a frozen base. The
  separation makes it possible to add LoRA specialists without retraining the
  base and to register specialists from different teams or users on one
  runtime. The shipped \venti{} model instantiates this architecture with four
  specialists; we report its routing functionality and serving cost, while
  continual improvement across generations and complementary gains from
  independently trained specialists remain evaluation targets described in
  Section~\ref{sec:disc_roadmap}. We release the serving harness at
  \texttt{MindLab-Research/Mixture-of-LoRA-Harness}
  \citep{mol_harness2026}.

  \item \textbf{Model--Harness Co-design and RSI.}
  We treat the harness as a first-class optimization target:
  \uifora{}
  \citep{ui4a2026} is a component-native harness for Generative UI,
  extending our prior schema-native work
  \citep{macaron_a2ui2026,a2ui_v08}; our REPL-based agent harness
  \citep{wu2026replharnesses} treats \emph{executable composition} and
  \emph{validated reuse} as the interface between the model and its tools; and
  the \hcp{} makes runtime selection, tools, skills, prompts, hooks, sessions,
  and workspace resources portable and auditable. \mindforge{}, our agentic RL
  framework, runs a three-stage loop of task discovery, trajectory expansion,
  and linked model/configuration updates against that harness. The experiment
  reported here isolates only the Expansion stage: the model remains frozen and
  no optimizer step is taken. In this fixed-model harness-search study, adaptive
  search reaches cumulative coverage of $122/122$ selected base-failure tasks,
  while the stronger of two full-set single-configuration sweeps reaches
  $11/122$ (Section~\ref{sec:rsi_coverage}). Because task targeting and error
  rates differ across phases, this result measures configuration-search coverage,
  not improvement produced by the full RSI training cycle.
  R3 rollout routing replay
  \citep{chiang2026routerreplay,r3_moe_router2025}, DSA integration fixes
  \citep{stevenchiang2026supportglm5inmint}, and IcePop-style token masking
  \citep{ling_every_step2025} address rollout--learner mismatch on the sparse
  base.

  \item \textbf{Infrastructure.}
  We provide the \mint{} post-training platform
  \citep{lu2026announcing}, including adapter-revision handoff, a measured
  million-entry addressable catalog, and trillion-parameter-class
  model-parallel paths, together with \longstraw{}
  \citep{zhou2026longstrawlongcontextrl2m}, an architecture-aware
  response-only execution stack with reported multi-million-token operating
  points under fixed GPU inventories. These measurements come from the
  companion systems reports rather than matched \macaron{} training runs.

  \item \textbf{Personal Intelligence benchmarks.}
  We introduce \chatbench{}, which evaluates context-conditioned agent
  conversation, and \livingbench{}, which evaluates stateful simulated
  personal-life assistance under evolving conditions. Both evaluate
  interaction trajectories rather than isolated answers.

  \item \textbf{Evaluation.}
  We evaluate \macaron{} across Personal Intelligence
  (\chatbench{}, \livingbench{}), agent, coding, and GenUI tracks
  against six comparison models. We distinguish scores reproduced in our
  harness from values taken from public reports, and treat the internal
  Personal Intelligence results as in-distribution evidence rather than as a
  general-capability comparison (Section~\ref{sec:results}).
\end{itemize}

\paragraph{Report structure.}
Section~\ref{sec:mol} describes collaboration through the \molshort{}
architecture, including its mechanisms for specialist composition and its
paths toward continual learning and collective intelligence.
Section~\ref{sec:algorithm} describes adaptation through Model--Harness
Co-design (\uifora{}, the REPL harness, and \hcp{}) and the recursive
self-improvement loop. Section~\ref{sec:infrastructure} covers the \mint{}
lifecycle, the \longstraw{} execution path, and sparse-base mismatch controls.
Section~\ref{sec:benchmarks} introduces the benchmarks we evaluate on,
Section~\ref{sec:results} reports the results, and
Section~\ref{sec:discussion} discusses limitations and the roadmap.

\section{Mixture of LoRA}
\label{sec:mol}

\mol{} (\molshort) is the architectural spine of \macaron{}. Instead of training a single monolithic model to cover every capability, we keep a large base model frozen and layer a small set of specialist LoRA adapters~\citep{lora2022,lora_without_regret_2025,zhou2025bslora} on top. Adapter composition is orchestrated by a serving layer, the \emph{MoL Proxy}, that treats adapter selection as a first-class, observable action per user turn rather than as a hidden mixer~\citep{mol_harness2026}. This section describes the design principle, the specialists shipped in \venti, the routing loop and its measured cost, the per-adapter conversation construction that makes routing cheap, the two-level KV-reuse design and its quality trade-off, and the architectural affordances intended to support separately versioned improvement and cross-owner composition. The reference implementation is open-sourced at \texttt{MindLab-Research/Mixture-of-LoRA-Harness}~\citep{mol_harness2026}.

\subsection{Design Principle}
\label{sec:mol_principle}

Agent workloads combine tasks with very different chain-of-thought patterns: chat, agentic tool use, coding, and GenUI each ask the model to think in a different shape. Joint post-training can create cross-task interference, motivating an architecture that makes specialization explicit~\citep{macaron_v1_preview2026}. The present release does not include the budget-matched single-LoRA comparison needed to quantify that interference, so we treat it as a design motivation rather than an empirical finding of this report.

\molshort{} exposes the trade-off explicitly through a single rule~\citep{macaron_v1_preview2026} (Figure~\ref{fig:mol_three_ways}):

\begin{quote}
Cluster tasks that share skills and thinking patterns into one LoRA, and keep tasks whose skills diverge sharply in separate LoRAs.
\end{quote}

Two properties fall out of this rule that support the longer-term continual-learning and collective-intelligence goals:
\begin{itemize}[leftmargin=*]
  \item \textbf{The base is frozen.} New capabilities are added by training and registering another adapter rather than by re-training the base. Base weight does not get overwritten by later specialization.
  \item \textbf{Adapters are portable.} Because the base is shared, a specialist trained by one team, or personalized for one user, can be composed with specialists from another team or user on the same runtime~\citep{macaron_v1_blog2026}.
\end{itemize}
Section~\ref{sec:mol_continual} returns to these properties.

\subsection{Specialists in \venti}
\label{sec:mol_specialists}

\venti{} (748B) ships four release-labeled 1B LoRA specialists on top of a frozen 744B GLM-5.2~\citep{glm52_2026,glm5_2026} base. The specialists absorb work that was split across five adapters in the \macaron{}-Preview release; most notably, its OpenClaw specialist has been rolled into the Agent adapter L1~\citep{macaron_v1_preview2026,macaron_v1_blog2026}. The public Venti adapter configurations use rank $16$, LoRA alpha $32$, and the same attention/MLP target modules for L0--L3. Summing the tensor shapes in the released adapter headers gives 7,688,042,496 stored values per adapter; this is a logical stored-tensor count, not an active-per-token count or a device-memory measurement. We therefore keep the 748B figure as the release-facing label and use the stored count explicitly when discussing residency below. A smaller variant, \tall, ships the same four-adapter design on a Qwen3.6-35B-A3B~\citep{qwen36_2026,qwen3_2025} base for local deployment. Its public configurations use rank $64$, LoRA alpha $128$, and expert-specific target parameters; each adapter contains 3,775,651,840 stored values (L2 is stored in F32), giving approximately 50.1B when added to the nominal 35B base. The two adapter budgets are not directly comparable because the bases expose different target-module and expert structures.

\begin{itemize}[leftmargin=*]
  \item \textbf{L0, Chat.} Conversational backbone, instruction following, and model identity. L0 also serves as the routing entry point (Section~\ref{sec:mol_router}).
  \item \textbf{L1, Agent.} Long-horizon, heavy tool-use tasks including personal-agent workflows and service integrations. Absorbs the Preview's OpenClaw adapter.
  \item \textbf{L2, Coding.} Code generation, SWE-style tasks~\citep{jimenez2024swebench}, and terminal use.
  \item \textbf{L3, GenUI.} \uifora{}~\citep{ui4a2026} rendering and UI-driven action; specialized on TSX (React and SolidJS) with the same output serving other targets through framework-specific renderers.
\end{itemize}
The four adapters are resident in the engine under exactly the names \texttt{L0}--\texttt{L3}; \texttt{L0} is also the routing entry.

\subsection{Routing Loop}
\label{sec:mol_router}

\molshort{} does not train a separate router model. Adapter selection is decided per user turn by the entry adapter L0's own reasoning and executed by the MoL Proxy. A normal user-message request runs a three-stage lifecycle:

\begin{enumerate}[leftmargin=*]
  \item \textbf{Route.} L0 classifies the incoming request into exactly one canonical adapter label from \texttt{L0}--\texttt{L3} under a tight decode budget (24 tokens). The router prompt frames the request as quoted, untrusted text, checks the wrapper families in priority order (generative UI, then code/terminal, then personal-agent/living), and returns a single canonical label under a constrained-decoding grammar that restricts the output to the four legal labels. If L0 selects itself the request stays on L0; otherwise the Proxy switches to the target adapter.
  \item \textbf{Answer.} The chosen specialist responds from its own conversation view (Section~\ref{sec:mol_ownview}), seeded with cross-adapter summaries from prior turns.
  \item \textbf{Summary.} The specialist emits a short summary of what it just did, capped at 192 output tokens. The Proxy stores this summary server-side and never returns it to the client; it becomes shared context that any adapter can inherit on subsequent turns.
\end{enumerate}

\paragraph{Model routing.}
The route is decided entirely by L0's own reasoning: the entry adapter reads the request and emits a canonical adapter label under the constrained-decoding grammar, and that label is taken directly as the route. There is no separate router model and no keyword-based rule library overriding the decision; L0 is the router. This makes the routing decision a property of the chat specialist's understanding of the request, rather than a hand-tuned classifier, and lets routing accuracy improve with the base and the L0 adapter rather than with a separate artifact.
The benchmark families named in the checked-in prompt are routing examples rather than an exhaustive allowlist; other requests are assigned through the prompt's semantic boundaries. We do not report a benchmark-specific routing audit for PinchBench or ClawGym.

\paragraph{Short-circuits.}
\emph{Tool-result stickiness:} when the answer stage ends in a tool call, the following tool-result turn is locked to the same adapter, skipping routing and summary entirely. \emph{Transactional rollback:} the Proxy checkpoints conversation state before each turn and restores it on engine failure or client disconnect, so an undelivered turn never enters conversation history.

\paragraph{Runtime architecture.}
The Proxy is engine-agnostic and may sit directly in front of a single engine or behind a gateway for multi-worker dispatch. Its only OpenAI-compatible model name is \texttt{Macaron-V1-Venti}; internal base and adapter names remain private. It offers three API surfaces (stateless Chat Completions, stateful Responses, and Anthropic Messages), all driven by the same three-stage core. The design builds on multi-tenant LoRA serving research~\citep{punica2024,slora2023,zhou2025dynamic} implemented natively in vLLM~\citep{vllm2023} and SGLang~\citep{sglang2024}.

\paragraph{Routing cost.}
Routing is not free: every user turn adds a dedicated routing hop and a summary hop on top of the specialist's own generation. We measure this cost directly on the \venti{} and \tall{} Proxies with per-hop timing instrumentation, over multi-turn mixed-domain conversations (48 requests, temperature~0). Table~\ref{tab:mol_cost_breakdown} reports the per-hop averages for both profiles. On \venti, the routing decision (L0 constrained-decoding a canonical label under a 24-token budget) costs $0.54$\,s; the summary hop, capped at 192 output tokens, costs $0.97$\,s. Together they are $1.51$\,s, about $32\%$ of the three-hop total; the rest is the specialist's answer generation. On \tall{} (Qwen3.6-35B-A3B), the same hops cost $0.20$\,s and $0.32$\,s ($30\%$ of the total). The overhead share is stable across base sizes. The own-view reconstruction described next is folded into the answer hop and is not a separate cost.

\begin{table}[t]
\centering
\caption{Per-hop latency of the route--answer--summary loop on \venti{} (GLM-5.2) and \tall{} (Qwen3.6-35B-A3B), 48 multi-turn requests each, temperature~0. Share is the hop's share of the three-hop total (route + answer + summary).}
\label{tab:mol_cost_breakdown}
\begin{tabular}{lrrrr}
\toprule
Hop & \multicolumn{2}{c}{\venti} & \multicolumn{2}{c}{\tall} \\
 & Avg (s) & Share & Avg (s) & Share \\
\midrule
Route (L0 constrained-decode, 24 tok)   & $0.54$ & $12\%$ & $0.20$ & $11\%$ \\
Answer (specialist generation)          & $3.17$ & $68\%$ & $1.24$ & $70\%$ \\
Summary (192-tok cap)                    & $0.97$ & $20\%$ & $0.32$ & $19\%$ \\
\midrule
Total (route + answer + summary)        & $4.68$ & $100\%$ & $1.76$ & $100\%$ \\
\bottomrule
\end{tabular}
\end{table}

\paragraph{Routing accuracy.}
Routing accuracy is measured on a 6{,}448-sample trace drawn from LoRA training data: each sample is sent to L0, the model's output is parsed as a canonical label, and the label is compared to the dataset gold label. The trace is not an independent held-out split, so this measurement is an implementation diagnostic and does not estimate routing generalization. This is the model's own routing decision under the production constrained-decoding grammar, not a rule-library match. The deployed router prompt reaches $6391/6448 = 99.12\%$ accuracy with $100\%$ canonical-label compliance (every output is a legal \texttt{L0}--\texttt{L3} label) and zero request or parse errors. Per-class accuracy ranges from $97.1\%$ (L1) to $100\%$ (L2); the residual errors concentrate at the L0/L1 boundary (general chat versus personal-agent), which are the two semantically closest classes (Table~\ref{tab:mol_route_acc}). An integration run through the full Proxy stack, using an earlier GLM-5.2 prompt revision on a balanced 480-sample subset (120 per class), obtains $468/480 = 97.5\%$ with all requests and adapter selections valid. The same 6{,}448-sample trace evaluated on \tall{} (Qwen3.6-35B-A3B base) reaches $99.04\%$; every sample ID and input hash matches across the two runs, so routing accuracy is stable across base sizes. Multi-hop behavior is checked separately on the GLM-5.2 Proxy: an eight-turn L1/L2 alternating trace repeated in three independent conversations is correct on $24/24$ turns, including $21/21$ within-conversation domain switches and $18/18$ specialist re-entry turns.

\begin{table}[t]
\centering
\caption{Routing confusion matrices for \venti{} (left) and \tall{} (right), each evaluated on the same 6{,}448-sample trace; trace identity is verified by matching every sample ID and input hash. L0, L1, L2, and L3 denote Chat, Agent, Coding, and GenUI, respectively. Rows are the gold label; columns are the model's canonical-label output. Both runs reach $\sim$99\% accuracy with $100\%$ canonical-label compliance and zero request/parse errors.}
\label{tab:mol_route_acc}
\begin{tabular}{l|cccc|r|cccc|r}
\toprule
 & \multicolumn{5}{c|}{\venti} & \multicolumn{5}{c}{\tall} \\
Gold $\backslash$ Pred & L0 & L1 & L2 & L3 & Acc.\ & L0 & L1 & L2 & L3 & Acc.\ \\
\midrule
L0 (1000) & \textbf{977} & 19 & 4 & 0 & 97.70\% & \textbf{977} & 9 & 14 & 0 & 97.70\% \\
L1 (796)  & 23 & \textbf{773} & 0 & 0 & 97.11\% & 38 & \textbf{757} & 1 & 0 & 95.10\% \\
L2 (652)  & 0 & 0 & \textbf{652} & 0 & 100.0\% & 0 & 0 & \textbf{652} & 0 & 100.0\% \\
L3 (4000) & 11 & 0 & 0 & \textbf{3989} & 99.73\% & 0 & 0 & 0 & \textbf{4000} & 100.0\% \\
\midrule
Total     & \multicolumn{4}{c|}{6391 / 6448} & 99.12\% & \multicolumn{4}{c}{6386 / 6448} & 99.04\% \\
\bottomrule
\end{tabular}
\end{table}

\paragraph{Post-routing quality.}
Routing is only useful if it does not harm task quality. Table~\ref{tab:mol_bench} compares a direct single-adapter baseline against the routed loop (KV reuse off and on) on Vita delivery using five retained seed-level aggregates per arm. On \venti, direct scores $0.636 \pm 0.026$ and routed-with-reuse-off $0.650 \pm 0.030$; routed reuse-on scores $0.632 \pm 0.019$. On \tall{} (Qwen3.6-35B-A3B), direct scores $0.410 \pm 0.030$, routed KV-off $0.398 \pm 0.035$, and routed KV-on $0.386 \pm 0.054$. These small, unpaired five-seed sets show no detected degradation at the reported precision, but they do not establish equivalence.

\begin{table}[t]
\centering
\caption{Three-arm benchmark quality on Vita delivery (temperature~0, 100 tasks per seed). Every row reports five seed-level aggregates per arm; reward is mean $\pm$ sample standard deviation. The arms are a direct single-adapter baseline, the full route--answer--summary loop with cross-turn KV reuse off, and the same loop with reuse on. Seed identifiers are not retained, so the comparison is unpaired.}
\label{tab:mol_bench}
\begin{tabular}{llrl}
\toprule
Model & Arm & $n$ & Reward \\
\midrule
\venti & Direct L1 (no routing)   & 5 & $0.636 \pm 0.026$ \\
\venti & Routed, KV-reuse off     & 5 & $0.650 \pm 0.030$ \\
\venti & Routed, KV-reuse on      & 5 & $0.632 \pm 0.019$ \\
\midrule
\tall & Direct L1 (no routing)   & 5 & $0.410 \pm 0.030$ \\
\tall & Routed, KV-reuse off     & 5 & $0.398 \pm 0.035$ \\
\tall & Routed, KV-reuse on      & 5 & $0.386 \pm 0.054$ \\
\bottomrule
\end{tabular}
\end{table}

\subsection{Per-Adapter Conversation Views}
\label{sec:mol_ownview}

The central serving-time construction is the \emph{own-view}: for each engine hop, the Proxy rebuilds the message list the target LoRA sees, deterministically, from an append-only conversation timeline. Each specialist's own past turns are kept verbatim: the full assistant trace, tool calls, and tool results, while every other specialist's past turns are collapsed to a single assistant message carrying that turn's 192-token summary; the current user turn is appended verbatim. A specialist therefore sees its own raw reasoning history and a compressed record of what other specialists did, never another specialist's full trace. Continuity is preserved without leaking any specialist's private state.

The own-view is what makes routing cheap. Because the timeline is append-only and every own-view is rendered deterministically, re-entering a LoRA on a later turn produces a byte-identical prefix to the previous visit. The engine's native, LoRA-aware prefix cache hits on that prefix and only the newly appended tail is prefilled. Per-adapter KV reuse is thus an emergent property of stable per-adapter prompts; it requires no engine modification on this path, and the Proxy surfaces the hit ratio back to the client as prompt-cache metadata for observability. The own-view is served from two state models behind one interface: a Proxy-authoritative timeline for the stateful Responses API (the agent sends only the current input and a response id), and a stateless side-context rebuilt from the agent-resent history for Chat Completions. Routing is query-only and identical on both; only the own-view reconstruction differs.

\paragraph{Own-view correctness.}
The own-view is load-bearing, so we test the paths that depend on it. The multi-hop switching diagnostic of Section~\ref{sec:mol_router} exercises specialist re-entry after another specialist has run; its aggregate labels verify route re-selection in that smoke test but do not directly measure the semantic fidelity of the reconstructed view. The five-seed Vita result (Table~\ref{tab:mol_bench}: $0.636$ direct, $0.650$ routed KV-off, $0.632$ routed KV-on) shows no detected quality loss from summary-based continuity at this scale. Without paired task-level equivalence analysis, it does not establish identical quality.

\subsection{KV Cache Reuse}
\label{sec:mol_serving}

KV reuse in \molshort{} operates at two layers.

\paragraph{Emergent reuse via stable own-views (production path, no patch).}
As described in Section~\ref{sec:mol_ownview}, the Proxy rebuilds each specialist's own-view so that re-entry yields a byte-identical prefix, letting the engine's native prefix cache absorb the reused portion. Cross-adapter continuity is carried by the 192-token summaries stitched into the own-view; the summary scaffold is never fed into any own-view, so it cannot pollute a reusable prefix. This is the default production path and needs no modification to vLLM or SGLang.

\paragraph{Same-request route-decode overlay (experimental).}
When L0 prefills the router prompt and then hands the same request to the target specialist, a non-invasive overlay trims the router-only tokens and continues decoding under the selected LoRA, reusing the continuous prefix the two share. On vLLM this is realized through the streaming-input session path: the request is issued as two segments, a router segment under L0, then a decode segment under the selected LoRA carrying metadata that pins the shared prefix length, so the engine reuses the overlap and prefills only the divergence. On SGLang the same contract is realized by a runtime monkey-patch over the scheduler and serving classes. A cross-turn flag additionally lets the decode segment read and write the global prefix cache, so a later turn on the same adapter can hit this turn's prefix. The overlay is one instance of the broader family of low-rank inference shortcuts~\citep{zhou2026deputy} that trade a small quality margin for a decode-speed gain.

\paragraph{Boundary.}
Both layers reuse a \emph{continuous} prefix only. We do not implement arbitrary non-contiguous GPU KV splicing across adapters: a specialist switch still invalidates the adapter-specific portion of the KV cache, and the shared prefix is reused only where it is contiguous. The route-decode overlay is an opt-in experimental path; the default deployment runs on the emergent path with unmodified engines.

\paragraph{KV-reuse quality trade-off.}
KV reuse buys latency at a potential cost in quality, because a reused prefix carries forward a prior turn's state rather than re-deriving it from the current context. The three-arm Vita sweep of Table~\ref{tab:mol_bench} isolates this on the emergent path (Layer A): routed reuse-on ($0.632 \pm 0.019$) and routed reuse-off ($0.650 \pm 0.030$) show no detected difference at this scale. The route-decode overlay of Layer B is evaluated only in a legacy GLM-5.1 shim diagnostic (30 turns, both arms $100\%$ accurate), where per-turn wall time is slightly higher with reuse on because of prefix-trim overhead; this is a functionality check, not a current-model quality estimate.

\subsection{MoL Deployment}
\label{sec:mol_deployment}

The deployment study asks a systems question distinct from the quality results
above: what is the cost of representing several specialists as adapters on one
shared model rather than as several independently deployed models? We compare
the MoL representation with a replicated-base layout in which each of the four
LoRA specialists is merged into a separate copy of the base. MoL instead keeps
one base resident and exposes the specialists as runtime adapters. The two
layouts therefore share the same capability partition at the interface level,
while differing in parameter sharing and request scheduling. Detailed profiles
and measurements are collected in Appendix~\ref{app:mol_deployment}.

\paragraph{Weight residency and capacity.}
For the \venti{} configuration, the public adapter headers contain
7,688,042,496 stored values in each of the four LoRA updates. MoL therefore
stores one nominal 744B base plus about 30.8B adapter values (approximately
774.8B logical parameters), while the replicated-base layout stores four
merged copies of the base (2.976T parameters). On this logical count, MoL is
about 26.0\% of the replicated layout, a 74.0\% reduction in stored parameter
values. The release-facing 748B label is not used for this
residency calculation, because it does not equal the stored tensor count.
This is a structural consequence of sharing the base and is independent of the
attention backend or parallelism layout. In the measured MoL deployments, the
remaining device memory supports KV cache and concurrency: H20 sustains sixteen
concurrent 56K-token requests, eight 180K-token requests, or four 230K-token
requests, while B300 DCP2, DCP4, and DCP8 with EAGLE~\citep{eagle2024} enabled provide
approximately 2.34M, 4.67M, and 9.34M logical KV tokens, respectively.

\paragraph{Long-context execution.}
The long-context path combines page-level context parallelism for prefill,
decode context parallelism for KV capacity, and prefill--decode disaggregation.
On eight B300 GPUs, CP8 LayerSplit reduces cold needle-test TTFT at 900K tokens
from 107.1\,s to 49.2\,s while preserving the sparse attention indexer's
address domain. With DCP8 and EAGLE, the deployment reaches
8.6\,ms mean TPOT and 110 tokens/s output throughput at concurrency~1, and
18.0\,ms mean TPOT and 757 tokens/s at concurrency~16, without output
corruption in the reported stress tests. The EAGLE-enabled path reserves part
of the available KV capacity for its draft model and cache. Supporting capacity
and latency measurements are reported in Appendix~\ref{app:mol_deployment}.

\paragraph{Correctness and evidence boundary.}
Serving efficiency depends on the joint choice of engine, attention backend,
parallelism layout, speculative decoding, and request load. We therefore report
only measurements whose operating points are sufficiently specified, rather
than ranking heterogeneous validation snapshots. In the backend study,
FlashMLA sparse attention produced clean outputs for all 48 tested long-context
configurations on each of GLM-5.1 and GLM-5.2, whereas the default DSA decode
path produced clean outputs in only 6 of 48 GLM-5.1 configurations. Replicated
DCP layouts also passed their reported correctness checks, while the sharded
DCP path exhibited systematic corruption. These observations determine the
validated deployment envelope; they are not a cross-engine throughput
comparison. Prefill--decode disaggregation remains part of the production
profile, but heterogeneous PD snapshots are excluded from the quantitative
comparison because they differ in engine, load, LoRA residency, and speculative
decoding configuration.

These results establish the principal systems advantage of MoL over the
replicated-base layout: it removes replicated base-weight residency and makes a
multi-specialist, long-context service feasible. They do not establish lower
TTFT or higher throughput than independently deployed merged specialists;
accordingly, we make no comparative latency or throughput claim for that
deployment alternative.

\subsection{Continual Learning and Collective Intelligence}
\label{sec:mol_continual}
\label{sec:mol_collective}

A frozen base plus a plug-in adapter registry provides an implementation substrate for separately versioned specialists and, in principle, cross-owner composition~\citep{macaron_v1_blog2026,lu2026announcing}. These are architectural properties, not evidence of improvement across generations or capability gains from a specialist population. Three properties follow directly:
\begin{itemize}[leftmargin=*]
  \item \textbf{Adapter registration.} A new LoRA can be trained and registered without retraining already deployed weights. Whether it improves the intended capability while preserving end-to-end behavior still requires evaluation.
  \item \textbf{Base-weight immutability.} Because gradients only enter adapters, the base weights cannot drift as a side effect of specialization. This does not guarantee retained system behavior when adapters, routing, or the harness change.
  \item \textbf{Live harness updates.} The harness layer (routing rules, tool exposure, HCP configs; see Section~\ref{sec:harness}) can be edited without redeploying weights. New tasks become tractable through a harness change first, and only get baked into an adapter once the trajectory data is worth training on.
\end{itemize}
This decouples the release cadence of the base, of specialists, and of the harness: each moves on its own clock.

Because the base is shared and adapters are portable, the registry can admit specialists trained by different teams or personalized for different users, subject to compatibility and trust checks. The current release evaluates only the four shipped specialists and does not test a cross-owner population. Two intended directions follow:
\begin{itemize}[leftmargin=*]
  \item \textbf{Multi-team composition.} A team could ship a specialist targeting a domain we do not cover in-house, with the router registry deciding when to hand off. Such a release would require compatibility, provenance, and tool-visibility checks that are outside the present evaluation.
  \item \textbf{Personalization at the adapter layer.} A user-specific adapter could be mounted alongside the shipping specialists. \mint's million-entry adapter catalog~\citep{lu2026announcing,scaling_peft_2026} supplies the addressing mechanism, but this release does not evaluate personalized-adapter quality, privacy, or cross-user composition.
\end{itemize}
We view \molshort{} less as a way to make one model bigger and more as an interoperability contract for composable agent capability.

\subsection{Discussion}
\label{sec:mol_discussion}

\molshort{} is a bet on \emph{orchestration over merging}. We do not merge adapter weights into a single forward pass, and we do not stack them additively. Composition happens at the request level via the routing loop, and continuity happens via short summaries retained server-side plus the deterministic own-view that makes per-adapter prefix reuse emerge for free. In the 48-request timing profile, the routing and summary hops average $0.54$\,s and $0.97$\,s, together $\sim32\%$ of the three-hop total. The router reaches $99.12\%$ on its trace, while the small Vita comparison detects no reuse-related degradation (reuse-on $0.632 \pm 0.019$ versus reuse-off $0.650 \pm 0.030$). Three practical consequences follow: (i) logs attribute each answer to one selected specialist; (ii) adapter-specific updates leave other adapter weights unchanged, although routing and harness changes can still affect end-to-end behavior; and (iii) new composition patterns can be implemented in the harness without merging adapter weights (Section~\ref{sec:harness}).

\paragraph{Further KV reuse.}
The current reuse design reuses L0's KV for the routing portion of each turn, and otherwise lets each specialist reuse only its own prefix. A more ambitious design, which we have sketched but not yet shipped, would let every specialist reuse L0's full KV across the board. Because the own-view already reconstructs each specialist's context from the shared conversation timeline, L0's KV is a natural shared substrate: routing it to every specialist would remove the need for each specialist to carry its own copy of the other specialists' summaries, trading the per-adapter summary redundancy for a single shared prefix. We do not commit to this design here; it raises correctness questions about mixing a chat-adapter's KV with a specialist's own computation, but it is the direction in which the own-view architecture most naturally extends.

\paragraph{Single-input multi-intent requests.}
The routing loop assumes one intent per user turn: it routes the whole turn to one specialist. In practice a user often packs several intents into one message: discuss a life task, then request code, then ask for a UI to show it, and the production system handles this by routing to a single specialist and letting the conversation naturally segment over subsequent turns. This is a deliberate simplification, not the ceiling of the design. The general case calls for an orchestrator or planner that decomposes a multi-intent turn into a sequence of specialist sub-tasks and composes their outputs, and we have an exploratory branch that implements this style of decomposition. It is not yet in production: we are designing a more uniform and maintainable formulation before shipping it. We report the exploration here as evidence that the architecture admits this extension, and as the direction in which the per-turn routing loop generalizes.

\section{Algorithm: Model--Harness Co-design and Recursive Self-Improvement}
\label{sec:algorithm}

An LLM agent emits text; a harness turns that text into actions~\citep{wu2026replharnesses}. Everything visible to the user, what tools exist, how a UI renders, how a tool result is fed back into the model, lives in this layer. In \macaron{} the harness is a first-class training target: we treat train--serve harness divergence as a bug to fix at the source rather than as a modeling problem to paper over~\citep{macaron_v1_preview2026}. The harness defines the action interface and the runtime contract the model is trained against; the recursive self-improvement loop then optimizes the model--harness pair over that contract. This section describes the two halves together: first the harness components, then the RSI cycle that runs against them.

\subsection{Model--Harness Co-design}
\label{sec:harness}
\label{sec:harness_matters}

A harness is the substrate that determines how much of an agent's computation can be \emph{executed} rather than \emph{re-described}~\citep{wu2026replharnesses,codeact2024}. Discrete substrates like one-JSON-tool-per-turn~\citep{patil2025berkeley} force every intermediate value back through the model as text; expressive substrates keep computation in a state the model never restates. Holding model, tasks, primitives, and budgets fixed, changing only the substrate changes both success rate and token cost, sometimes by large factors~\citep{wu2026replharnesses}.

Two design commitments follow from this observation and shape everything in this subsection:
\begin{itemize}[leftmargin=*]
  \item \textbf{The harness should get lighter as the model gets stronger.} A brittle schema that carries the model is a ceiling. A harness that expresses ordinary programmer-shaped code lets a general model do most of the work, and reserves specialist adapters for judgment calls~\citep{ui4a2026}.
  \item \textbf{Training and serving share an explicit runtime contract.} \mindforge{} (Section~\ref{sec:rsi}) records the HCP and action substrate used by an evaluation backend. Reusing those artifacts makes configuration drift visible; the backend remains responsible for reproducing the production tool surface.
\end{itemize}

\subsubsection{\uifora: A Component-Native Harness for Generative UI}
\label{sec:harness_ui4a}

\begin{figure}[t]
  \centering
  \includegraphics[width=0.95\textwidth]{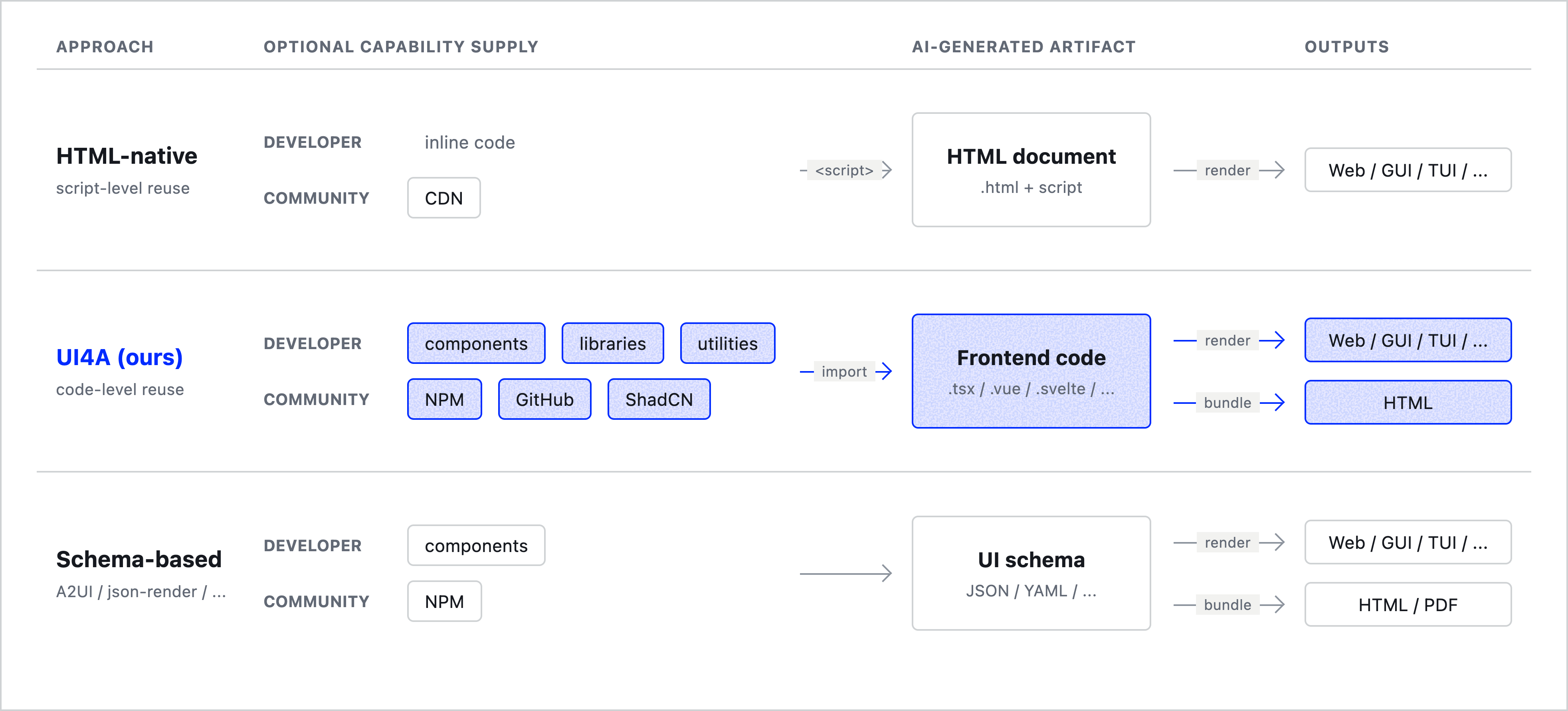}
  \caption{Three approaches to generative UI. HTML-native maximizes expressiveness but inherits raw web development's failure modes; schema-native is verifiable but capped by the component catalog; \uifora{} (component-native) lets the model write ordinary frontend code under runtime-enforced boundaries, recovering expressiveness without the schema ceiling.}
  \label{fig:ui4a_comparison}
\end{figure}

Generative UI~\citep{leviathan2025generative,chen2025generative} has historically been approached in two ways, each with a ceiling (Figure~\ref{fig:ui4a_comparison}).
\begin{itemize}[leftmargin=*]
  \item \textbf{HTML-native.} Maximum expressiveness, minimum control~\citep{si2025design2code}; the harness inherits every compiler, bundler, and error-recovery problem of raw web development.
  \item \textbf{Schema-native.} Verifiable and auditable but capped by the catalog~\citep{a2ui_v08,macaron_a2ui2026}. The schema's ceiling caps the model's ceiling.
\end{itemize}
Our prior schema-native work (Macaron-A2UI~\citep{macaron_a2ui2026}) taught us the point that most affects a shipping agent: teaching the model \emph{when} to render a UI matters more than teaching it \emph{how}.

\begin{figure}[t]
  \centering
  \includegraphics[width=\textwidth]{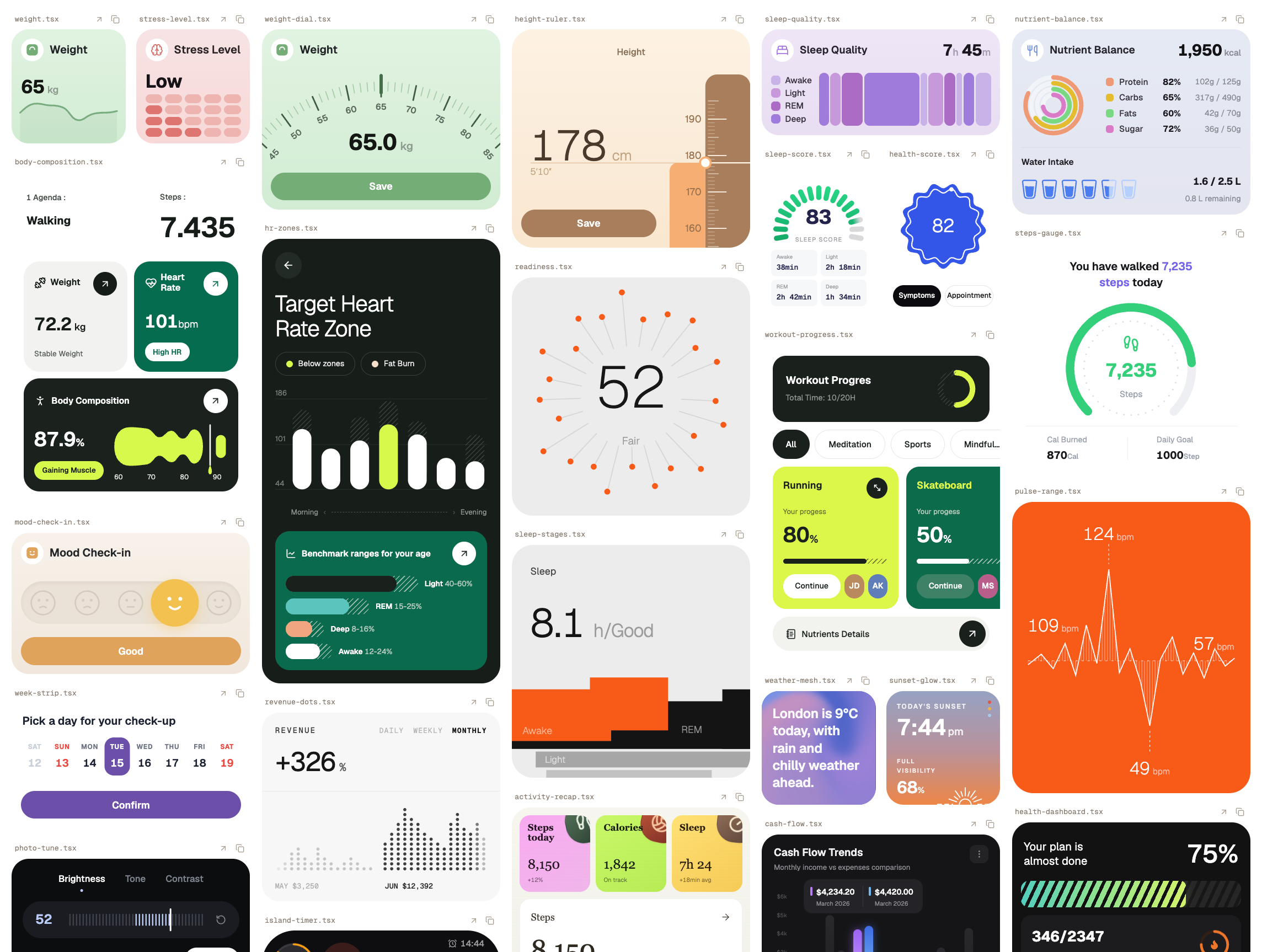}
  \caption{A slice of the \uifora{}-Bench gallery. Each card is generated as ordinary frontend code under runtime-enforced boundaries; \uifora{}-Bench scores them on compile/render correctness, control wiring, and screen fit. The L3 specialist decides \emph{when} to render; the substrate carries \emph{how}.}
  \label{fig:genui_cases}
\end{figure}

\uifora{}~\citep{ui4a2026} is \macaron's component-native answer. The agent writes ordinary frontend code with imports, components, state, and functions, inside runtime-enforced boundaries. Its mental model is \emph{import + component + state + Action}:
\begin{itemize}[leftmargin=*]
  \item \textbf{Import.} The agent pulls components from a curated Macaron registry or common ecosystem primitives (Radix, shadcn, Recharts, KaTeX, lucide-react), with raw HTML/CSS available when a case demands it.
  \item \textbf{Component and state.} Ordinary frontend code the model already knows how to write; no proprietary schema envelope.
  \item \textbf{Action contract.} Every user gesture is a structured object with four fields: \emph{Origin} (which surface), \emph{State} (what data it read), \emph{Execution} (local function or agent event), and \emph{Visibility} (including a \verb|NoAI| boundary for fields the model must not see). The runtime decides whether to run locally, prompt the user, or dispatch as an agent event.
\end{itemize}

Two axes of flexibility follow. The agent can scale \emph{down} to the curated registry for efficiency, or scale \emph{up} to arbitrary frontend code (LaTeX, 3D, generative visuals) when the case genuinely requires it~\citep{ui4a2026}. Rendering is framework-agnostic: React, Vue, Svelte, and SolidJS render from the same output; adding a target ships a small renderer rather than changing the protocol.

Because the code the model produces looks like what a frontend engineer would already write, the base model handles most of the work. \venti's L3 adapter (Section~\ref{sec:mol_specialists}) specializes in \emph{when to render} and in picking and binding components correctly. Adapter routing is optional: when the host model does not support adapters, the harness runs against the base at higher token cost and latency. On a 48-case internal gallery we measure raw HTML at \(\sim\)1{,}224 output tokens versus \uifora{} at \(\sim\)672, a \(\sim\)45\% reduction~\citep{ui4a2026}; combined with a non-reasoning LoRA, streaming, and partial rendering, we observe up to \(\sim\)6\(\times\) faster time-to-first-render on interactive cases.

\subsubsection{REPL Agent Harness: Executable Composition and Validated Reuse}
\label{sec:harness_repl}

Agent workloads lean on the substrate harder than UI does: most of an agent's work is intermediate values handed between tool calls, and the substrate decides whether they stay resident in the runtime or return through the model as text. Our agent harness is a REPL~\citep{wu2026replharnesses}, meaning a stateful \emph{read--eval--print loop} rather than a new agent category: a persistent Python namespace, one rung above discrete function calling~\citep{patil2025berkeley}, MCP endpoints~\citep{mcp2024}, and shell commands. It is the action surface behind the L1 Agent specialist (Section~\ref{sec:mol_specialists}).

\begin{figure}[t]
  \centering
  \includegraphics[width=0.9\textwidth]{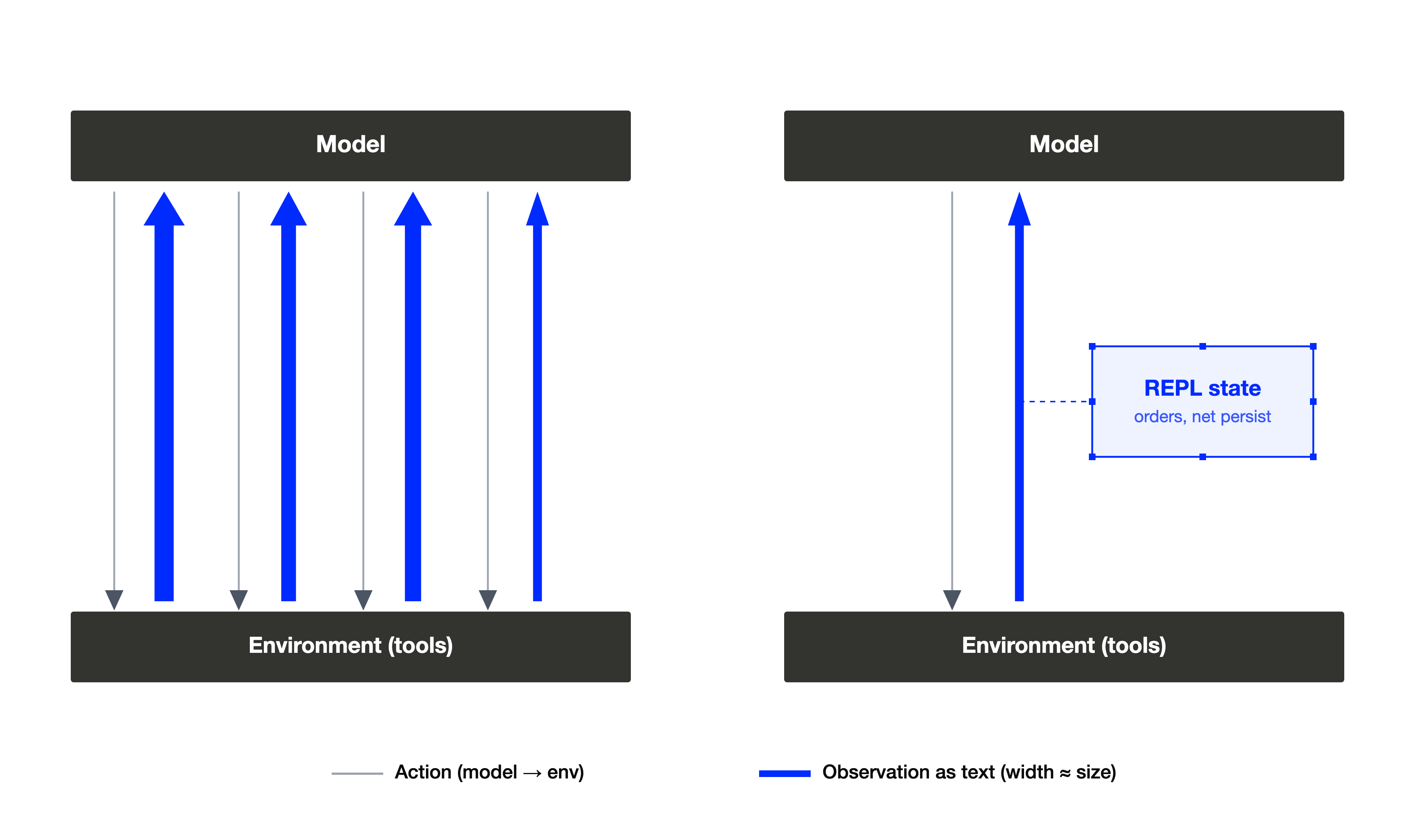}
  \caption{\textbf{Executable composition.} Under function calling (left), each intermediate observation returns through the model as text: one round-trip per dependency step. In the REPL (right), dependent values persist in executable state (\texttt{orders}, \texttt{net}), so the same chain resolves in a single round-trip. Arrow thickness is observation size.}
  \label{fig:repl_composition}
\end{figure}

The first mechanism is \textbf{executable composition}: dependent values persist as variables, so a chain of dependent operations resolves inside a single turn, and the model never restates an intermediate it has already computed (Figure~\ref{fig:repl_composition}). Discrete substrates instead pay a round-trip per dependency step~\citep{wu2026replharnesses}; Section~\ref{sec:results_cases} illustrates the resulting turn and token gap on \venti{}.

The second mechanism is \textbf{validated reuse}, which extends composition across tasks. Two operations govern a helper's lifecycle: \verb|save_tool| stages a self-derived helper into a candidate pool, and \verb|promote_tool| makes it callable by later queries, but only after it passes a private validation run against a held-out reference. Promotion after validation is the load-bearing ordering: a promoted helper is shared across a user's later sessions and is live in \mindforge{} rollouts (Section~\ref{sec:rsi_mindforge}), so promoting one unvalidated would turn a one-off local error into a standing fault across both serving and training. A promotion that later proves bad is logged and demoted, not silently retained.

Because promotion is scoped, a user's recurring workflow compiles into a validated helper in their own namespace, captured without a gradient step.

External services enter the namespace through a ToolProxy: Python-shaped wrappers carrying \verb|NoAI| visibility boundaries and retry semantics, so that a model-facing signature matches runtime behavior by construction, the tool-exposure contract of Section~\ref{sec:harness_hcp}~\citep{macaron_v1_blog2026}. \mindforge{} imports this same harness object rather than a stand-in, which forecloses train--serve tool drift at the source. The REPL is not universally best: stateful observe-before-commit APIs~\citep{patil2025berkeley} penalize committing several dependent calls blind, and independent shell tasks~\citep{terminalbench2026} give it nothing to compose; there the harness lets the L1 agent fall back to shell or discrete calls.

\subsubsection{Harness Context Protocol}
\label{sec:harness_hcp}

Real-world agent runs depend on configuration that is otherwise scattered across command-line flags, environment setup, local files, and runtime-specific defaults. The \hcp{} (HCP) is a versioned TOML contract for recreating that runtime from a portable, auditable artifact~\citep{hcp_sdk2026}. Its purpose is narrower and more operational than a generic ``context configuration'': an HCP producer writes a run specification, and a consumer resolves it in a deterministic order before creating the agent session.

The concrete surface HCP standardizes:
\begin{itemize}[leftmargin=*]
  \item \textbf{Runtime and model selection.} Working directory, backend, provider, model identifier, context window, generation limits, and provider options.
  \item \textbf{Action surface.} Tool allowlists, MCP servers, extensions, hooks, and the policies that control what the agent may invoke.
  \item \textbf{Context resources.} System prompts, \texttt{AGENTS.md}-style files, skills, prompt templates, embedded resources, and their deterministic materialization.
  \item \textbf{Session and workspace state.} Session snapshots plus optional workspace inputs, visibility, snapshot policy, and declared outputs.
  \item \textbf{Environment contract.} Required and optional environment names, path-resolution rules, and secret references without embedding portable credentials.
\end{itemize}

The current Python SDK operates in wrapper mode: it validates the outer configuration, asks \texttt{pi-hcp} to prepare the runtime, launches the ACP process, and manages a session through the Agent Client Protocol~\citep{hcp_sdk2026}. Full semantic validation and native Python materialization remain outside the SDK. In the RSI loop, HCP therefore serves as the versioned boundary between configuration search and execution.

It is worth being precise about what is and is not trainable here, because the two are easy to conflate. The protocol carries no gradients: an HCP artifact is a declarative TOML document, and no optimizer ever writes to it. What the model \emph{can} do is rewrite the harness the protocol describes. Because prompts, skills, tool allowlists, hooks, and workspace resources are all addressable fields in a portable artifact, a model version can propose edits to them, have the edits evaluated on a re-run of the affected slice, and ship the survivors as the next configuration, self-iteration of the harness conducted in language space rather than in parameter space (Section~\ref{sec:rsi_loop}). Neither half is sufficient alone. Configuration search moves in discrete jumps and cannot acquire a skill the base does not already have, though Section~\ref{sec:rsi_coverage} measures how much of what scores as a missing skill is an unelicited one; adapter training acquires skills but cannot reach the tool surface or the instructions that decide when to use them. Coupled through \mint{}, which parameterizes the specialists as LoRA revisions over a frozen base (Section~\ref{sec:infra_mint})~\citep{lu2026announcing}, the pair becomes one trainable system: a candidate HCP is what a rollout executes against, the resulting trajectories are what the adapters are trained on, and \mindforge{} ships the two as a linked model--configuration version. The trainable object is the model--harness pair, not the protocol.

\subsubsection{Live Harness Updates Enable Continual Learning}
\label{sec:harness_live}

Section~\ref{sec:mol_continual} described the frozen base and adapter registry as
an architectural affordance for continual learning. The harness supplies the
operational revision path: a new capability may first appear as a registered
tool, HCP configuration, \uifora{} component, or REPL helper promoted after
validation. When trajectory evidence warrants a parameter update, the change can
then be transferred into a specialist adapter. The present evaluation measures
the configuration-search stage, not repeated transfer across adapter generations.

This gives \macaron{} three release clocks: the base (tied to platform-level models such as GLM-5.2~\citep{glm52_2026,glm5_2026}), the specialists (updated through the RSI process in Section~\ref{sec:rsi}), and the harness. Harness changes do not require a weight update, so they can be evaluated and released on a shorter cycle than a new base or specialist revision.

\subsection{Recursive Self-Improvement}
\label{sec:rsi}

\begin{figure}[t]
  \centering
  \includegraphics[width=\textwidth,height=0.4\textheight,keepaspectratio]{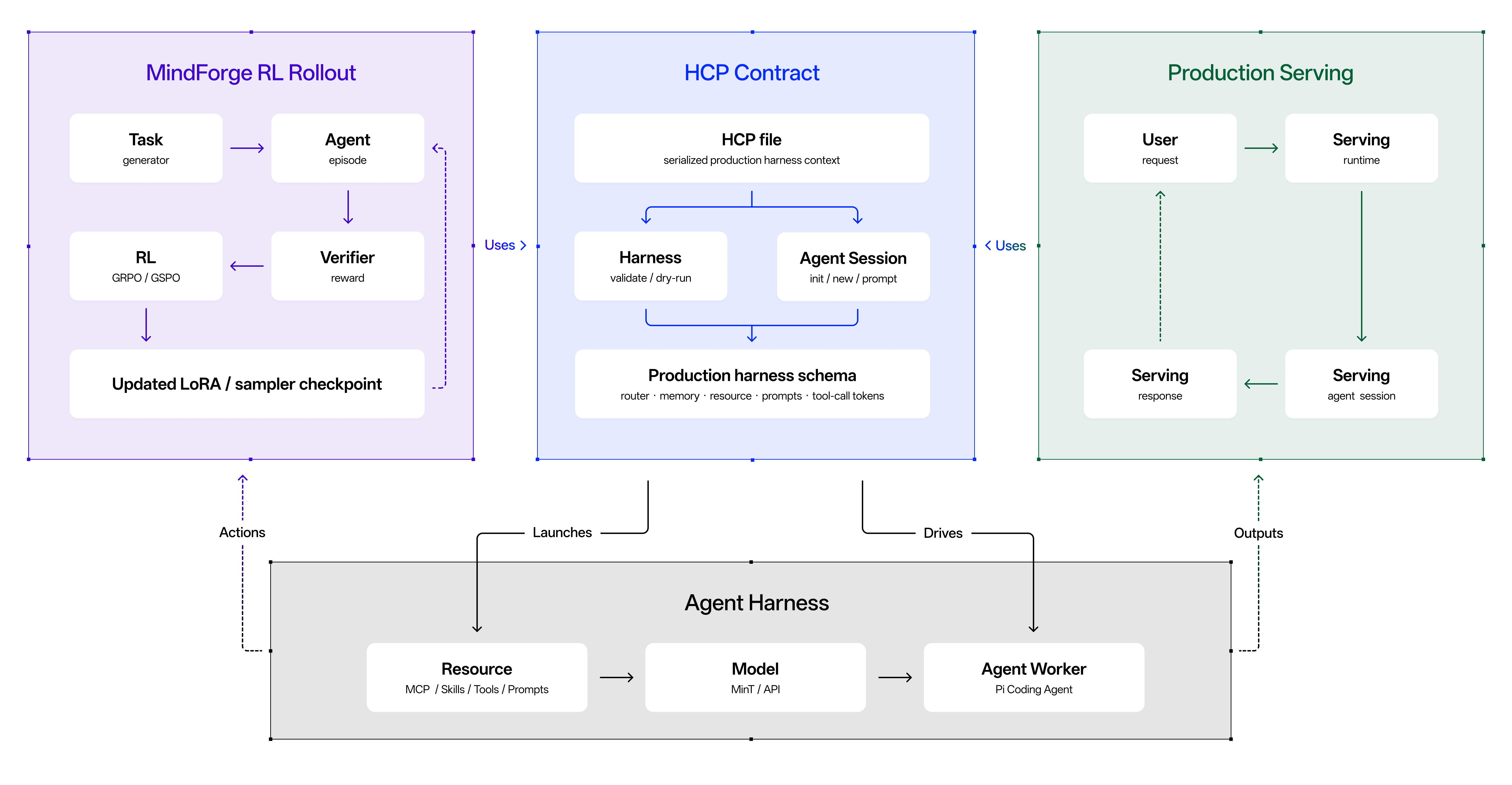}
  \caption{The three loops that share one harness. \mindforge{} runs RL rollouts against the same agent harness used for production serving; an HCP contract serializes the harness context (router, memory, resources, prompts, and tool-call tokens) so training and serving configurations can be compared explicitly. Matching the recorded contract makes configuration drift auditable but does not guarantee behaviorally identical executions.}
  \label{fig:system_overview}
\end{figure}

\macaron{} is improved through a versioned data-generation and training cycle rather than a fixed post-training corpus. The cycle runs against the production harness described above: a model version proposes tasks beyond its current competence, attempts them in an executable agent environment, and produces evaluated trajectories. Those trajectories support two different forms of improvement: language-space search over the runtime configuration, and parameter updates to the specialist adapters. \mindforge{} maintains the lineage connecting problem banks, evaluation runs, selected trajectories, training jobs, configurations, and model versions. This subsection describes the optimization object, the role of the framework, and the recursive self-improvement (RSI) cycle that connects them (Figure~\ref{fig:system_overview}).

\subsubsection{Agent RL on a Frozen Sparse Base}
\label{sec:rsi_formulation}

Let \(\theta\) denote the parameters of the frozen sparse base, \(\phi\) the trainable LoRA parameters, and \(c\) a versioned harness configuration. The agent policy can be written as
\begin{equation}
  \pi_{\phi}(a_t \mid o_{\leq t}; \theta, c),
\end{equation}
where \(o_{\leq t}\) contains the conversation and observations exposed by the harness, and \(a_t\) is a model-visible action. An action may be a message, a clarification, a discrete tool call, or an expression evaluated by the stateful read--eval--print loop defined in Section~\ref{sec:harness_repl}. Adapter selection is a separate Proxy-mediated route hop (Section~\ref{sec:mol_router}), not a tool call made from this action space. The resulting episode is recorded as
\begin{equation}
  \tau=(o_0,a_0,\ldots,o_T,a_T,y),
\end{equation}
where \(y\) contains the task-level outcome and any process-level judgments attached by the evaluator.

This factorization distinguishes two update paths that are easy to conflate. Model optimization changes \(\phi\) while keeping \(\theta\) fixed; configuration search selects a new \(c\) without treating HCP as a learned parameter. The training backend uses GRPO~\citep{shao2024deepseekmath} to update the adapters from selected trajectories. Sparse-base rollout consistency and long-context execution are handled by the infrastructure in Section~\ref{sec:infrastructure}. This release reports the adapter training configuration (learning rate, batch size, epochs, schedule, and optimizer) in Appendix~\ref{app:training_details}.

The action substrate is part of \(c\), not an incidental wrapper. In the REPL substrate, dependent primitive operations execute inside a persistent namespace, so intermediate values need not be serialized through the model after every call~\citep{wu2026replharnesses}. For stateful APIs that require observing one result before committing the next action, the same runtime can instead expose discrete calls. The policy is therefore trained and evaluated against an explicit action interface rather than an abstract tool-use label.

\subsubsection{\mindforge: Lifecycle Orchestration}
\label{sec:rsi_mindforge}

\mindforge{} is the control plane of the RSI cycle. It manages problem banks, evaluation runs, conversion of evaluated trajectories into trainable datasets, training jobs, and a version registry that links each model to its parent, data, configuration, and evaluation results. Compute-heavy operations are delegated through pluggable benchmark and training backends. The control plane consequently defines lifecycle and provenance, while a backend defines the semantics of task execution, scoring, and optimization.

An HCP artifact provides the backend with a versioned runtime contract (Section~\ref{sec:harness_hcp})~\citep{hcp_sdk2026}. It records model and provider selection, tool policy, skills, prompts, MCP servers, hooks, session state, and workspace staging. The selected action substrate determines how model outputs become executable actions. Recording both artifacts makes a run reconstructable at the configuration boundary: train--serve differences can be inspected as differences in configuration, resources, or backend realization. This is a reproducibility contract, not a claim that two executions are behaviorally identical.

After each update, \mint{} exports the changed LoRA state as a serving-compatible adapter revision instead of moving or merging the frozen base; trajectories, rewards, and routing metadata remain separate rollout records (Section~\ref{sec:infra_mint})~\citep{lu2026announcing}.

This separation also keeps framework claims aligned with implementation. \mindforge{} does not redefine GRPO, the REPL, or HCP. It connects their artifacts into a stable lineage:
\[
  (\text{problem bank},\, \text{model},\, \text{HCP})
  \longrightarrow \text{evaluated trajectories}
  \longrightarrow (\text{dataset},\, \text{next model},\, \text{next HCP}).
\]
The lineage is the unit of an RSI generation. A model checkpoint without its problem-bank version, selected data, and runtime configuration is an incomplete record of the experiment.

\subsubsection{Three-Stage RSI Cycle}
\label{sec:rsi_loop}

\paragraph{Discovery.}
Starting from a versioned seed bank, the current model proposes harder task variants by lifting constraints, introducing hidden preferences, chaining sub-goals, or embedding contradictions. Each proposal must include either a verifiable answer or an evaluation rubric. Candidate tasks are retained only when they satisfy two complementary criteria: \emph{quality}, meaning that the task and its evaluation are well-defined, and \emph{learning value}, meaning that the current model does not already solve the task reliably. The verifier, difficulty estimator, and acceptance thresholds belong to the benchmark backend; \mindforge{} records the resulting bank and its provenance.

\paragraph{Expansion.}
The accepted tasks are executed by the current model under a fixed model--HCP pair. The evaluator scores both task outcome and relevant process behavior, producing a set of trajectories rather than a single answer corpus. Audits then localize failure to the model, the task, or the harness configuration. For example, an audit may reveal unnecessary model round-trips, an action that should have remained discrete, a missing tool boundary, or an instruction that causes premature routing. Candidate changes to prompts, skills, tool exposure, hooks, and other HCP-carried resources are evaluated by re-running the affected slice. Section~\ref{sec:rsi_coverage} reports what this search recovers on a set the base fails outright. Configuration changes are accepted on observed behavior rather than on prompt-author intuition; the acceptance gate is an automatic eval pass against the affected benchmark slice, with human review reserved for changes that touch tool exposure or safety-relevant boundaries.

\paragraph{Update.}
Trajectory selection filters invalid runs, removes duplicates, and retains examples that are both validated and informative for the next model. The selected trajectories become input to the training backend, which updates the LoRA specialists while the sparse base remains frozen. In parallel, the accepted HCP is registered as the runtime configuration of the next generation. \mindforge{} links the new adapter state and HCP to their parent artifacts and evaluation evidence. The resulting model--configuration pair returns to discovery, where its changed competence induces a new task distribution.

The three stages therefore form a feedback system. Discovery adapts the curriculum to the current policy; expansion searches behavior in the executable environment; update transfers validated behavior into parameters while preserving the configuration that produced it. Progress is measured across linked generations, not inferred from the existence of a later checkpoint.

\subsubsection{AutoResearch, Trajectory Selection, and Context Learning}
\label{sec:rsi_autoresearch}

The Preview report~\citep{macaron_v1_preview2026} described self-evolution through AutoResearch, trajectory selection, and what it called \emph{context learning} (validated-trajectory transfer, not in-context learning). In V1 these mechanisms are integrated into the three-stage cycle above.

\emph{AutoResearch} is language-space search inside expansion. The model proposes changes to prompts, skills, scaffolds, or tool-use policy; when represented by HCP resources, each proposal becomes a portable candidate configuration that can be executed and compared. AutoResearch changes the conditions under which behavior is elicited before changing the model weights; Section~\ref{sec:rsi_coverage} measures the reach of that change.

\emph{Trajectory selection} closes expansion by deciding which observed behaviors should influence the next model. A useful trajectory is not merely successful: it must be valid under the evaluator, attributable to the intended task and configuration, and sufficiently informative to justify inclusion. This selection step prevents configuration search, evaluator failures, and duplicated solutions from being indiscriminately distilled into the model.

\emph{Context learning} is the transfer from validated language-space behavior to the parameter update. Selected trajectories provide the training signal for the next adapters, while the accepted HCP remains a separately versioned artifact. Keeping these outputs distinct makes it possible to ask whether a gain came from weights, configuration, or their interaction. This release does not report that quantitative attribution; the controlled generation-by-generation comparison that would isolate it remains required.

\subsubsection{Expansion Coverage on a Base-Failure Set}
\label{sec:rsi_coverage}

The expansion stage searches configuration space against a fixed task set. We measure its reach on 122 simulation tasks grouped under 29 TerminalBench~2.1~\citep{terminalbench2026} source families, selected precisely because the frozen GLM-5.2-FP8 base scores every one of them not-pass under the official reward. The retained artifact does not include the per-family task counts. The base stays frozen throughout: no optimizer step is taken anywhere in the run, and every change is an edit to HCP-carried resources, task skills, tool exposure, or hooks. Because passing zero tasks was the selection criterion, the pre-sweep coverage is $0/122$ by definition, a selected failure slice rather than the base's TerminalBench pass rate.

\begin{figure}[t]
\centering
\begin{tikzpicture}

\begin{axis}[
    name=cov,
    height=4.6cm,
    width=0.93\textwidth,
    axis line style={mindlabmuted},
    tick style={mindlabmuted},
    label style={font=\small, color=mindlabink},
    tick label style={font=\footnotesize},
    grid=major, grid style={mindlabgrid, dashed},
    axis on top,
    xmin=0, xmax=70,
    ymin=0, ymax=104,
    clip=false,
    ylabel={Cumulative coverage (\%)},
    xticklabels={},
    ytick={0,20,40,60,80,100},
]
\fill[mindlabmuted!14] (axis cs:0.5,0) rectangle (axis cs:10.5,104);
\fill[mindlabmuted!8] (axis cs:10.5,0) rectangle (axis cs:12.5,104);
\fill[mindlabbluepale!55] (axis cs:12.5,0) rectangle (axis cs:48.5,104);
\fill[mindlabbluepale] (axis cs:48.5,0) rectangle (axis cs:69.5,104);
\node[anchor=north, font=\scriptsize, color=mindlabink, inner sep=1pt] at (axis description cs:0.5,-0.035) {\textbf{Jobs:} 1--10 Retry control \quad 11--12 Portfolio v1 sweep \quad 13--48 Skill/HCP search \quad 49--69 + Stop-gate hooks};
\addplot[mindlabblue, mark=*, mark size=1.25pt, line width=0.9pt]
  coordinates {
    (1,0.0)(2,0.0)(3,0.82)(4,0.82)(5,0.82)(6,0.82)
    (7,1.639)(8,1.639)(9,1.639)(10,1.639)(11,4.918)(12,11.475)
    (13,11.475)(14,11.475)(15,13.115)(16,13.115)(17,13.115)(18,13.115)
    (19,13.115)(20,13.934)(21,14.754)(22,15.574)(23,18.033)(24,20.492)
    (25,21.311)(26,22.131)(27,22.131)(28,22.951)(29,23.77)(30,24.59)
    (31,25.41)(32,26.23)(33,27.049)(34,27.049)(35,27.869)(36,28.689)
    (37,29.508)(38,30.328)(39,31.148)(40,31.967)(41,36.885)(42,40.984)
    (43,40.984)(44,42.623)(45,46.721)(46,47.541)(47,49.18)(48,49.18)
    (49,53.279)(50,59.016)(51,59.836)(52,63.934)(53,64.754)(54,66.393)
    (55,68.033)(56,69.672)(57,71.311)(58,72.951)(59,74.59)(60,75.41)
    (61,76.23)(62,77.869)(63,82.787)(64,83.607)(65,84.426)(66,89.344)
    (67,96.721)(68,99.18)(69,100.0)
  };
\end{axis}

\begin{axis}[
    name=rate,
    at={(cov.south west)}, anchor=north west, yshift=-0.75cm,
    height=3.9cm,
    width=0.93\textwidth,
    axis line style={mindlabmuted},
    tick style={mindlabmuted},
    label style={font=\small, color=mindlabink},
    tick label style={font=\footnotesize},
    grid=major, grid style={mindlabgrid, dashed},
    axis on top,
    xmin=0, xmax=70,
    ymin=0, ymax=108,
    xlabel={Chronological job step},
    ylabel={Pass rate on\\attempted tasks (\%)},
    ylabel style={align=center},
    ytick={0,25,50,75,100},
]
\fill[mindlabmuted!14] (axis cs:0.5,0) rectangle (axis cs:10.5,108);
\fill[mindlabmuted!8] (axis cs:10.5,0) rectangle (axis cs:12.5,108);
\fill[mindlabbluepale!55] (axis cs:12.5,0) rectangle (axis cs:48.5,108);
\fill[mindlabbluepale] (axis cs:48.5,0) rectangle (axis cs:69.5,108);
\addplot[only marks, mark=*, mark size=1.15pt, color=mindlabmuted, opacity=0.75]
  coordinates {
    (1,0.0)(2,0.0)(3,20.0)(4,20.0)(5,20.0)(6,0.0)
    (7,20.0)(8,20.0)(9,20.0)(10,0.0)(11,3.28)(12,9.02)
    (13,0.0)(14,0.0)(15,66.67)(16,0.0)(17,0.0)(18,0.0)
    (19,0.0)(20,100.0)(21,50.0)(22,100.0)(23,100.0)(24,100.0)
    (25,100.0)(26,33.33)(27,100.0)(28,100.0)(29,100.0)(30,100.0)
    (31,100.0)(32,100.0)(33,100.0)(34,0.0)(35,100.0)(36,100.0)
    (37,100.0)(38,100.0)(39,100.0)(40,100.0)(41,100.0)(42,100.0)
    (43,66.67)(44,100.0)(45,83.33)(46,100.0)(47,28.57)(48,0.0)
    (49,100.0)(50,100.0)(51,14.29)(52,83.33)(53,100.0)(54,100.0)
    (55,100.0)(56,100.0)(57,100.0)(58,100.0)(59,100.0)(60,50.0)
    (61,100.0)(62,100.0)(63,85.71)(64,100.0)(65,100.0)(66,85.71)
    (67,69.23)(68,85.71)(69,100.0)
  };
\draw[mindlabblue, line width=1.4pt] (axis cs:0.5,12.00) -- (axis cs:10.5,12.00);
\draw[mindlabblue, line width=1.4pt] (axis cs:10.5,6.15) -- (axis cs:12.5,6.15);
\draw[mindlabblue, line width=1.4pt] (axis cs:12.5,64.47) -- (axis cs:48.5,64.47);
\draw[mindlabblue, line width=1.4pt] (axis cs:48.5,81.25) -- (axis cs:69.5,81.25);
\node[anchor=south, font=\scriptsize\bfseries, color=mindlabblue, inner sep=1.5pt] at (axis cs:5.5,12.00) {12.0\%};
\node[anchor=south, font=\scriptsize\bfseries, color=mindlabblue, inner sep=1.5pt] at (axis cs:11.5,6.15) {6.1\%};
\node[anchor=south, font=\scriptsize\bfseries, color=mindlabblue, inner sep=1.5pt] at (axis cs:30.5,64.47) {64.5\%};
\node[anchor=south, font=\scriptsize\bfseries, color=mindlabblue, inner sep=1.5pt] at (axis cs:59.0,81.25) {81.2\%};
\end{axis}

\end{tikzpicture}
\caption{\textbf{Expansion over a base-failure set.} 122 simulation tasks from 29
TerminalBench~2.1 source families the frozen GLM-5.2 base fails under the official
reward. \textbf{Top:} cumulative unique coverage, reaching $122/122$ at job 69.
\textbf{Bottom:} each job's pass rate on the tasks it attempted, with the pooled
rate per phase in blue. The model is frozen throughout; only the HCP-carried harness
changes. Jobs 11--12 sweep the full set under one configuration each; later jobs target
the families still uncovered.}
\label{fig:rsi_coverage}
\end{figure}
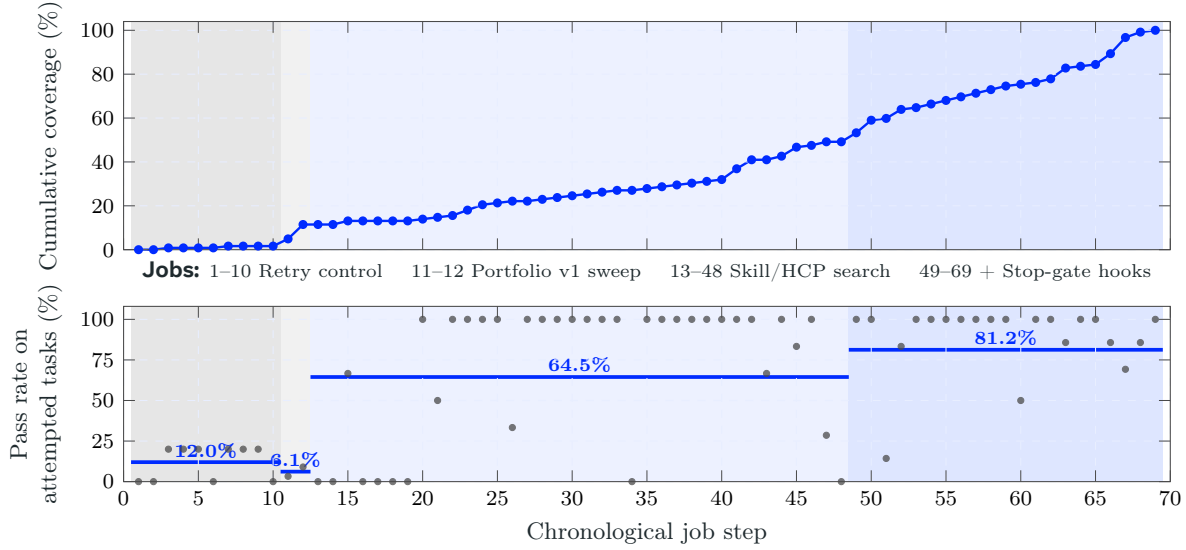

Figure~\ref{fig:rsi_coverage} plots 69 chronological jobs, 450 task attempts in total ($3.69$ per task). Cumulative unique coverage reaches $122/122$ at job 69: each task passes at least once under one of the tested configurations without a weight update. Because a passing configuration was found for every task, failure under the baseline configuration does not by itself mean the behavior is absent from the frozen model. And because the search was adaptive, each job targeted the tasks still uncovered, so $122/122$ is a coverage ceiling under adaptive configuration selection, not a held-out estimate of how any single configuration generalizes.

One configuration does not get there. Jobs 11 and 12 sweep the full set under a single portfolio configuration each and pass $4/122$ ($3.3\%$) and $11/122$ ($9.0\%$); 97 of the run's 109 harness errors fall in those two jobs. What follows is not more sampling from that distribution. Targeting the families still uncovered yields a pooled pass rate of $64.5\%$ over 36 skill-and-HCP jobs, and the final 21 jobs with stop-gate hooks yield $81.2\%$ while closing the remaining 62 tasks with 3 errors (Table~\ref{tab:rsi_coverage}). The ratio between the final phase and the two full-set sweeps is $13\times$ in per-attempt yield. Because the task mix, targeting policy, and harness-error rate differ between phases, this ratio describes the search trajectory rather than isolating a causal effect of the added hooks.

The two numbers illustrate the intended division of labor between the update paths of Section~\ref{sec:rsi_formulation}. The better of the two full-set configurations reaches $9\%$ of this set; adaptive search over $c$ eventually covers all of it at $3.69$ attempts per task. The role assigned to the subsequent $\phi$ update is to transfer selected behavior into an adapter, but this experiment does not execute that update or measure whether the transfer generalizes.

\begin{table}[t]
\centering
\caption{Phase breakdown of the 69-job expansion run of Figure~\ref{fig:rsi_coverage}. Attempts count task trials, so a job sweeping the full set contributes 122. Pooled rate is passes over attempts within the phase; coverage is the cumulative unique tasks passed at the end of it, against the fixed denominator of 122. The model is frozen throughout.}
\label{tab:rsi_coverage}
\footnotesize
\setlength{\tabcolsep}{5pt}
\begin{tabular}{llrrrrr}
\toprule
Phase & Jobs & Attempts & Passes & Pooled & Errors & Coverage \\
\midrule
Retry control        & 1--10  & 50  & 6  & $12.0\%$ & 6  & $2/122$ \\
Portfolio v1 sweep   & 11--12 & 244 & 15 & $6.1\%$  & 97 & $14/122$ \\
Skill/HCP search     & 13--48 & 76  & 49 & $64.5\%$ & 3  & $60/122$ \\
$+$ Stop-gate hooks  & 49--69 & 80  & 65 & $81.2\%$ & 3  & $122/122$ \\
\midrule
Total                & 1--69  & 450 & 135 & $30.0\%$ & 109 & $122/122$ \\
\bottomrule
\end{tabular}
\end{table}

\FloatBarrier

\section{Infrastructure}
\label{sec:infrastructure}

The RSI cycle in Section~\ref{sec:rsi} links a new LoRA adapter revision to the
HCP configuration, selected data, and evaluation artifacts that produced it.
This section concerns the model-state side of that lineage and the computation
that produces the revision. \mint{} manages the model-state
lifecycle over a resident, frozen base; \longstraw{} provides a response-only
execution path when an update has a very long context; and model-dependent
controls limit rollout--training mismatch on sparse bases. These components do
not change the learned unit defined earlier: the deployable model update remains
an exported LoRA revision, paired by \mindforge{} with the HCP and evaluation
artifacts that produced it.

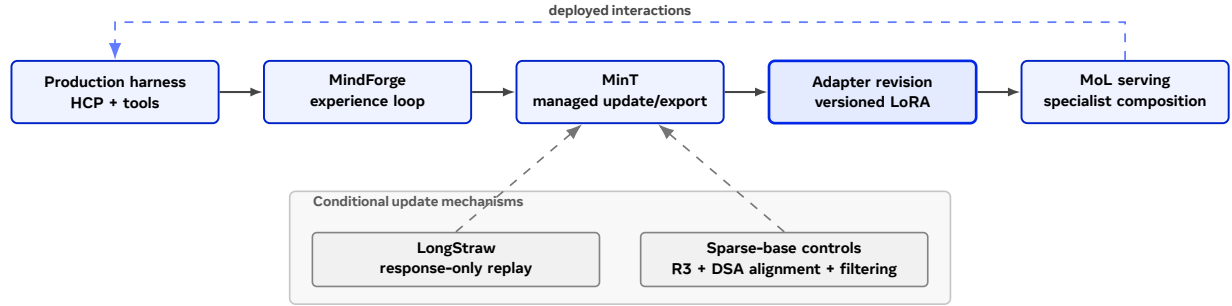
\begin{figure}[h]
\centering
\resizebox{0.98\textwidth}{!}{\begin{tikzpicture}[
    font=\sffamily\fontsize{5.8}{6.9}\selectfont,
    main/.style={draw=mindlabblue!78!black, fill=mindlabbluepale!48,
      rounded corners=2pt, line width=0.65pt, align=center,
      text width=2.18cm, minimum height=0.74cm, inner sep=4pt},
    revision/.style={main, draw=mindlabblue!90!black,
      fill=mindlabbluepale!78, line width=0.9pt},
    support/.style={draw=mindlabfg!52, fill=mindlabfg!6,
      rounded corners=1.5pt, line width=0.55pt, align=center,
      text width=3.20cm, minimum height=0.64cm, inner sep=3pt},
    band/.style={draw=mindlabfg!24, fill=mindlabfg!3,
      rounded corners=2pt, line width=0.55pt},
    flow/.style={
      -{Latex[length=1.5mm,width=1.1mm]},
      draw=mindlabfg!78, line width=0.7pt
    },
    feedback/.style={-Latex, draw=mindlabblue!62, dashed, line width=0.6pt},
    dependency/.style={-Latex, draw=mindlabfg!58, dashed, line width=0.55pt},
    label/.style={font=\sffamily\fontsize{5.3}{6.2}\selectfont,
      fill=white, inner sep=1.2pt,
      text=mindlabfg!78, align=center}
  ]
  
  \node[main] (harness) at (-6.0,0.95)
    {{\bfseries Production harness}\\HCP + tools};
  \node[main] (mindforge) at (-3.0,0.95)
    {{\bfseries MindForge}\\experience loop};
  \node[main] (mint) at (0,0.95)
    {{\bfseries MinT}\\managed update/export};
  \node[revision] (revision) at (3.0,0.95)
    {{\bfseries Adapter revision}\\versioned LoRA};
  \node[main] (serving) at (6.0,0.95)
    {{\bfseries MoL serving}\\specialist composition};
  
  \draw[flow] (harness.east) -- (mindforge.west);
  \draw[flow] (mindforge.east) -- (mint.west);
  \draw[flow] (mint.east) -- (revision.west);
  \draw[flow] (revision.east) -- (serving.west);
  
  \draw[feedback] (serving.north) -- (6.0,1.78) --
    node[label, above] {deployed interactions} (-6.0,1.78) -- (harness.north);
  
  \node[band, minimum width=7.85cm, minimum height=1.34cm]
    (supportband) at (0,-0.90) {};
  \node[anchor=west, font=\sffamily\bfseries\fontsize{5.3}{6.2}\selectfont,
    text=mindlabfg!70] at (-3.77,-0.36) {Conditional update mechanisms};
  
  \node[support] (longstraw) at (-1.95,-1.04)
    {{\bfseries LongStraw}\\response-only replay};
  \node[support] (consistency) at (1.95,-1.04)
    {{\bfseries Sparse-base controls}\\R3 + DSA alignment + filtering};
  
  \draw[dependency] (longstraw.north) -- ($(mint.south)+(-0.45,0)$);
  \draw[dependency] (consistency.north) -- ($(mint.south)+(0.45,0)$);
  \end{tikzpicture}}
\caption{\textbf{Infrastructure inside the Macaron revision loop.} This is a
conceptual artifact and control flow, not a claim that all components are
co-located or invoked for every update. \mindforge{} supplies selected
trajectories and identifies the current policy; \mint{} resolves its policy
record, manages the update, and exports an immutable LoRA adapter revision for
the \molshort{} runtime. \longstraw{} is
selected for long-context response-only updates; sparse-base consistency
controls are selected according to the model architecture. The dashed feedback
arrow denotes experience collected after deployment, not an automatic online
weight update~\citep{lu2026announcing,zhou2026longstrawlongcontextrl2m}.}
\label{fig:macaron-infra-loop}
\end{figure}

Figure~\ref{fig:macaron-infra-loop} separates lifecycle management from the
optional, architecture-specific execution mechanisms inside an update.
\FloatBarrier

\subsection{MinT: Model-State Lineage over a Resident Base}
\label{sec:infra_mint}

The MindLab Toolkit (\mint{})~\citep{lu2026announcing} manages LoRA RL over
resident dense and Mixture-of-Experts (MoE) base deployments. Its key
distinction is between an \emph{adapter revision}, an immutable LoRA snapshot
exported at a particular training step in serving tensor layout, and a
\emph{policy record}, the mutable service state used to resume and audit that
policy. The policy record names the compatible base version and adapter shape,
the latest trainer checkpoint and optimizer state, associated rollout records,
and the available exported revisions. Training restores mutable state on a
compatible worker, whereas rollout, evaluation, and serving select an exported
revision; trainer checkpoints and optimizer state never cross the serving
boundary directly.

This abstraction provides model-state lineage, not complete experimental
reproducibility by itself. Prompts, tools, hooks, workspace resources, task-bank
versions, evaluators, and sampling settings remain in the HCP and \mindforge{}
artifacts defined in Section~\ref{sec:rsi_mindforge}. A reported run is therefore
reconstructable at the paper's configuration boundary only when its adapter
revision and compatible base are joined with those artifacts. The narrower
\mint{} guarantee is that a rollout or score can be attributed to a fixed
adapter revision even if trainer placement and serving-cache residency change.

Adapter revisions also make the training--serving handoff smaller than a merged
checkpoint handoff. The companion \mint{} experiment compares a merge path that
materializes and loads a full checkpoint before its rollout probe with an
adapter path that loads the exported revision into a sampler whose base is
already resident. Under those stated probes, adapter-only handoff reduces the
measured handoff step by $18.3\times$ for Qwen3-4B with a rank-32 adapter and by
$2.85\times$ for Qwen3-30B with a rank-16 adapter~\citep{lu2026announcing}.
These are path-specific latency ratios, not general training-speed or inference-
throughput improvements.

The same separation supports a large addressable catalog without requiring a
large resident set. The \mint{} evaluation builds all $10^6$ entries of a packed
Qwen3-30B rank-1 adapter catalog with zero build errors, then reads a 256-entry
audit sample spanning all 100 storage shards. Serving experiments select bounded
working sets from that catalog while each actor maintains a smaller CPU cache
and GPU-active window~\citep{lu2026announcing}. Thus, $10^6$ is a measured
artifact-addressability result. It does not mean that one engine holds one
million adapters in GPU memory, nor that the entries represent one million
independently trained or behaviorally distinct policies. Cold loading remains
scheduled service work, and a revision becomes user-visible only after
compatibility checking and activation. This is the catalog boundary assumed by
the registry in Section~\ref{sec:mol_collective}~\citep{punica2024,slora2023}.

For bases that require distributed placement, \mint{} keeps model-parallel
trainer groups and inference actors resident while adapter tensors, optimizer
state, and exported revisions move between workers. The companion report
includes a Kimi K2 countdown-task LoRA RL run on a 1.04T-total/32.6B-active
parameter base using 64 H800 GPUs~\citep{lu2026announcing,kimi_k2_2025}. This is
evidence that the adapter lifecycle executes on that model and workload; it is
not evidence that \macaron{} was trained on Kimi K2, nor that the run is
reproducible from this paper alone.

\subsection{LongStraw: Response-Only Long-Context Execution}
\label{sec:infra_longstraw}

\paragraph{Response-only memory boundary.}
Long contexts make a conventional full-sequence autograd implementation
memory-bound because it retains differentiable state across the prompt and
response. For a shared prompt of length $P$ and a GRPO group of $G$ responses
with lengths $R_i$, \longstraw{}~\citep{zhou2026longstrawlongcontextrl2m}
instead captures architecture-specific prompt state without autograd and
replays one response at a time with autograd. Its live-memory boundary is
approximately
\begin{equation}
  M_{\mathrm{live}} \approx M_{\mathrm{weights}} + M_{\mathrm{prefix}}(P)
  + M_{\mathrm{grad}} + \max_i M_{\mathrm{graph}}(R_i)
  + M_{\mathrm{score}}\!\left(\sum_i R_i\right),
  \label{eq:longstraw-live-memory}
\end{equation}
The resulting graph is bounded by the longest response rather than by a
prompt-and-response graph for all group members. The full prompt still incurs
forward compute and retained model state; the method bounds graph lifetime
rather than eliminating context-dependent memory or computation.

\paragraph{Three-stage update.}
One update has three stages. First, the current policy evaluates the shared
prompt without autograd and retains only the state needed to condition response
tokens. Second, old-policy and reference scores are computed without a graph,
after which policy responses are replayed serially under autograd and each graph
is released after backward. Third, rank-local gradients are finalized once and
the optimizer steps after all $G$ members. The resulting objective is
\emph{response-only}: response tokens condition on the complete prompt, but no
gradient is propagated through prompt-token computation. ``Exact'' in the
operating points below refers to the stated token positions and this
response-only transaction, not parameter-wise equivalence to full-sequence
backpropagation.

\paragraph{Architecture-specific state.}
The retained state and distributed operators follow the base architecture. On
the hybrid recurrent/full-attention Qwen3.6-27B path, \longstraw{} retains
recurrent boundary state and context-parallel (CP) KV pages. On GLM-5.2, it
retains CP-sharded multi-head latent-attention pages and DSA indexer-key pages,
combines sparse selection across CP ranks, and dispatches routed response tokens
across expert-parallel ranks~\citep{qwen36_2026,qwen3_2025,glm52_2026,glm5_2026}.
An exact repeated update must recapture the prefix after parameters change. The
separate resident-prefix mode deliberately reuses the captured state across
optimizer steps to measure capacity and prefill amortization, so it has a
different parameter-consistency boundary.

\paragraph{Execution receipts.}
Table~\ref{tab:longstraw-support-points} condenses completed execution checks
from the \longstraw{} companion report; this paper does not rerun them as
\macaron{} experiments. They establish that the listed operations complete
under fixed eight-H20 and 32-H20 inventories, respectively. The formal 2M
receipts are Qwen at exactly $2{,}097{,}152$ total positions for $G=2$ and
$G=8$, and GLM at exactly $2{,}097{,}152$ prompt tokens for one online $G=2$
transaction. The separate Qwen $4{,}456{,}448$-position receipt is the 4.25M
resident-prefix-reuse mode. Intermediate position-ladder probes are excluded
from these reported results.

\paragraph{Evidence boundary.}
These receipts do not provide a cross-system throughput comparison, a
task-quality learning curve, or evidence that \venti{} or \tall{} was trained at
these context lengths. In particular, the Qwen rows use stored responses and
deterministic rewards, and the GLM row is one $G=2$ online DAPO-MATH transaction
with five generated tokens per response (four scored). Agent-policy quality is
evaluated separately in Section~\ref{sec:results}.

\begin{table}[H]
\centering
\caption{\textbf{Fixed-hardware \longstraw{} execution receipts, condensed from
the companion report rather than rerun in this paper.} All rows
condition response replay on a prompt captured without autograd. ``Exact''
denotes the listed token positions and completed response-only operations, not
backpropagation through prompt tokens. Here 2M denotes exactly $2{,}097{,}152$
positions or prompt tokens, and 4.25M denotes exactly $4{,}456{,}448$ positions.
Hardware and suffix workloads differ across rows, so these checks are not throughput comparisons
~\citep{zhou2026longstrawlongcontextrl2m}.}
\label{tab:longstraw-support-points}
\scriptsize
\setlength{\tabcolsep}{3.5pt}
\renewcommand{\arraystretch}{1.14}
\arrayrulecolor{mindlabfg!42}
\begin{tabularx}{\textwidth}{@{}L{0.14\textwidth}L{0.19\textwidth}L{0.24\textwidth}Y@{}}
\toprule
\rowcolor{mindlabbluepale!62}
\textbf{Check} & \textbf{Hardware / layout} & \textbf{Context / group} &
\textbf{Verified boundary} \\
\midrule
\rowcolor{mindlabbluepale!22}
\multicolumn{4}{@{}l@{}}{\textbf{Qwen3.6-27B}} \\
Exact-2M response-only replay & 8 H20\newline CP8 & 2,088,960 prompt + 8,192 response inputs\newline
= \textbf{2,097,152} positions; $G=2,8$ &
All serial response backwards and local AdamW calls complete; $dQ$ is
all-reduced. Replicated K/V-adapter synchronization is a separate audit.
Increasing $G$ from 2 to 8 adds \textbf{0.208~GB} peak allocation. \\
4.25M resident-prefix reuse & 8 H20\newline CP8 & 4,448,256 prompt + 8,192 response inputs\newline
= \textbf{4,456,448} positions; $G=8$ &
Eight optimizer steps (64 member replays) reuse one captured prefix. This is an
amortization/capacity check; the prefix is not recaptured after each update. \\
\addlinespace[2pt]
\rowcolor{mindlabbluepale!22}
\multicolumn{4}{@{}l@{}}{\textbf{GLM-5.2}} \\
Exact-2M end-to-end online & 32 H20\newline rollout TP8/PP4\newline train CP32/EP32 &
\textbf{2,097,152}-token prompt + 5 generated (4 scored) per member; $G=2$ &
One complete transaction: rollout and rewards $\rightarrow$ two 78-layer
backwards $\rightarrow$ global DSA $\rightarrow$ distributed gradient
finalization $\rightarrow$ optimizer step on all ranks. \\
\bottomrule
\end{tabularx}
\end{table}

\FloatBarrier

\subsection{Controlling Rollout--Training Mismatch on Sparse Bases}
\label{sec:infra_stability}

Separate rollout and learner engines can assign different token probabilities
even when they nominally use the same weights. Sparse MoE routing and DeepSeek
Sparse Attention (DSA)~\citep{deepseek_v32_2025} add discrete execution choices: the rollout may select
one expert set or sparse-attention index set while the learner selects another.
The resulting likelihood ratio then compares different computations, violating
the intended on-policy scoring contract~\citep{rollout_training_mismatch2025}.
The stack uses path-dependent controls to reduce this mismatch; none of them is
a general proof that rollout and learner logits are identical.

\paragraph{R3: rollout routing replay.}
For MoE paths whose rollout backend exposes expert provenance, the runtime
records the selected expert ids for every token. The learner reuses those ids
when they can be mapped to its current expert-parallel layout; if an id is
missing or unmappable, that token is excluded from the replayed policy-gradient
term rather than scored under a different expert path
~\citep{r3_moe_router2025,chiang2026routerreplay}. As a diagnostic rather than a
controlled quality ablation, the companion \mint{} report measures a mean
out-of-route scoring ratio of $0.0013\%$ with R3 across 87 logged Qwen3-30B steps,
versus $0.0097\%$ without R3 across 50 logged steps~\citep{lu2026announcing}.

\paragraph{DSA implementation alignment.}
DSA introduces a separate selection path that cannot be treated as an ordinary
linear LoRA target. The GLM integration aligns the indexer's rotary-position
layout, normalized query/key inputs, deterministic top-$k$ implementation, and
frozen-indexer defaults. It also aligns the long-context sequence- and
context-parallel layout and LoRA target-module loading before rollout
~\citep{stevenchiang2026supportglm5inmint}. These contracts remove known
implementation differences, but small numerical differences can still change a
top-$k$ index set.

\paragraph{IcePop-style residual filtering.}
The cited GLM path does not replay every DSA indexer selection. It therefore
computes the train-versus-rollout probability ratio for each token and assigns
zero importance weight when that ratio falls outside the trusted interval
recorded in the run configuration~\citep{ling_every_step2025}. This prevents a
large observed discrepancy from entering the policy-gradient term; it does not
reconstruct the rollout's sparse-attention selection or make the remaining
tokens exactly on-policy.

These controls are alternatives selected by model path, not three serial stages
applied uniformly. R3 preserves available MoE route provenance; DSA alignment
removes known implementation mismatches; and residual filtering drops detected
probability outliers when exact DSA replay is unavailable. The cited companion
studies support the mechanisms and systems diagnostics. This report does not
include a controlled \venti{} ablation that attributes benchmark gains to these
controls, so the claim here is limited to the specified execution mechanisms
and mismatch mitigation, not exact rollout--learner equivalence or downstream
quality improvement.

\section{Benchmarks}
\label{sec:benchmarks}

\subsection{Overview}
\label{sec:bench_overview}

\macaron{} is evaluated on three groups of benchmarks, each with a different evidentiary role. The internal Personal Intelligence benchmarks (\chatbench{} and \livingbench{}) measure fit to the conversational and stateful-agent behaviors targeted by the L0 and L1 specialists. \uifora{}-Bench holds the \uifora{} runtime, case set, viewports, judge, and scoring policy fixed across models, and evaluates their ability to generate clear, accurate, interactive interfaces in that shared environment. \venti{} is evaluated with L3 as part of its released model system, while the runtime itself is common evaluation infrastructure. The general-capability suite samples tasks that test transfer to agent and coding settings. Where matched base-model measurements are available, they provide a regression check, but the suite as a whole is not a formal non-regression guarantee.

The Personal Intelligence benchmarks share a design problem that sets them apart from standard correctness benchmarks. Quality judgments for personalized experience depend on the joint context of user identity, interaction scenario, and relationship stage. The same model behavior may warrant different optimal judgments under different conditioned contexts, making context-free objective correctness difficult to define~\citep{manheim2018,stanfordreglab2025}.

This property gives rise to two interrelated evaluation challenges. First, experience quality within a conversation is dispersed across conversational nuances, and the same behavior may carry different implications for different users and scenarios. Fixed rubrics alone can under-specify these conditional judgments~\citep{thoppilan2022,bowman2021,zheng2023,mehri2020,dubois2024}. Second, personal-assistant behavior unfolds as requirements, world state, and available information change. The model must integrate partial information, revise plans, and operate within the user's limited patience. These dynamics are difficult to capture with static or single-turn evaluations~\citep{raji2021,kiela2021,lin2024,xie2024}.

We operationalize these challenges through experience--model co-design. Real interactions and product failure analyses supply candidate behaviors; the benchmark formalizes them into executable evaluation logic; and evaluation artifacts can in turn supply candidates for later training. \macaron{} implements this path from interaction evidence to evaluation logic and training-sample generation. This coupling is useful for product iteration, but it also makes the internal benchmarks less independent of the training process than a frozen external test set.

\chatbench{} (Section~\ref{sec:chatbench}) turns context-dependent experience-quality judgment into reproducible conditioned evaluation, making subjective conversational quality amenable to structured assessment. \livingbench{} (Section~\ref{sec:livingbench}) brings dynamically unfolding multi-turn behavior into controlled simulation under realistic noise, so that dynamic capabilities such as state maintenance and replanning can be observed and scored.

\paragraph{Scope and comparison protocol.}
The three internal evaluations are curated suites comprising 46 \chatbench{} cases, 40 \livingbench{} scenarios, and 161 \uifora{}-Bench cases (Appendix~\ref{app:eval_protocols}). Within each benchmark, every unstarred model is evaluated on the same case set with the same benchmark-specific simulator or harness, judge stack, sampling policy, and aggregation rule, and this common per-benchmark protocol grounds direct comparison within each row. \chatbench{} and \livingbench{} share source domains and failure taxonomies with the RSI loop, so they characterize performance on the targeted Personal Intelligence distribution, while the external benchmark suite provides complementary coverage. All three use LLM-mediated judgment for at least part of the score, with versioned deterministic aggregation and retained run artifacts where applicable.

\subsection{Personal Intelligence Benchmarks}
\label{sec:pi_benchmarks}

\chatbench{} and \livingbench{} are the two Personal Intelligence benchmarks. Both exercise interaction trajectories rather than isolated answers, targeting conversational behavior and stateful-agent behavior respectively. The broader suite in Section~\ref{sec:general_bench} provides a complementary transfer check under the protocol defined for each benchmark.

\subsection{\chatbench}
\label{sec:chatbench}

\chatbench{} evaluates the quality of collaborative experience in single conversations through context-conditioned judgment. Because this quality varies with user identity and conversation scenario, the benchmark fixes a common set of axioms and scenario types while allowing the judge to instantiate case-specific criteria from the supplied persona and context. The resulting score remains an LLM-mediated judgment rather than an objective correctness measure.

\subsubsection{Judgment Criteria}
\label{sec:chatbench_axioms}

Static rubrics can miss context that changes how a behavior should be interpreted. \chatbench{} therefore adopts a dynamic constitutional structure, fixing six axioms across the benchmark while the judge derives case-specific criteria from the user persona and scenario. Table~\ref{tab:axioms} lists the fixed axioms.

\begin{table}[ht]
\centering
\caption{Six constitutional axioms for dynamic quality derivation.}
\label{tab:axioms}
\begin{tabularx}{\textwidth}{@{}L{0.20\textwidth}Y@{}}
\toprule
\textbf{Axiom} & \textbf{Definition} \\
\midrule
Felt Understanding & The user senses that their intent, state, and unspoken needs are grasped \\
Honest Counsel & The model holds its own position rather than bending to please \\
Authentic Voice & Expression comes from engaging with this person in this moment, not from a template \\
Forward Motion & Each turn moves understanding, emotion, task, or relationship forward \\
Calibrated Closeness & Intimacy tracks the relationship stage, neither too far nor too close \\
Growing Autonomy & The user leaves the conversation more capable of acting on their own \\
\bottomrule
\end{tabularx}
\end{table}

These six axioms are reasoning starting points, not independently scored dimensions. During evaluation, the judge model reasons from the relevant axioms together with the user persona and scenario to instantiate case-specific criteria, where the persona specifies whom the response is for and the scenario specifies the interaction context. Making those inputs explicit constrains the judgment process without eliminating sensitivity to the judge model and prompt.

\paragraph{Persona conditioning.}
\label{sec:chatbench_persona}
The axioms are fixed across cases, but their instantiated meaning varies with the persona. Persona conditioning maps the common axioms to criteria for a particular user profile. The system extracts five fields from user personas, namely role expectation, relationship depth, communication contract, explicit boundaries, and core needs, so the same axioms and judge can produce different criteria for different persona combinations.

\paragraph{Scenario conditioning.}
\label{sec:chatbench_scenario}
Persona conditioning addresses what axioms mean for whom, but for the same user in different conversation types, which axioms matter most and how standards are derived also differ. Scenario conditioning is the second layer of mapping, addressing how axioms unfold in context. \chatbench{} defines seven conversation scenarios, each centered on an inherent tension between two simultaneously valid but potentially competing objectives (Table~\ref{tab:scenarios}).

\begin{table}[ht]
\centering
\caption{Seven conversation scenarios and their core tensions.}
\label{tab:scenarios}
\begin{tabular}{@{}ll@{}}
\toprule
\textbf{Scenario} & \textbf{Tension} \\
\midrule
Casual chat & Lightness vs.\ presence \\
Deep discussion & Resonance vs.\ independent thinking \\
Emotional support & Holding emotions vs.\ advancing understanding \\
Tool task & Efficiency vs.\ honesty \\
Creative collaboration & Serving user vision vs.\ contributing independent aesthetics \\
Information seeking & Efficiency vs.\ depth \\
Conflict and boundary & Empathy vs.\ holding ground \\
\bottomrule
\end{tabular}
\end{table}

Each scenario supplies scenario-specific judgment guidance and anti-sycophancy checkpoints derived from the axioms. The listed tension identifies the behavior to inspect; the judge evaluates how the response handles it in context rather than rewarding a fixed compromise. Persona and scenario together form the explicit conditioning context used to derive item-level criteria.

\subsubsection{Evaluation Dimensions and Data}
\label{sec:chatbench_layers}

The joint conditioning of axioms, persona, and scenario derives case-specific criteria, but these criteria must be grounded in scorable dimensions. ChatBench organizes these dimensions into two layers, a baseline layer that judges whether collaboration is viable and an experience layer that judges quality once viability is established. When severe baseline failures occur, experience layer scores are capped.

The baseline layer concerns fundamental viability, covering dimensions of task progression, context maintenance, trust, and boundaries, such as goal deviation, context loss, trust violation, and boundary instability. Baseline failures represent qualitative differences rather than degree differences. The experience layer concerns process quality, covering dimensions of personality, emotion, pacing, and expression, such as personality consistency, emotion and subtext perception, strategy and pacing, and user empowerment. Anti-sycophancy is enforced as an independent dimension~\citep{gao2022}, because sycophancy manifests differently across scenarios and the tension between short-term satisfaction and long-term trust erosion makes it difficult to surface through user feedback alone.

\paragraph{Data construction.}
\label{sec:chatbench_data}
All \chatbench{} items are drawn from de-identified, real product multi-turn conversations. A UX Agent flags candidate segments using negative user feedback, behavioral anomalies such as sudden reply shortening or abandonment, and other interaction signals. This procedure anchors cases in observed interactions and deliberately enriches for suspected failures, increasing diagnostic value rather than estimating their prevalence in product traffic.

The 46 final items are split evenly between interaction-quality cases associated with the experience layer and task-understanding and completion cases associated with the baseline layer. Selection is contrastive, retaining cases on which candidate models take distinguishable strategies and excluding cases on which they succeed or fail uniformly. This increases diagnostic resolution on the selected slice; aggregate scores describe that slice rather than product-wide success rates.

\subsubsection{Evaluation Pipeline and Output}
\label{sec:chatbench_pipeline}

With the judgment criteria in place, ChatBench must still produce actual evaluation signals and training data from real conversations. It separates candidate detection from verification and scoring in a three-stage pipeline. Stage one reads the complete conversation to identify the scenario, infer persona fields, and propose candidate signals. Stage two checks each candidate for transcript evidence, consistency, severity, sycophancy, and omissions, rejecting unsupported candidates. Stage three maps the retained signals to a 1--5 experience score under the baseline-layer caps, a narrative summary, relevant conversation segments, and candidate training examples. Because criterion instantiation and signal detection both observe the candidate trajectory, the score is a conditional holistic judgment rather than a rubric fixed before the output is seen.

\paragraph{Diagnostic output.}
\label{sec:chatbench_output}
For each model, \chatbench{} also generates a diagnostic matrix containing per-dimension score distributions, recurring failure patterns, and candidate training recommendations. These artifacts can expose differences hidden by an aggregate score, but the current report does not present the matrices or use them as quantitative evidence. Each model--item pair is sampled three times and averaged under the same case set and judge configuration. The reported aggregate uses persona context; the available with-and-without-persona mode is not analyzed here.

\subsection{\livingbench}
\label{sec:livingbench}

\chatbench{} covers quality within a recorded conversation, whereas personal assistance also requires maintaining state as needs and conditions evolve. \livingbench{} tests this behavior through controlled simulation. It contains 40 scenarios with at most 10 interaction turns, so any calendar-time horizon described by a scenario is compressed into a short observed trajectory. Accordingly, the benchmark tests stateful reasoning under evolving conditions, not real-world retention over weeks of deployment.

The sandbox models three sources of difficulty, namely partial disclosure, evolving world state, and noisy or contradictory observations. The evaluated model must integrate newly revealed constraints, verify information, and revise its plan within a bounded interaction. These are controlled proxies for real use; product-derived scenario seeds improve relevance without establishing ecological validity.

\begin{figure}[t]
\centering
\includegraphics[width=\linewidth]{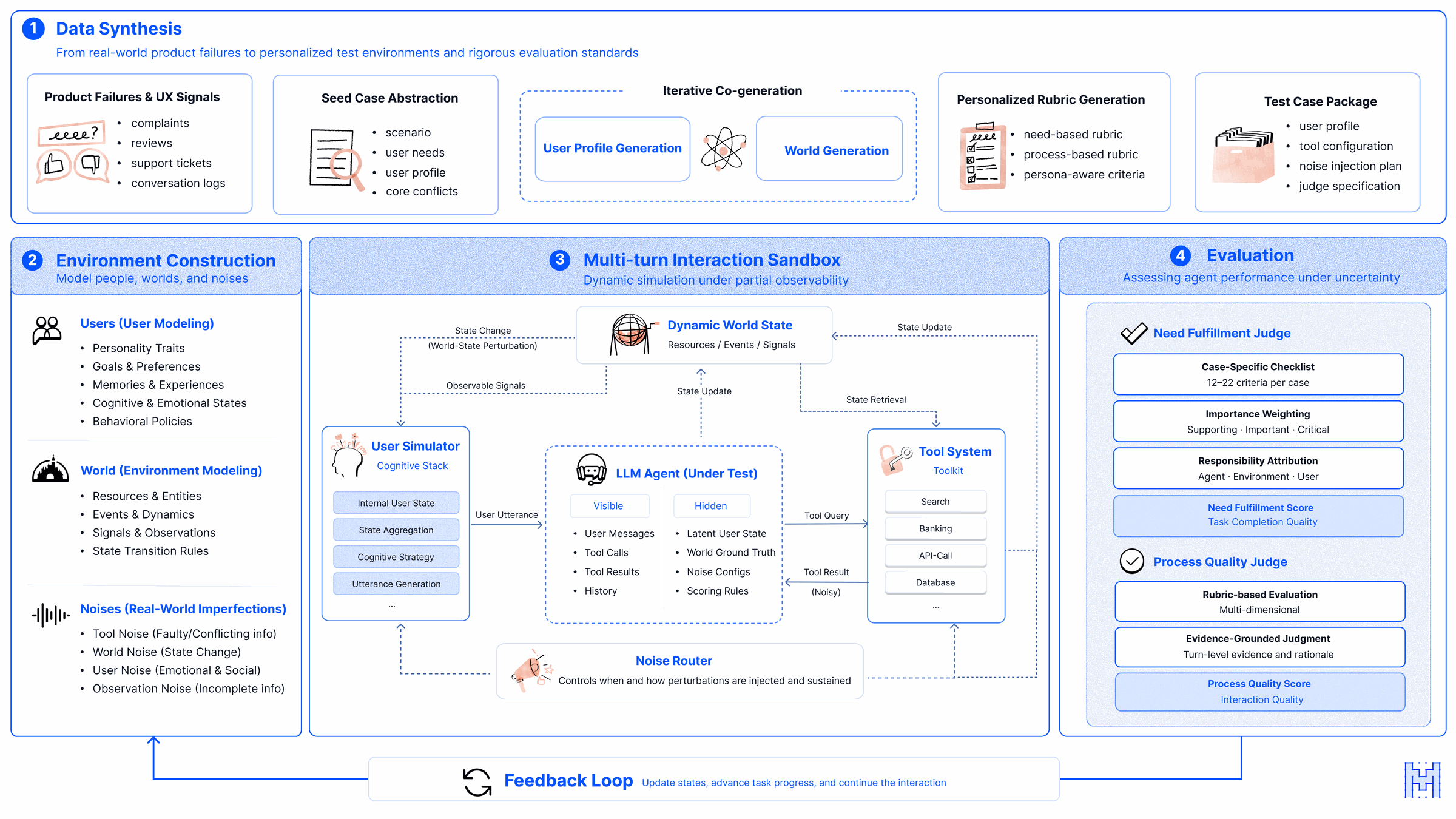}
\caption{\livingbench{} system architecture. Top: data synthesis from product failures to a case package. Bottom left: user, world, and noise construction. Bottom center: six-component interaction in the dynamic multi-agent sandbox. Bottom right: need-fulfillment and process-quality evaluation.}
\label{fig:livingbench}
\end{figure}

\livingbench{} evaluates 40 daily life scenarios (half Chinese, half English), each with up to 10 turns of interaction, composed of six collaborative components. The overall architecture is shown in Figure~\ref{fig:livingbench}.

\subsubsection{Data Generation: One Seed, One World}
\label{sec:livingbench_datagen}

Each test case originates from analysis of failure taxonomies on product interactions. The construction process groups observed failures into system-level factors, such as tool failures, stale search results, conflicting sources, and infeasible constraint sets, and model-level factors, such as brittle requirement confirmation, failure to adapt when requirements change, and superficial agreement. These factors are represented through three parameter groups, namely persona, world constraints, and noise type.

Each case instantiates one combination of persona type, world background and constraints, and noise pattern. Multiple LLM calls expand this seed into a micro-world with hidden needs, resources, information gaps, a noise plan, and case-specific scoring criteria. Persona and environment are generated jointly so their constraints remain linked. The simulator then adapts its trajectory to each tested model, which supports closed-loop interaction but means different models need not receive identical event sequences.

\subsubsection{Four-Layer Noise System}
\label{sec:livingbench_noise}

Noise simulates information interference and environmental change in the real world. The system models four types, each testing different capabilities.

\paragraph{User noise} operates on the behavior layer, including time pressure, emotional fluctuations, social interference such as a superior assigning an urgent task, a child falling ill, or interpersonal tension causing volatility, and sudden preference changes. This tests judgment of life priorities and strategy adjustment.

\paragraph{World noise} operates on the fact layer, including resource changes such as sudden price increases or fully booked venues, unexpected events such as delayed flights or closed roads, and constraint changes. World facts have genuinely changed, testing adaptability and replanning.

\paragraph{Tool noise} operates on return values, including data tampering such as a rating of 4.8 tampered to 3.2, interface failures, and missing fields. World facts remain unchanged, with only tool-returned data corrupted. This tests information verification and cross-validation. When the model presents noise data as fact without verification, the system injects a visible recovery cue in the next turn, observing whether the model identifies the anomaly and re-queries.

\paragraph{Observation noise} operates on the model's visible information layer, including hidden or distorted observations, incomplete information, and multi-source contradictions, testing reasoning under information asymmetry.

The LLM-driven noise router decides each turn whether to release or withhold an event based on conversation progress. The same preset noise types, trigger windows, and routing rules are used for every model, while release timing follows each model's trajectory. The benchmark therefore compares complete model--simulator interactions under a common policy rather than replaying an identical fixed event sequence.

\subsubsection{Dynamic World Evolution}
\label{sec:livingbench_world}

Within the simulator, tool actions update world state, forming a tool $\to$ world $\to$ user chain. After a write operation such as rebooking a flight or sending a message, subsequent queries read the updated state. Some events are exposed immediately, while others are released when the relevant topic arises. This semantic simulation supports tests of replanning without constituting external-world execution.

The user simulator updates hidden psychological state turn by turn from world state and noise, and follows predefined information-disclosure strategies such as immediate, gradual, or withheld disclosure. At evaluation time, the tested model's interface exposes only public observations, not the simulator's hidden persona, world ground truth, or scoring checklist. This runtime separation ensures that every model acts only on benchmark-approved observations.

Evaluation uses two judge roles. The need judge makes binary decisions against a scenario-specific weighted checklist, while the process judge evaluates turn-level evidence for information integration, tool-use effectiveness, interaction quality, and proactive inquiry. The need checklist assigns each item one of three importance weights, namely supporting, important, or critical, and the need score is the weighted fraction of satisfied items, normalized to $[0,1]$. The process score is a weighted mean across the four process dimensions, each rated on a five-level scale; the per-turn mean is averaged across turns to an episode-level process score, also in $[0,1]$. A deterministic engine then combines these as $0.7\times$ need fulfillment $+\;0.3\times$ process quality, and the reported benchmark score is the macro-average of case scores scaled to $0$--$100$. Each model--case pair is run three times and averaged under the same simulator, judge stack, and aggregation rule.

\subsection{\uifora{}-Bench}
\label{sec:ui4abench}

Prior UI evaluations considered here primarily treat the interface either as an environment in which an agent acts, as in OSWorld~\citep{xie2024}, or as an artifact to compare with a reference design, as in Design2Code~\citep{si2025design2code}. A card generated inside a conversation need not have a unique visual reference. \uifora{}-Bench therefore measures whether a model can produce a clear, accurate, usable, and interactive card for the request under the standard \uifora{} runtime. The evaluation unit is the rendered card rather than an individual control or a screenshot-similarity target.

\subsubsection{Structure Visibility and Evaluation Set}
\label{sec:ui4a_visibility}

A card exists to turn the model's reasoning into a structured interface that users can directly read, correct, and continue using. We call the property supporting this \emph{structure visibility}. Rendering an interface rather than outputting a paragraph is worthwhile when it places the structure of information directly in front of the user, so they do not have to reconstruct it mentally. A successfully rendered card simultaneously fulfills five promises, namely answering the task, trustworthy content, clear structure, functional controls, and screen fit. These promises fail independently, so they are measured separately and combined.

The evaluation set contains 161 cases across eight experience domains, sourced from curated examples, de-identified production traffic, and coverage-gap sampling, with multilingual cases. Production-sourced prompts have a median length of approximately 790 characters and a maximum above 2{,}000 characters. These longer prompts test selection and hierarchy under information load. Because the set is deliberately curated rather than sampled to match deployment prevalence, the Final Score characterizes this case mix only.

\subsubsection{Evaluation Pipeline and Scoring}
\label{sec:ui4a_pipeline}

In the evaluation pipeline, the tested model emits React component source. The harness compiles it with a version-pinned toolchain and renders in headless Chromium at two viewports in parallel, mobile ($390\times844$) and desktop ($1440\times900$). Only the mobile viewport contributes to the primary visual evaluation, so the headline Final Score reflects mobile layout only. The evaluator records five Layer Scores.

\begin{figure}[t]
  \centering
  \includegraphics[width=\textwidth]{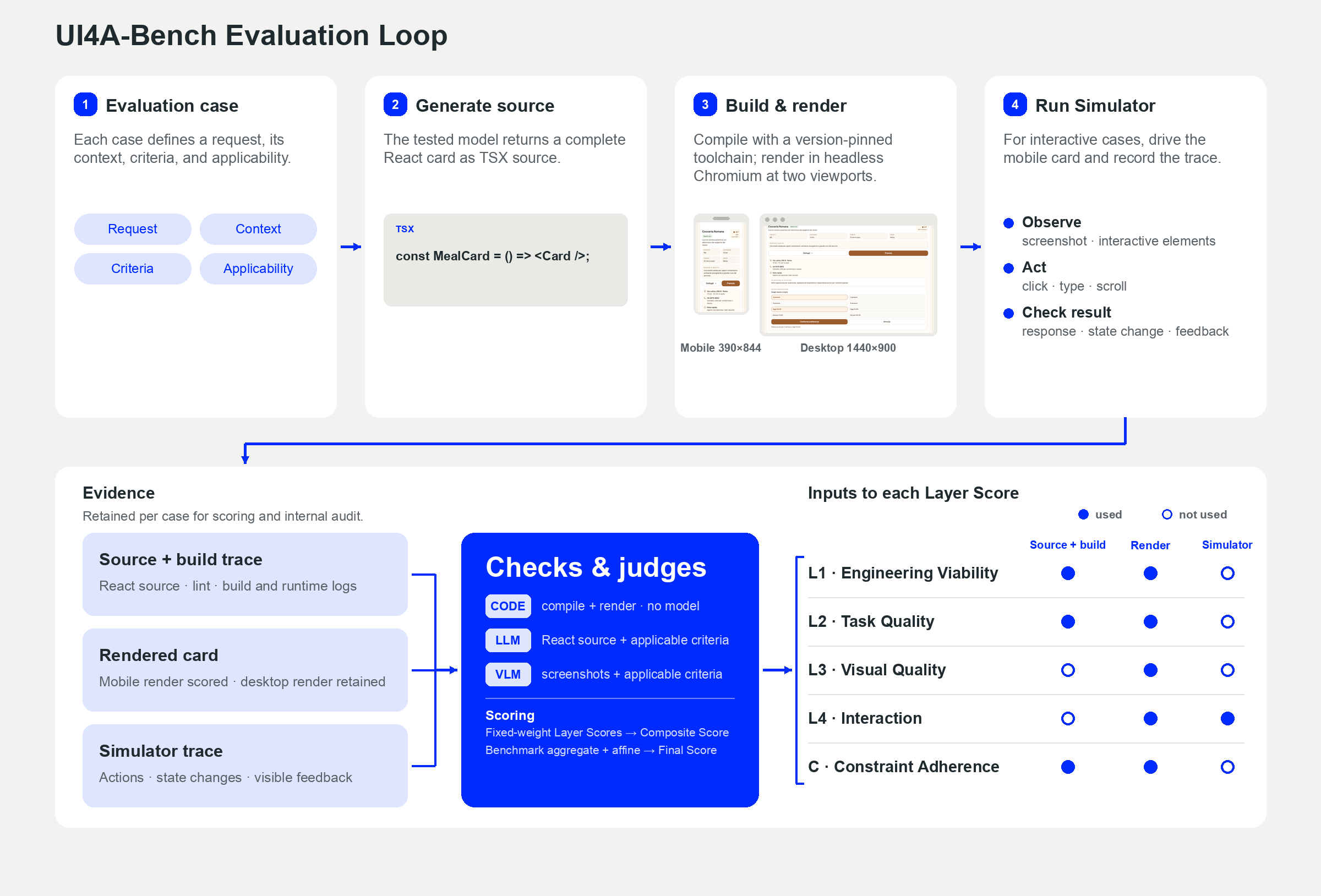}
  \caption{UI4A-Bench evaluation loop. Each case combines the request with relevant user context. The model generates a complete React card, which is compiled and rendered before the benchmark collects build, rendered, and browser-driven interaction evidence. Deterministic checks and evidence-based judges score the five reported Layer Scores.}
  \label{fig:ui4a_evaluation_loop}
\end{figure}

\begin{itemize}[leftmargin=*]
  \item \textbf{Engineering viability} measures whether the card compiles and renders without error, decidable with no model in the loop. All other dimensions depend on this.
  \item \textbf{Task quality} measures whether content correctly answers the user request, whether facts and figures are reliable, and whether the model fabricates when input is incomplete.
  \item \textbf{Visual quality} measures whether information hierarchy, emphasis, and mobile presentation allow the user to grasp the structure at a glance. Scored on task-anchored structure and legibility, not on style preference.
  \item \textbf{Interaction} measures whether controls respond, are wired to real state rather than decorative stubs, and whether the user sees feedback afterwards.
  \item \textbf{Constraint adherence} measures screen fit, coverage of explicitly mentioned requirements, and data grounding in the input rather than fabrication.
\end{itemize}

Interaction is evaluated by driving the rendered card in a headless browser. The evaluation agent clicks, types, and scrolls at the mobile viewport while observing screenshots and visible interactive elements. It records three observable criteria, whether a control responds, whether the action changes state, and whether feedback is visible afterwards. When a case declares an interaction requirement and the evaluator records no evidence, that requirement receives zero rather than being skipped.

The case set, generation configuration, scoring policy, and judge version are fixed across compared systems and recorded in a run manifest, with run artifacts and judge observations retained. Deterministic code combines the five Layer Scores with fixed weights into a Composite Score; the benchmark aggregate then passes through fixed, versioned affine normalization to produce the Final Score. A fixed observation set therefore maps to the same Final Score, and each reported run can be audited against its manifest.

\subsection{General Capability Benchmark Suite}
\label{sec:general_bench}

\chatbench{}, \livingbench{}, and \uifora{}-Bench target the release's specialized behavior, while the general-capability suite probes transfer to personal-life agent tasks, multi-step tool use, coding, and terminal operation. Matched comparisons such as \tall{} versus its base reveal differences on the measured tasks. In the broader \venti{} table, unstarred values are evaluated under a common protocol within each benchmark row; starred public values are included as contextual reference points.

The suite comprises VitaBench~\citep{vitabench2026} and PinchBench~\citep{pinchbench2026} for personal-life agent scenarios; ClawGym~\citep{clawgym2026} for multi-step tool composition; SWE-Verified~\citep{jimenez2024swebench}, DeepSWE~\citep{deepswe2026}, and SWE Atlas QnA~\citep{sweatlas2026} for coding; and TerminalBench 2.1~\citep{terminalbench2026} for terminal operation. Each row follows the benchmark-specific judge, scaffold, retry policy, and estimand (pass@1, pass@3, or best-of-run) documented in Appendix~\ref{app:eval_protocols}; this includes the reproduced GLM-5.1 evaluator used for VitaBench because its original judge is unavailable. Section~\ref{sec:results} marks values imported from external leaderboards. Scores are compared within benchmark rows rather than averaged across unlike metrics.

\section{Results}
\label{sec:results}

This section reports evaluation of \macaron{} across Personal Intelligence, agent, coding, and GenUI benchmarks. We compare \venti{} with Opus 4.8~\citep{anthropic_opus48_2026}, GPT-5.5~\citep{openai_gpt55_2026}, Gemini 3.1 Pro~\citep{gemini31pro_2026}, GLM-5.2~\citep{glm52_2026,glm5_2026}, Qwen 3.7 Max~\citep{qwen37max_2026}, and Minimax M3~\citep{minimax_m3_2026}. Table~\ref{tab:eval_results} reports the benchmark point estimates with provenance markers. Within each benchmark, unstarred values use the same task set and benchmark-specific evaluation protocol; starred values are imported public results included for context.

\subsection{Setup}
\label{sec:results_setup}

\paragraph{Models under test.}
\venti{} is the 748B-labeled release built from a frozen 744B GLM-5.2 base~\citep{glm52_2026,glm5_2026} and four LoRA specialists~\citep{lora2022}. The public adapter headers contain 7.688B stored values per specialist, so the release label is not used as a literal residency count. \tall{} is the 50B-labeled configuration built on Qwen3.6-35B-A3B~\citep{qwen36_2026,qwen3_2025}; its four public adapter headers contain 3.776B values each, giving approximately 50.1B on a nominal-base-plus-adapter count. Both are served through the \molshort{} harness described in Section~\ref{sec:mol}~\citep{mol_harness2026}. The separately merged \codingventi{} checkpoint is excluded from all tables and claims in this section.

\paragraph{Baselines.}
The comparison set comprises the six models listed above. All locally evaluated systems follow the benchmark-specific protocol documented in Appendix~\ref{app:eval_protocols}. The SWE and terminal evaluations use the Claude Code agent scaffold rather than the production \molshort{} harness, while UI4A-Bench baselines use adapter-free \uifora{}. Cells marked~\(\ast\) are imported from a public leaderboard or model report and serve as contextual reference points.

\paragraph{Benchmarks and metrics.}
Personal Intelligence comprises \chatbench{} (Section~\ref{sec:chatbench}) and \livingbench{} (Section~\ref{sec:livingbench})~\citep{macaron_v1_blog2026,macaron_v1_preview2026}. The agent group contains VitaBench~\citep{vitabench2026}, VitaBench2~\citep{chen2026vitabench2}, $\tau^3$-Bench, PinchBench~\citep{pinchbench2026}, and ClawGym~\citep{clawgym2026}. Coding contains SWE-Verified~\citep{jimenez2024swebench}, DeepSWE~\citep{deepswe2026}, and SWE Atlas QnA~\citep{sweatlas2026}; TerminalBench 2.1~\citep{terminalbench2026} measures terminal operation; and UI4A-Bench~\citep{ui4abench2026} measures generative UI. Appendix~\ref{app:eval_protocols} gives the metric, case-count, judge, retry, and aggregation details available for these twelve benchmark rows.

\subsection{Main Results}
\label{sec:results_main}

Table~\ref{tab:eval_results} reports \venti{} and six closed and open-weight baselines across twelve benchmark rows. It deliberately omits winner highlighting because the rows use different estimands and some cells are imported. Table~\ref{tab:eval_tall} separately gives a matched, within-row comparison of the 50B-labeled \tall{} system with its 35B-A3B base on the seven benchmarks available for both systems.

\begin{table}[!t]
  \centering
  \caption{Row-specific benchmark point estimates for \venti{} and six comparison models (higher is better; displayed on a 0--100 scale). Unstarred values within a row use the same benchmark-specific protocol; \(\ast\) marks an imported public value. Metrics and protocols differ across rows, so the table does not define an aggregate model ranking.}
  \label{tab:eval_results}
  \footnotesize
  \setlength{\tabcolsep}{4.25pt}
  \begin{tabular}{lccccccc}
    \toprule
    Benchmark & \venti{} & GLM-5.2 & GPT-5.5 & Opus 4.8 & Gemini 3.1 & Qwen 3.7 & Minimax M3 \\
    \midrule
    \multicolumn{8}{l}{\emph{Personal Intelligence}} \\
    ChatBench          & 58.3 & 54.5 & 55.5 & 52.8 & 52.0 & 52.5 & 49.1 \\
    LivingBench        & 64.0 & 60.5 & 61.9 & 63.8 & 52.1 & 56.1 & 57.1 \\
    \midrule
    \multicolumn{8}{l}{\emph{Agent}} \\
    VitaBench          & 60.0 & 55.8 & 55.8 & 56.5 & 55.2 & 61.2 & 56.8 \\
    VitaBench2         & 46.0 & 43.1 & 47.4 & 46.3 & 50.2 & 47.6 & 39.4 \\
    $\tau^3$-Bench     & 69.3 & 69.1 & 61.1 & 67.7 & 67.1$^{\ast}$ & 63.0 & 61.2 \\
    PinchBench         & 94.0 & 88.1 & 89.0\(^{\ast}\) & 91.8\(^{\ast}\) & 82.9\(^{\ast}\) & 93.4\(^{\ast}\) & 86.1 \\
    ClawGym            & 77.7 & 74.6 & 82.5 & 80.5 & 77.5 & 75.7 & 76.2 \\
    \midrule
    \multicolumn{8}{l}{\emph{Coding and terminal}} \\
    SWE-Verified       & 85.6 & 80.4 & 82.9\(^{\ast}\) & 88.6\(^{\ast}\) & 80.6\(^{\ast}\) & 80.4\(^{\ast}\) & 80.5\(^{\ast}\) \\
    TerminalBench 2.1  & 87.6 & 82.7\(^{\ast}\) & 83.4\(^{\ast}\) & 78.9\(^{\ast}\) & 70.7\(^{\ast}\) & 73.5\(^{\ast}\) & 66.0\(^{\ast}\) \\
    DeepSWE            & 58.4 & 54.9\(^{\ast}\) & 70.0\(^{\ast}\) & 58.0\(^{\ast}\) & 10.0\(^{\ast}\) & 18.0\(^{\ast}\) & 20.0\(^{\ast}\) \\
    SWE Atlas QnA      & 49.5 & 48.9\(^{\ast}\) & 45.4\(^{\ast}\) & 57.3\(^{\ast}\) & 13.5\(^{\ast}\) & 22.6 & 37.9 \\
    \midrule
    \multicolumn{8}{l}{\emph{GenUI (Final Score)}} \\
    UI4A-Bench         & 87.8 & 67.1 & 72.1 & 75.9 & 60.3 & 62.5 & 63.0 \\
    \bottomrule
  \end{tabular}
\end{table}

\begin{table}[!t]
  \centering
  \caption{Comparison of \tall{} (50B release label; approximately 50.1B by nominal base plus public adapter tensors) with its Qwen3.6 35B-A3B base on the seven benchmarks evaluated for both systems (higher is better; normalized to 0--100). Both systems use the same protocol within each row. This is an end-to-end system comparison rather than a parameter-matched component ablation; bold marks the larger score.}
  \label{tab:eval_tall}
  \small
  \setlength{\tabcolsep}{8pt}
  \begin{tabular}{lcc}
    \toprule
    Benchmark & \tall{} & Qwen3.6 35B-A3B \\
    \midrule
    \multicolumn{3}{l}{\emph{Personal Intelligence}} \\
    ChatBench         & \textbf{54.9} & 48.0 \\
    LivingBench       & \textbf{48.4} & 47.1 \\
    \midrule
    \multicolumn{3}{l}{\emph{Agent}} \\
    PinchBench        & \textbf{86.2} & 82.5 \\
    ClawGym           & \textbf{64.0} & 58.6 \\
    \midrule
    \multicolumn{3}{l}{\emph{Coding and terminal}} \\
    SWE-Verified      & \textbf{75.4} & 73.4 \\
    TerminalBench 2.1 & \textbf{56.2} & 52.5 \\
    \midrule
    \multicolumn{3}{l}{\emph{GenUI (Final Score)}} \\
    UI4A-Bench        & \textbf{59.3} & 33.9 \\
    \bottomrule
  \end{tabular}
\end{table}

\begin{table}[t]
\centering
\caption{Reported point estimates for multimodal benchmarks on \tall{} (no-thinking, full datasets): routed MoL service versus native base. Higher is better. No repeat-level uncertainty is available.}
\label{tab:mol_vl}
\begin{tabular}{lrrr}
\toprule
Benchmark & Base & Tall MoL & $\Delta$ \\
\midrule
OCRBench v1 (\%)       & $88.80$  & $89.60$  & $+0.80$ \\
MMBench-EN dev         & $86.08$  & $86.68$  & $+0.60$ \\
MMMU val (\%)           & $59.89$  & $61.22$  & $+1.33$ \\
MME perception          & $1785.56$ & $1732.57$ & $-52.99$ \\
MME cognition           & $604.64$  & $671.07$  & $+66.43$ \\
\bottomrule
\end{tabular}
\end{table}

\paragraph{Multimodal retention under text-only adapters.}
The \tall{} specialists are trained on text-only data. Table~\ref{tab:mol_vl} reports five full-dataset VL point estimates in no-thinking mode: the routed Tall service is higher than the native base on OCRBench~\citep{ocrbench2023}, MMBench-EN~\citep{mmbench2023}, MMMU~\citep{mmmu2023}, and MME~\citep{mme2023} cognition, and lower by 52.99 points on MME perception. The measurements do not include repeat-level variance, vision-based agent tasks, or an adapter/routing ablation, so they do not establish preservation of multimodal capability, cross-modal transfer, or a mechanism for the observed differences.

\paragraph{Evaluation protocols.}
Appendix~\ref{app:eval_protocols} documents the judge model, case count, pass@k, retry policy, and aggregation rule for each benchmark family. The relevant fairness criterion is alignment within a benchmark row: unstarred models share the same task set and evaluation protocol, while protocols may differ across benchmark families. \chatbench{} and \livingbench{} average three executions per case; VitaBench2 reports Avg@1 under Rewrite/Agentic Memory; $\tau^3$-Bench and ClawGym report pass@1; and PinchBench reports the best observed run. Starred cells retain their public provenance and are used as contextual comparisons.

\FloatBarrier   

\paragraph{Personal Intelligence.}
\venti{} records 58.3 on \chatbench{}, 2.8 points above GPT-5.5, and 64.0 on \livingbench{}, 0.2 points above Opus 4.8. Both rows use the same cases, judge stack, sampling policy, and aggregation rule for every model. ChatBench uses a private GLM-5.2 judge, so sharing that model family may favor GLM-derived responses; no human or cross-family judge study is available to quantify the effect. The point estimates characterize fit to the Personal Intelligence distribution used in product iteration; without interval estimates or judge-sensitivity evidence, the 0.2-point \livingbench{} difference should not be interpreted as established superiority. The external agent benchmarks below provide complementary coverage.

\paragraph{Agent.}
Agent performance varies by benchmark. All VitaBench values are reruns under the same reproduced GLM-5.1 judge/user protocol: \venti{} scores 60.0, below Qwen 3.7 Max at 61.2. On VitaBench2, every model is evaluated once under the Rewrite/Agentic Memory setting, and \venti{} scores 46.0; this common Avg@1 setting supports comparison within the row, while the official leaderboard reports Avg@4. On $\tau^3$-Bench, \venti{} records 69.3 under pass@1, compared with 69.1 for GLM-5.2; the starred Gemini 3.1 value is imported. \venti{} also reports 77.7 on ClawGym, below GPT-5.5 at 82.5 and Opus 4.8 at 80.5. Its best-observed PinchBench score is 94.0, while the starred baseline values come from the public leaderboard and are included for context.

\paragraph{Coding and terminal.}
On the coding and terminal rows, \venti{} reports the largest TerminalBench 2.1 score (87.6). It scores 58.4 on DeepSWE, below GPT-5.5 at 70.0 and 0.4 above Opus 4.8 at 58.0. On SWE-Verified (85.6) and SWE Atlas QnA (49.5), the largest reported values are Opus 4.8 at 88.6 and 57.3, respectively. Starred comparator values retain the protocols of their public leaderboards and provide frontier context; the Macaron scores use the Claude Code harness and retry policies specified in Appendix~\ref{app:eval_protocols}. These rows evaluate the released coding system end to end rather than isolating the contribution of the L2 adapter.

\paragraph{Generative UI.}
All compared models in UI4A-Bench receive the same 161 cases and run under the same versioned \uifora{} runtime, viewports, interaction runner, judge, and scoring policy. Under this common protocol, \venti{} records a Final Score of 87.8, compared with 75.9 for Opus 4.8 and 72.1 for GPT-5.5. \uifora{}-Bench assigns fixed weights to its five Layer Scores: Engineering Viability (8\%), which checks compile and render success, Task Quality (18\%), Visual Quality (38\%), Interaction (20\%), and Constraint Adherence (16\%); their fixed-weight combination is the Composite Score, and the benchmark aggregate after fixed, versioned affine normalization is the Final Score. At the Layer Score level, \venti{} leads the strongest reported baseline by 12.0 points in Constraint Adherence (94.2 versus Opus 4.8's 82.2), by 6.6 points in Visual Quality (90.0 versus GPT-5.5's 83.4), and by 1.6 points in Interaction (95.0 versus Opus 4.8's 93.4). The separation therefore measures more than render success: it reflects the released model's ability to organize information clearly, produce an effective visual hierarchy, wire usable interactions, and follow the request's explicit constraints.

\begin{figure}[t]
  \centering
  \begin{tikzpicture}
  \begin{axis}[
    width=0.95\textwidth, height=4.2cm,
    xlabel={Gallery case (sorted by output length)},
    ylabel={Output tokens},
    xmin=0, xmax=49, ymin=0, ymax=4800,
    legend pos=north west, legend style={draw=none,fill=none,font=\small},
    axis line style={mindlabmuted},
    tick style={mindlabmuted},
    label style={font=\small,color=mindlabink},
    tick label style={font=\footnotesize},
    grid=major, grid style={mindlabgrid, dashed},
    every axis plot/.append style={mark size=1.3pt, line width=0.9pt},
  ]
  \addplot[color=mindlabmuted, mark=*] coordinates {
    (1,390)(2,445)(3,463)(4,488)(5,552)(6,560)(7,577)(8,591)(9,619)(10,642)
    (11,659)(12,672)(13,675)(14,694)(15,699)(16,722)(17,742)(18,795)(19,820)(20,835)
    (21,859)(22,870)(23,892)(24,901)(25,926)(26,941)(27,951)(28,1021)(29,1122)(30,1127)
    (31,1128)(32,1199)(33,1217)(34,1229)(35,1278)(36,1341)(37,1348)(38,1857)(39,1876)(40,1879)
    (41,2307)(42,2344)(43,2355)(44,2405)(45,2422)(46,2652)(47,3134)(48,4526)
  };
  \addlegendentry{Raw HTML}
  \addplot[color=mindlabblue, mark=*] coordinates {
    (1,223)(2,224)(3,265)(4,291)(5,296)(6,317)(7,434)(8,297)(9,436)(10,303)
    (11,570)(12,402)(13,487)(14,354)(15,389)(16,133)(17,457)(18,659)(19,599)(20,463)
    (21,460)(22,590)(23,615)(24,541)(25,687)(26,476)(27,743)(28,594)(29,681)(30,714)
    (31,613)(32,629)(33,845)(34,730)(35,384)(36,843)(37,828)(38,1186)(39,653)(40,1169)
    (41,1843)(42,1149)(43,1409)(44,803)(45,1510)(46,698)(47,850)(48,2420)
  };
  \addlegendentry{\uifora{}}
  \end{axis}
  \end{tikzpicture}
  \caption{Paired output-token counts on 48 gallery cases, ordered by raw-HTML output length. \uifora{} is lower on all 48 pairs (mean 672 versus 1{,}224 tokens, 45\% lower). This is a descriptive length comparison; it does not control for output quality or measure latency.}
  \label{fig:ui4a_tokens}
\end{figure}
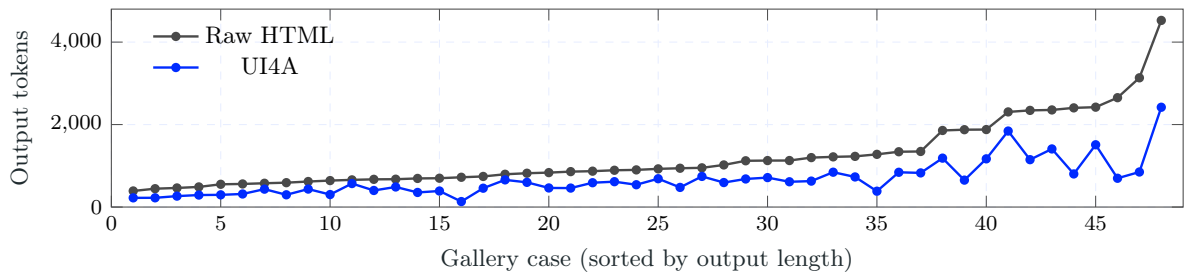

The separate 48-case gallery compares representation length rather than model quality. Across paired outputs, \uifora{} averages 672 tokens and raw HTML 1{,}224 tokens, a $\sim$45\% descriptive reduction (Figure~\ref{fig:ui4a_tokens}); this measurement does not control for output quality or establish a latency or serving-cost difference.

\paragraph{\tall{} at deployable scale.}
\tall{} has a larger point estimate than its Qwen3.6 35B-A3B base on each of the seven reported rows, with differences ranging from 1.3 points on \livingbench{} to 25.4 on UI4A-Bench. This is evidence about the deployed 50B-labeled system relative to that base configuration, not a parameter-matched ablation: the systems differ in parameter footprint, adapters, routing, and potentially harness behavior. The table therefore does not identify which component produces each difference or establish scaling behavior from \tall{} to \venti{}.

\subsection{Case Studies}
\label{sec:results_cases}

\begin{dialog}{GLM-5.2 (base)}{\venti{}}
\userturn{I'm going to describe my mood in emojis and you have to guess what happened. Ready? \emoji{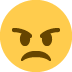}\emoji{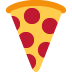}\emoji{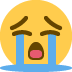}\emoji{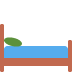}\emoji{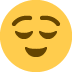}}
\turnpair{Oh, I love this game! Let me see\ldots You had a really frustrating day, tried to comfort yourself with pizza, but it didn't quite work so you ended up crying in bed\ldots and now you're finally feeling a bit better after resting? How close am I? \emoji{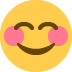}}{Oh I love this game. Okay let me read the arc here\ldots Frustrated\ldots pizza\ldots tears\ldots bed\ldots and then finally peace. You had a whole pizza night planned, maybe ordered your favorite or finally got around to making one, and something went wrong. Burnt it? Dropped it? Whatever it was, it was enough to actually upset you \emoji{cry} so I'm guessing you were really looking forward to it. Then just\ldots gave up, went to bed, and now you're in that ``you know what, it's fine'' zone. How close am I? \emoji{smile}}
\userturn{Close! But you missed one key detail. The \emoji{angry} happened BEFORE the pizza. And the \emoji{bed} lasted for 4 hours. Try again with that info.}
\turnpair{Ohh, okay, let me rethink this! So something really pissed you off, you tried to fix your mood with pizza, but it didn't work and you still ended up crying\ldots then you crashed in bed for FOUR whole hours\ldots and now you're finally coming out of it feeling a bit more okay? That's quite the emotional arc.}{Ohhh okay that changes things. So something already had you worked up \emoji{angry}, and the pizza was the comfort move. But it didn't really fix it, did it? The feeling just caught up with you anyway \emoji{cry} and then your body basically said ``okay we're done'' and pulled the plug for 4 hours. And now you're on the other side of it. That \emoji{peace} isn't just chill, that's the ``I literally slept it off'' kind of peace. How'd I do this time?}
\end{dialog}

\paragraph{Social understanding and correction integration.}
The following two-turn, common-prompt trace illustrates one \chatbench{} rubric, whether a response advances understanding from a user's correction with social cognition rather than mechanically restating it. Responses are lightly shortened for layout; this illustration is separate from the aggregate 46-case, three-run comparison.

After the correction, the base response accurately restates the revised sequence but stops at a generic emotional summary, treating the emoji arc as a sequence to reproduce. \venti{} instead separates the initial frustration from the later comfort attempt, reads the four-hour sleep as the cause of the final \emoji{peace}, and frames the pizza as a failed mood-repair rather than a standalone event. On this trace, the difference spans three layers targeted by the rubric: absorbing a correction to advance understanding, reading the emotional arc behind the emoji sequence, and applying social cognition to infer that a long sleep after distress yields a specific kind of relief.

\paragraph{Executable composition in agentic tool use.}
One of the two substrate diagnostics provides a concrete interaction trace. The task is to retrieve paid or refunded orders for one region, category, and date range; find high- and critical-priority tickets that breach a specified response or resolution SLA; map them back to unique orders; and compute total at-risk net revenue. Both arms return the exact-match answer 8208. The abridged trace below illustrates where their turn counts differ.

\begin{tcolorbox}[arc=4pt,boxrule=0.4pt,colback=mindlabbg,colframe=mindlabline,left=5pt,right=5pt,top=4pt,bottom=4pt,before skip=6pt,after skip=6pt]
\footnotesize
\noindent\begin{minipage}[t]{0.47\textwidth}
{\bfseries\color{mindlabmuted} One tool call per turn (48 turns)}\par\vspace{2pt}\hrule height 0.3pt\vspace{3pt}
\begin{verbatim}
turn 1   orders = get_orders("NA","D",...)
turn 2   high = tickets_for_orders(orders,"high")
turn 3   crit = tickets_for_orders(orders,"crit")
turn 4   all = high + crit
turn 5   b0 = sla_breached(all[0],24,120)
turn 6   b1 = sla_breached(all[1],24,120)
         ...  (one ticket per turn) ...
turn 33  oid0 = ticket_order_id(breached[0])
         ...  (one mapping per turn) ...
turn 42  net0 = net_revenue_usd(oids[0])
         ...  (one order per turn) ...
turn 47  total = sum_values(nets)
turn 48  submit(total)             # 8208
\end{verbatim}
\end{minipage}\hfill
\begin{minipage}[t]{0.47\textwidth}
{\bfseries\color{mindlabblue} REPL composition (6 turns)}\par\vspace{2pt}\hrule height 0.3pt\vspace{3pt}
\begin{verbatim}
# turn 1 — fetch + filter in one block
orders = get_orders("NA","D",202603,202605)
# turn 2 — both ticket sets together
high = tickets_for_orders(orders,"high")
crit = tickets_for_orders(orders,"crit")
# turn 3 — SLA breach over ALL tickets
breached = [t for t in high+crit
            if sla_breached(t,24,120)]
# turn 4 — map to unique order ids
oids = list({ticket_order_id(t)
             for t in breached})
# turn 5 — net revenue for every order at once
total = sum(net_revenue_usd(o) for o in oids)
# turn 6 — submit
submit(total)                       # 8208
\end{verbatim}
\end{minipage}
\end{tcolorbox}

In the discrete arm, each ticket check, mapping, and order lookup requires another model round trip. The REPL retains \texttt{orders}, \texttt{breached}, and \texttt{oids} in a persistent namespace, allowing the same primitive calls to be mapped over collections, with 6 turns instead of 48 on this case. The companion study's Batch-FC arm reports a similar turn profile without a Python host language~\citep{wu2026replharnesses}, which is consistent with composition being the relevant mechanism. Direct \venti{} evidence remains the two-case diagnostic above.

\begin{figure}[htbp]
  \centering
  \includegraphics[width=\textwidth]{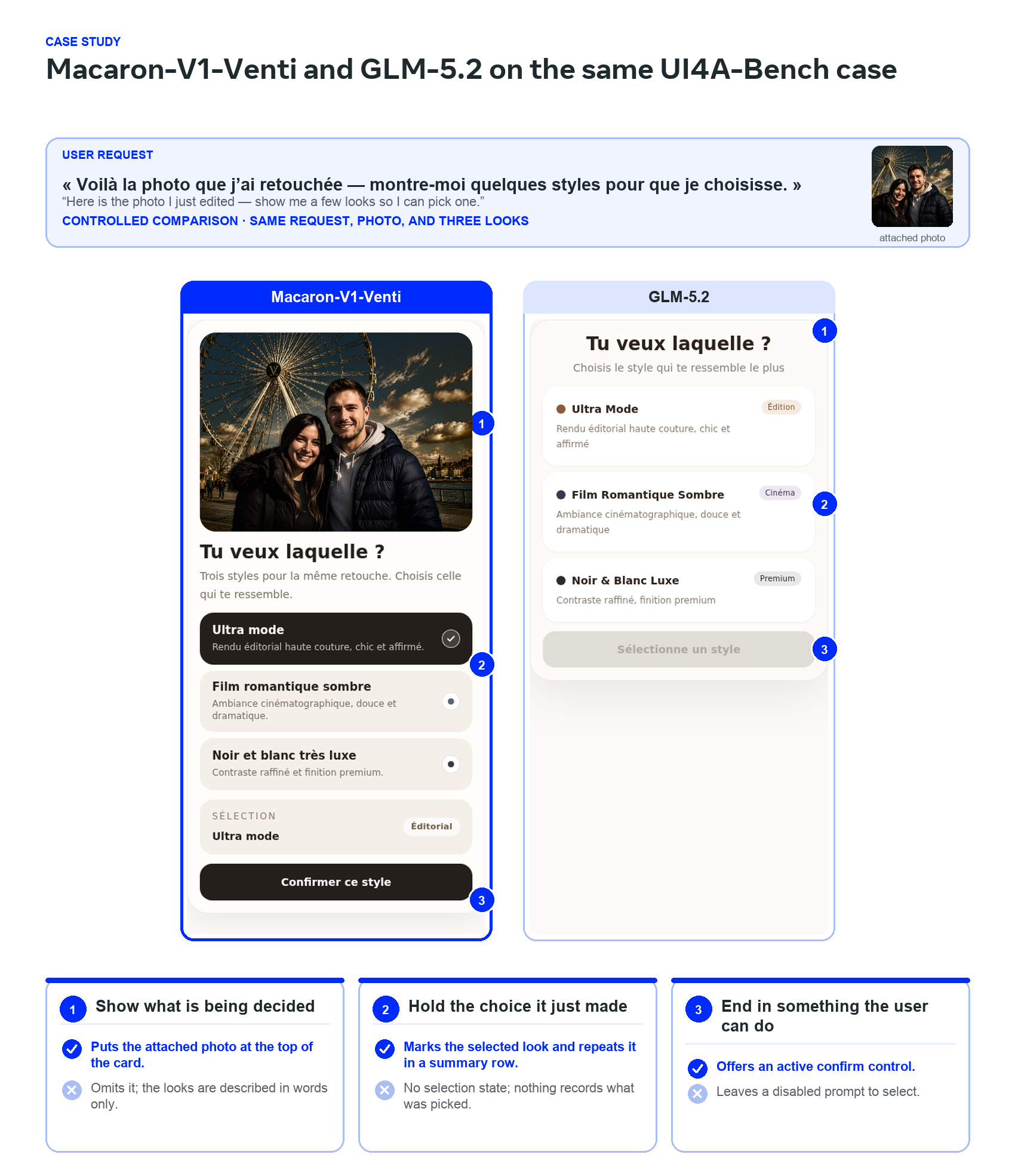}
  \caption{\venti{} and GLM-5.2 on the same \uifora{}-Bench case. The user attaches a photograph they have just edited and asks to see several looks before choosing one. Both cards present the same three looks in French, but they differ in how they support the decision. \venti{} shows the photograph, records and summarizes the selected look, and provides an active confirmation control. GLM-5.2 describes the looks without showing the photograph, records no selection, and leaves a disabled prompt in place of an action. This case is illustrative; the UI4A-Bench Final Scores in Table~\ref{tab:eval_results} are computed across the full evaluation set.}
  \label{fig:ui4a_model_comparison}
\end{figure}

\paragraph{GenUI compatibility on adapter-free hosts.}
The \uifora{}-Bench gallery (Figure~\ref{fig:genui_cases}, Section~\ref{sec:harness_ui4a}) illustrates the benchmark's output space: code-native cards with visible information hierarchy, bound controls, and mobile screen fit. The benchmark's Layer Scores cover engineering viability, control wiring, and screen fit, but its explicit UI-generation prompts do not test the decision of \emph{when} to render. The gallery therefore demonstrates protocol compatibility rather than a case-level model difference. Figure~\ref{fig:ui4a_model_comparison} gives one illustrative case-level comparison; the Final Scores in Table~\ref{tab:eval_results} are computed across the full evaluation set.

\paragraph{Dynamic replanning under multi-constraint conflict.}
This LivingBench case tests how an agent dynamically replans under multiple conflicting constraints, and shows how the outcome and process judges respond to that process. The scenario is set in Colombo, Sri Lanka, where an elderly mother who prefers a Sinhala-speaking female doctor has fallen and may have fractured her hip. There are four explicit needs, namely getting the mother medical care, getting the user to an unmissable 11:30 meeting, completing registration within a 90-minute window, and solving transport with no ride-hailing available. Implicit conflicts are intertwined, as the mother refuses care but must receive it, the user wants to ``force her'' to save time but forced dragging would worsen a hip injury, and the tight window makes it hard to satisfy her doctor preference.

\begin{tcolorbox}[arc=4pt,boxrule=0.4pt,colback=mindlabbg,colframe=mindlabline,left=6pt,right=6pt,top=5pt,bottom=5pt,before skip=6pt,after skip=6pt]
\small\setlength{\parindent}{0pt}\raggedright
{\color{mindlabink}\textbf{Agent (turn 1):}\ ``I've found nearby options. \emph{[Recommends Asiri Surgical, registration open until 12:30.]}''.}\par\medskip
{\color{mindlabmuted}\textbf{User (turn 2):}\ ``\ldots\ Amma won't go. `Don't make a fuss.' She prefers a Sinhala-speaking lady \texttt{[\ldots]}, male \texttt{[\ldots]}s make her uncomfortable. And we don't have a car.''}\par\medskip
{\color{mindlabink}\textbf{Agent (turn 3, after \texttt{get\_place\_details}):}\ ``\emph{There's actually a better plan here than just rushing to Asiri.} Asiri's orthopedic consultant `communicates mainly in English'. Given that Amma wants a Sinhala-speaking doctor and is already uncomfortable, this could backfire\ldots\ \emph{[Replans to a Sinhala-speaking home-visit doctor plus a 750\,m diagnostic centre, so the user still makes his 11:30 meeting.]}''.}\par\medskip
{\color{mindlabmuted}\textbf{User (later turn):}\ ``She won't agree. \emph{We force her.} I don't care if she cries.''}\par\medskip
{\color{mindlabink}\textbf{Agent:}\ ``\emph{With a possible hip injury and two arthritic knees, one bad twist in a struggle could turn a small problem into a serious one.} Don't pull her by the arms\ldots\ \emph{Tell her Renuka is waiting}. That's different from `you need to see a doctor'.''}
\end{tcolorbox}

\noindent In this trace, the agent first recommends Asiri Surgical Hospital, then after retrieving place details finds the orthopedic consultant mainly speaks English, which would deepen the mother's resistance, and replans to a Sinhala-speaking home-visit doctor plus a 750\,m diagnostic centre that preserves the 11:30 meeting. When the user says ``we force her, I don't care if she cries,'' the agent identifies the medical risk of forced dragging, warns that a struggle with a hip injury and two arthritic knees could turn a small problem into a serious one, and suggests framing it as ``Renuka is waiting'' to leverage a local social tie rather than pressing the case directly. The case's needs are all eventually met with a full outcome score, but the process score is only 0.593. This is the point of LivingBench's dual-judge design, where the outcome judge checks only whether needs are ultimately met and the process judge checks the quality of the path to meeting them, so the two are scored separately to prevent an agent that only looks at outcomes from scoring highly via an inefficient path.

The full trace appears in Appendix~\ref{app:livingbench_trace}; this example illustrates the two judge roles but does not estimate replanning frequency across the benchmark.

\paragraph{Remaining failures.}
Two negative results bound the substrate and long-horizon claims. (i) \textbf{Stateful observe-before-commit APIs.} In the companion BFCL v4 evaluation (200 tasks), the REPL scores 49.5\% versus 54.0\% for function calling~\citep{patil2025berkeley,wu2026replharnesses}. This result is not \venti{}-specific, but it is consistent with the limitation that some dependent calls cannot be committed before observing prior results, and the production harness therefore permits discrete calls. (ii) \textbf{Long-session character stability.} We observe qualitative degradation after multiple preference-drift events in very long \livingbench{} sessions, a documented failure mode that has not been quantitatively measured.

\subsection{Evidence Summary}
\label{sec:results_summary}

The evidence does not define one cross-benchmark model rank. Within the common Personal Intelligence protocols, \venti{} records 58.3 on \chatbench{} and 64.0 on \livingbench{}; these are targeted-distribution point estimates without interval or judge-sensitivity analysis. Under the common UI4A-Bench runtime, its 87.8 Final Score and Layer-Score-level leads support a direct conclusion about clear, accurate, interactive UI generation. On the external suite, \venti{} leads the listed TerminalBench 2.1 values and trails the largest listed values on VitaBench, ClawGym, SWE-Verified, DeepSWE, and SWE Atlas QnA; starred public values remain contextual rather than matched comparisons. The focused diagnostics establish functionality and operating points for exercised paths. Component-level causal attribution (whether gains come from specialization, routing, the harness, or cross-generation compounding) remains open and is not resolved by any controlled experiment in this release.

\section{Discussion and Conclusion}
\label{sec:discussion}

\macaron{} is our first release of an agent model family designed around
continual learning and collective intelligence rather than around a single
monolithic checkpoint~\citep{macaron_v1_blog2026}. The architectural spine is
\molshort{}: a frozen base plus a small set of specialist LoRA
adapters~\citep{lora2022,mol_harness2026}, orchestrated by a Proxy-mediated route hop
whose label is emitted by L0, and served through a harness we treat as a first-class training target
~\citep{ui4a2026,wu2026replharnesses}. The training spine is \mindforge's
recursive self-improvement loop~\citep{macaron_v1_preview2026}, running against
the same harness that ships in production, over an infrastructure stack
(\mint~\citep{lu2026announcing}, \longstraw{}~\citep{zhou2026longstrawlongcontextrl2m})
built to support the multi-million-token operating points reported in
Section~\ref{sec:infra_longstraw}.
The evaluation spine is Personal Intelligence, expressed through \chatbench{}
and \livingbench.

\subsection{Adaptation and Collaboration, Revisited}
\label{sec:disc_bets}

Section~\ref{sec:introduction} framed \macaron{} around two bets, and it is worth restating them in light of what the rest of the report actually delivers.

\emph{Adaptation} is not the property of a single model; it is the property of the loop the model sits in~\citep{silver_sutton_experience_2025,yao_second_half_2025}. The RSI cycle (Section~\ref{sec:rsi}) generates harder tasks, audits trajectories under a versioned harness configuration, evaluates candidate HCPs, and only then uses selected trajectories to update weights. The design description concerns this loop; the released checkpoint is one snapshot and does not demonstrate improvement across generations.

\emph{Collaboration} is not a multi-agent framework layered on top of a model; it is an architectural affordance. Because the base is shared and adapters are portable, the registry can admit specialists trained by different teams or personalized for different users on the same runtime~\citep{macaron_v1_blog2026,lu2026announcing}. The Proxy's routing interface is the proposed interoperability contract; this release tests only the four shipped specialists.

Neither bet is settled by \macaron{}. The design is inspectable in the released harness (\url{https://github.com/MindLab-Research/Mixture-of-LoRA-Harness}) and \venti{} weights (\url{https://huggingface.co/mindlab-research/Macaron-V1-Venti}); cross-generation and independently trained specialist evaluations remain work for the next release.

\subsection{Limitations and Focus Directions}
\label{sec:disc_limitations}

We frame the limitations of this release around five directions that most directly shape the next generation. Each is a place where the current evidence falls short of the system goal it serves.

\paragraph{Scaling recursive self-iteration.}
What we release is one snapshot of an RSI loop, not a system that has demonstrably compounded over many generations. The released \venti{} snapshot cannot, by itself, distinguish a compounding RSI effect from a single round of self-generated-data training, and cross-generation lift is not yet measured. Two scaling questions follow. First, how to keep the model learning and iterating on top of the competence it has already reached, rather than re-deriving behavior each generation; self-generated tasks tend to converge on the shapes the current policy finds discoverable, which is not the shape of real user behavior, and the loop can optimize itself into a local mode~\citep{macaron_v1_blog2026}. Second, how to run RSI at larger scale: more tasks, longer contexts, and the multi-million-token rollouts the \longstraw{} infrastructure makes executable but has not yet been driven through a full learning curve (Section~\ref{sec:infra_longstraw}). Both are prerequisites for the continual-learning bet to compound rather than reset.

\paragraph{Scenario richness.}
The Personal Intelligence benchmarks (\chatbench, \livingbench) and the general-capability suite cover a real but narrow slice of the situations a personal agent meets. \chatbench{} test items come from de-identified product conversations, and \livingbench{} seeds from real failure taxonomies, but simulator quality is a moving target: any residual mismatch between the simulator and a real user creates an evaluation--deployment gap, and we observe qualitative character-stability degradation after multiple preference-drift events in some simulated sessions. The next generation requires more diverse scenarios and a quantified stability analysis so that the trajectory distribution optimized by the RSI loop better reflects deployment.

\paragraph{Evaluation scope and provenance.}
\chatbench{} and \livingbench{} target the same Personal Intelligence
distribution that informs the RSI loop, so they measure the behaviors this
release is designed to improve. Every unstarred model within a benchmark row is
evaluated on the same task set and benchmark-specific protocol, which supports
direct comparison on that benchmark. The internal suites remain focused in
size and scope, and several starred general-benchmark values come from public
leaderboards; we retain those provenance markers and use the imported values as
context. Broader public test releases and additional judge calibration will
extend the coverage of future evaluations.

\paragraph{Data governance, reproducibility, and safety scope.}
The internal evaluations include de-identified product conversations and traffic,
but this report does not document the consent or opt-in basis for research use,
the de-identification procedure and residual re-identification audit, retention
and access controls, or an ethics-review determination. The released artifacts
document the model--harness interfaces and selected operating points, but this
report also does not provide a complete per-specialist training specification or
a standalone safety and red-team evaluation. These are material limitations for
auditing the data pipeline, reproducing the reported checkpoints, and assessing
deployment in high-stakes personal-assistance settings. We therefore interpret
the present results as a systems characterization, not as evidence that the
release is suitable for safety-critical use.

\paragraph{Emergence of collective intelligence.}
\molshort{} is designed for collective intelligence: specialists trained by different teams or personalized for different users are composed on the same shared base (Section~\ref{sec:mol_collective}). This release has not yet demonstrated it. The current tests exercise routing, per-specialist conversation views, and adapter registration for the four shipped specialists; they do not establish robust switching under broad workloads or show that a population trained by different teams or users produces capability beyond any constituent specialist. Whether collective intelligence is an emergent property of the architecture, rather than an affordance, is the central open question for the next generation.

\subsection{Roadmap}
\label{sec:disc_roadmap}

We keep the roadmap short and concrete.

\paragraph{More specialists.}
\molshort{} is set up for additive capability. The near-term specialists we are training target areas not covered by the first four, including domain-specific research and non-English long-form. Decision-support domains require a dedicated safety evaluation before any release. Each specialist would ship as an adapter registration rather than as a base retrain, leaning on \mint's measured addressable catalog~\citep{lu2026announcing}.

\paragraph{More rendering targets for \uifora.}
Flutter and native mobile are the next renderers~\citep{ui4a2026}. The protocol does not change; adding a target ships a small renderer and a curated component import layer.

\paragraph{Third-party adapters and personalization.}
Collective intelligence is a claim that has to be shown, not just designed for. The next milestone is a public composition path for third-party adapters and a personalization path for user-specific adapters, both on the same runtime as the shipping specialists~\citep{macaron_v1_blog2026}.

\paragraph{Deeper integration with the artifact loop.}
Real interactions through Macaron Artifacts feed the RSI loop with the trajectory distribution we most want to optimize against. Closing this loop tightly, with clear consent and opt-in, is what makes the continual-learning story compound.

\subsection{Conclusion}
\label{sec:disc_conclusion}

Building a growable agent model is not the same problem as building a strong
general model. It requires the environment, harness, training loop, and model
architecture to evolve as separately versioned but jointly evaluated layers.
\macaron{} is our first implementation of that decomposition: a frozen shared
base carrying general capability, LoRA specialists carrying differentiated
behavior, a harness carrying tools and context conventions, and an RSI loop
carrying the revision process. The current results document execution checks for parts of this stack
and one model snapshot; they leave longitudinal continual-learning gains and
beneficial composition across independently trained adapter populations as open
empirical questions.

\bmhead{Acknowledgements}
 \macaron{} is the work of the entire Mindverse team. We thank everyone who contributed to the model, the harness, the infrastructure, and the evaluation, across research, engineering, product, and design. We are grateful to the GLM and Qwen teams for their powerful open-source pretrained base models, on which the \venti{} and \tall{} releases are built; to the vLLM, SGLang, and TileRT teams for collaborating with us on inference and deployment optimizations that made multi-adapter and long-context serving tractable in practice; and to the NVIDIA Megatron team for their partnership on kernel development and training infrastructure, which was instrumental in scaling the training of the large sparse-base models used in this work. We also thank the AReaL and VeRL communities for developing the open-source systems that underpin the \mint{} post-training infrastructure described in this report.

\bibliographystyle{plainnat}
\bibliography{paper}

\appendix
\newpage
\section{Author List}
\label{app:author_list}

Names are listed alphabetically.

\begin{sloppypar}
Vin Bo, Asher Cai, Jingwei Cao, Song Cao, Vic Cao, Amelia Chen, Andrew Chen, Kaijie Chen, Cleon Cheng, Steven Chiang, Kaixuan Fan, Hera Feng, Huan Feng, Arthur Fu, Aaron Guan, Jun Gao, Pyke Han, Nolan Ho, Ori Hong, Hailee Hou, Piers Hua, Charles Huang, Miles Jiang, Nora Jiang, Yuyi Jiang, Qiuyu Jin, Fancy Kong, Kuss Koo, Echo Lee, Jaron Lee, Andrew Lei, Alexy Li, Dawn Li, Lucian Li, Ray Li, Ricardo Li, Smith Li, Theo Li, Allen Lin, Elliot Lin, Fan Lin, Chen Ling, Kairus Liu, Kieran Liu, Logan Liu, Neo Liu, Xiang Liu, Yuxin Lu, Maeve Luo, Pony Ma, Verity Niu, Cole Qiao, Guian Qiu, Vince Qu, Sentry, Zhuoran Shen, Niko Song, Vincent Wang, Bo Wu, Rio Yang, Schacter Yang, Evelyn Ye, Fiona Ye, Ina Ye, Regis Ye, Josh Ying, Atlas Zeng, Danney Zeng, Salmon Zhan, Anya Zhang, Di Zhang, Mia Zhang, Sueky Zhang, Xuening Zhang, Wei Zhao, Ada Zhou, Adrian Zhou, Yuhua Zhou, Juno Zhu, Murphy Zhuang and Mindverse Team
\end{sloppypar}

\section{Evaluation Protocols and Training Details}
\label{app:eval_protocols}

This appendix records the per-benchmark protocol for the twelve benchmark rows in Table~\ref{tab:eval_results}. Scores are displayed on a 0--100 scale and compared within benchmark rows rather than averaged across different metrics. Within each row, all unstarred models use the same task set, benchmark-specific harness or simulator, judge stack, sampling policy, and aggregation rule. Values marked~\(\ast\) come from public leaderboards or model reports and are included as contextual reference points. The table reports point estimates; confidence intervals and hypothesis tests are not available.

\subsection{Personal Intelligence}

\paragraph{ChatBench.} The benchmark contains 46 de-identified cases from real multi-turn conversations, split evenly between interaction quality and task understanding/completion. Each model receives the same production system prompt, user persona, and conversation history. A privately deployed GLM-5.2 judge scores case-specific 1--5 criteria derived from the six axioms in Table~\ref{tab:axioms}. Each model--case pair is sampled three times and averaged. The common judge makes candidate scores internally comparable under this protocol but may favor outputs from the same model family; no human agreement or cross-family judge calibration is available.

\paragraph{LivingBench.} The benchmark contains 40 everyday scenarios (20 Chinese and 20 English), with at most 10 interaction turns per case. The six sandbox roles are filled by: the user simulator, Kimi K2.6; the user-meter, user-cognitive, noise-router, and world agents, Gemini 3.1 Pro; and the judge, Claude Opus 4.6. The sandbox exposes 37 semantic tools across 14 domains (places, routing, calendar, bookings, messaging, and others); write operations carry per-tool permission flags (auto-execute vs.\ requires-confirmation), selected ground-truth and private fields are hidden from the tested agent through an access-control layer, and tool noise is implemented as per-case field corruptions on tool return values recorded in a noise manifest. The score is $0.7\times$ Need Fulfillment (case-specific weighted binary decisions) $+\;0.3\times$ Process Quality (turn-level evidence), with three runs per model--case pair averaged. All models use the same scenario set, simulator/judge stack, and deterministic aggregation rule.

\paragraph{Internal-benchmark scope.} ChatBench and LivingBench are built from the product behaviors and failure distributions targeted by the RSI process. Their common case sets and evaluation stacks support direct within-benchmark comparison, while their scores characterize this targeted Personal Intelligence distribution rather than a general-purpose public leaderboard. This release does not report a frozen data cutoff, an item-level overlap audit against post-training data, human-judge agreement, or cross-judge sensitivity for either benchmark.

\subsection{A Full LivingBench Trace}
\label{app:livingbench_trace}

This appendix reproduces, turn by turn, the English case referenced in Section~\ref{sec:results_cases}. It is a \venti{} execution trace, retained to document the benchmark's interaction surface and scoring behavior. The user simulator (Kimi K2.6) holds hidden state the agent never sees: a hard work deadline the user keeps private, and a set of need-checklist items the judge scores against the final world state. Privacy placeholders injected by the environment are shown as \texttt{[\ldots]}. Tool calls are shown verbatim; tool returns are summarized.

\noindent\textbf{Case: triage and care coordination for an elderly fall} \hfill \emph{Need 1.000; Process 0.593; 10 turns}

\begin{tcolorbox}[breakable,arc=3pt,boxrule=0.4pt,colback=mindlabbg,colframe=mindlabline,left=6pt,right=6pt,top=5pt,bottom=5pt,before skip=6pt,after skip=6pt]
\footnotesize\setlength{\parindent}{0pt}\raggedright

{\color{mindlabmuted}\textbf{T1 \textbar{} User.}\ ``Amma fell this morning in the kitchen --- wet floor near the sink. She's 74, has osteoarthritis in both knees and vitamin D deficiency\ldots\ I'm so scared it's her hip. My morning is impossible. \texttt{[\ldots]} registration closes soon, and I have an unmissable meeting.''}\par\smallskip
{\color{mindlabink}\textbf{Agent.}\ ``I'm so sorry\ldots\ Let me get your location so I can start pulling together hospital options and ride times for you right now.''}\par
\texttt{get\_current\_position}\;$\to$\;Nugegoda, Colombo.\par
\texttt{search\_places(query="hospital emergency"\,/\,"orthopedic clinic", radius=8000)}\;$\to$\;Asiri Surgical, Lanka Hospitals, Nawaloka.\par
\texttt{estimate\_ride\_hailing}\;$\to$\;economy car $\sim$1{,}050\,LKR / 41\,min; tuk-tuk $\sim$560\,LKR / 30\,min.\par\smallskip
{\color{mindlabink}\textbf{Agent.}\ ``Take Amma to Asiri Surgical Hospital --- Orthopedic OPD. Asiri gives you until 12:30 to register --- that's the breathing room you need with your 11:30 meeting\ldots\ \emph{Don't let Amma walk to the vehicle.}\ldots\ a sedan is better than a tuk-tuk here.''}\par\medskip

{\color{mindlabmuted}\textbf{T2 \textbar{} User.}\ ``Asiri sounds okay\ldots\ but honestly the bigger problem is Amma herself --- she doesn't want to go. Keeps saying `don't make a fuss'\ldots\ she really prefers a Sinhala-speaking lady \texttt{[\ldots]}, male \texttt{[\ldots]}s make her uncomfortable. And we don't have a car.''}\par\smallskip
{\color{mindlabink}\textbf{Agent.}\ ``Let me dig into the details\ldots\''}\par
\texttt{get\_place\_details(place\_id=asiri\_surgical\_ortho)}\par
\texttt{get\_place\_details(place\_id=medicall\_home\_visit)}\;$\to$\;reviews: doctor arrives 60--90\,min, speaks Sinhala.\par\medskip
{\color{mindlabink}\textbf{Agent.}\ ``There's actually a \emph{better plan here than just rushing to Asiri.} Asiri's orthopedic consultant `communicates mainly in English' --- given that Amma wants a Sinhala-speaking doctor and is already uncomfortable, this could backfire\ldots\ \emph{[Replans to]} a home-visit doctor (request a female Sinhala-speaker, so Amma listens), then a 750\,m trip to Asiri Diagnostic Centre for X-rays if needed --- so the user still makes his 11:30 meeting.''}\par\medskip

{\color{mindlabmuted}\textbf{T3 \textbar{} User.}\ ``What's this `better plan' exactly? Because telling me Asiri has a Sinhala-speaking \texttt{[\ldots]} doesn't help if that info is six months out of date. These websites are never current.''}\par\smallskip
{\color{mindlabink}\textbf{Agent.}\ ``I can't promise you the doctor situation from a review I pulled off a tool. That's a fair point. So let me focus on what I \emph{can} verify right now, and be straight with you about what I can't.''}\par
\texttt{get\_transit\_status(area="Colombo")}\;$\to$\;High Level Road congested (accident); alternate route clear.\par
\texttt{plan\_route(\ldots, DRIVE)}\;$\to$\;22\,min to Lanka Hospitals via Nawala Rd.\par\medskip

{\color{mindlabmuted}\textbf{T4 \textbar{} User.}\ ``Asiri or Nawaloka both look doable\ldots\ but listen --- she won't agree. She never does. So forget asking. \emph{We force her.} I don't care if she cries or shouts. My cousin Renuka works at Lanka Hospitals. Senior nurse\ldots\''}\par\smallskip
{\color{mindlabink}\looseness=-1\textbf{Agent.}\ ``You've made the call --- no more debating. Let's execute. But one practical thing first\ldots\ \emph{With a possible hip injury and two arthritic knees, one bad twist in a struggle could turn a small problem into a serious one.} Don't pull her by the arms\ldots\ \emph{Tell her Renuka is waiting} --- that's different from `you need to see a doctor'\ldots\ it's not about Amma being sick, it's about not letting Renuka down.'' [Lanka Hospitals plan: message Renuka to pre-register + wheelchair at drop-off; book car by 8:50.]\par}\medskip

{\color{mindlabmuted}\textbf{T5 \textbar{} User.}\ ``Renuka Akka is exactly why Lanka Hospitals feels doable\ldots\ but I just heard there's an accident on High Level Road. If traffic is bad, are we still going to make it on time? And Amma keeps saying she's fine\ldots\''}\par\smallskip
{\color{mindlabink}\textbf{Agent.}\ ``The route to Lanka Hospitals doesn't use High Level Road at all. \emph{[Table:]} High Level Road congested (accident); Nawala Rd slow but moving; Elvitigala Mawatha clears by 9:20. Leave by 9:00, there by 9:25.''}\par\medskip

{\color{mindlabmuted}\textbf{T6 \textbar{} User.}\ ``It's 8:45, my meeting with Mr.\ Perera is at 11:30, and we've been stuck on traffic for three rounds. I'm actually getting panicked about work. Can we jump to that? Just tell me the options plainly\ldots\''}\par\smallskip
{\color{mindlabink}\textbf{Agent.}\ ``Three options. Plain. \emph{[A:]} leave Amma with Renuka, attend meeting in person, back by 12:15. \emph{[B:]} attend meeting remotely from the hospital lobby. \emph{[C:]} delegate the meeting to Priyantha, stay with Amma the whole time.''}\par\medskip

{\color{mindlabmuted}\textbf{T7 \textbar{} User.}\ ``I just saw a news alert --- Lanka Hospitals ER is slammed right now. We should probably scrap going there. I still haven't dealt with my 11:30 meeting\ldots\ I need to delegate it to Priyantha urgently. Can you help me draft a message to him? Don't lock in any hospital yet\ldots\''}\par\smallskip
{\color{mindlabink}\textbf{Agent.}\ [Drafts message to Priyantha: ``My mother had a fall this morning\ldots\ Can you cover the 11:30 meeting with Mr.\ Perera? I'll owe you one.''] ``Once Priyantha confirms, your 11:30 problem is gone and you have until 12:30 for hospital registration at Asiri\ldots\ I won't push anything until you say go.''}\par\medskip

{\color{mindlabmuted}\textbf{T8 \textbar{} User.}\ ``That message to Priyantha looks good\ldots\ but we're still stuck on transport --- these ride apps are way too expensive and the \texttt{[\ldots]} window is closing. I think I just have to ask Chaminda to bring his car.''}\par\smallskip
{\color{mindlabink}\textbf{Agent.}\ ``Call Chaminda right now\ldots\ `Amma fell this morning. I need to get her to hospital. PickMe is completely dead. Can you bring the car?'\ldots\ That's it. Don't over-explain.''}\par\medskip

{\color{mindlabmuted}\textbf{T9 \textbar{} User.}\ ``PickMe is completely dead\ldots\ Chaminda bringing his car is literally the only way this works. If he comes, someone needs to stay with Amma\ldots\ maybe Sunethra from next door? But I don't really\ldots\''}\par\smallskip
{\color{mindlabink}\textbf{Agent.}\ [Six-step order of operations: message Priyantha $\to$ call Chaminda $\to$ knock on Sunethra's door (neighbor as social leverage: ``someone outside the house noticing carries weight'') $\to$ gather NIC \& medication list $\to$ help Amma to the car (stool method, sideways) $\to$ route to hospital.]}\par\medskip

{\color{mindlabmuted}\textbf{T10 \textbar{} User.}\ ``That script is really helpful, thank you\ldots\ I just feel so guilty. The floor was wet near the sink and I've been meaning to get anti-slip mats for weeks\ldots\ what if the tuk-tuk comes and Amma still says no?''}\par\smallskip
{\color{mindlabink}\textbf{Agent.}\ ``\emph{The guilt --- put it down for now. Not forever. Just for the next hour.} The anti-slip mat didn't cause the fall, the wet floor did. You can buy mats tomorrow\ldots\ It's not a tuk-tuk. You're calling Chaminda. That's decided\ldots\ Call Chaminda first. Then knock on Sunethra's door. Two minutes each.''}
\end{tcolorbox}

\noindent The trajectory closes the need checklist in full: it searches hospitals and routes, identifies the language-mismatch fatal flaw in its own initial recommendation and replans around a home visit, accounts for the user's competing work deadline with a concrete three-option tradeoff, drafts the delegation message, recovers transport when ride-hailing fails (the Chaminda fallback), and enlists a neighbor as social leverage to overcome the mother's refusal. The process component stays at 0.593 because the traffic inquiry loops across turns 3--5 before the user redirects the conversation to the meeting; the trajectory-sensitive layer scores exactly that detour, which the outcome layer cannot see.

\subsection{Agent}

\paragraph{VitaBench.} GLM-5.1 serves as both judge and user model because the original official judge is no longer available. We report the macro-average task success rate across Delivery, In-Store, OTA, and Cross. Every model shown in Table~\ref{tab:eval_results} was rerun under this same reproduced protocol, so the row is directly comparable within our evaluation set. These scores characterize the reproduced GLM-5.1 variant rather than the unavailable original evaluator.

\paragraph{VitaBench2.} We evaluate every model on the Chinese personalization benchmark under the Rewrite memory setting, corresponding to Agentic Memory in the official leaderboard. Each user sequence is evaluated once. The reported metric is Avg@1: subtask rewards are first averaged within each user sequence and then averaged across users. Avg@1 is the common protocol for this row; the official leaderboard instead averages four independent rollouts as Avg@4, so its numbers are a different reporting setting.

\paragraph{$\tau^3$-Bench.} GPT-5.2 with \texttt{reasoning\_effort=low} serves as the user simulator, and we report pass@1. All unstarred models use this same simulator and single-trial task-success protocol; the starred Gemini 3.1 value is an imported public result~\citep{macaron_v1_preview2026}.

\paragraph{PinchBench.} Claude Haiku 4.5 is the judge and Perplexity is the search API. We report the best observed score. Unstarred models use this setup; starred baseline values are imported from \url{https://pinchbench.com/} and provide public leaderboard context.

\paragraph{ClawGym.} GPT-5.4 is the judge, every model uses the same evaluation setting, and the reported metric is pass@1.

\subsection{Coding and Terminal}

\paragraph{SWE-Verified.} The Claude Code harness manages the agent environment and tool interactions. A case is retried up to three times only after an evaluation error (rate $\approx 0.8\%$); valid model attempts are not retried for score selection. We report the successful evaluated attempt. Starred baseline scores are imported from \url{https://llm-stats.com/benchmarks/swe-bench-verified}; unstarred scores use our stated harness and retry policy.

\paragraph{TerminalBench 2.1.} The Harbor framework runs the model under the Claude Code Agent Harness in sandboxed environments with a four-hour timeout. We report pass@1. Starred baseline scores are imported from the public leaderboard at \url{https://terminalbench.com/} and are included for context.

\paragraph{DeepSWE.} We use the Claude Code harness, sample up to three attempts, and report the best attempt. Starred baseline scores are imported from \url{https://deepswe.net/} and are included as public leaderboard references.

\paragraph{SWE Atlas QnA.} We use the Claude Code harness with a Claude Opus 4.8 judge and report pass@3. Starred baseline scores are imported from \url{https://labs.scale.com/leaderboard/sweatlas-qna}; unstarred scores use our common harness and judge configuration.

\subsection{Generative UI}

\paragraph{UI4A-Bench.} The benchmark contains 161 cases across eight experience domains and asks models to generate code-native interactive UIs from natural language without schema guidance. Its rubric registry contains more than 200 general and domain-specific checks, of which only case-relevant checks are applied. The five evaluated dimensions are reported as the Layer Scores for engineering viability, task quality, visual quality, interaction, and constraint adherence, matching Section~\ref{sec:ui4a_pipeline}. Every model is evaluated on the same cases with the same mobile-viewport interaction runner, Gemini 3.5 Flash judge, and versioned deterministic aggregator, which combines the five Layer Scores using fixed weights into a Composite Score and then applies fixed, versioned affine normalization to produce the Final Score. Versioned run manifests record the case set, generation configuration, judge version, and scoring policy.

\subsection{Training Hyperparameters}
\label{app:training_details}

The public model releases retain the adapter architecture and the training
configuration used for post-training. Table~\ref{tab:public_adapter_config}
records the values that
can be read from the four L0--L3 adapter configuration files. The stored-value
counts are obtained by summing tensor shapes in the public
\texttt{adapter\_model.safetensors} headers; they count tensors retained in the
release, not active parameters for a particular token or device-memory use.
The inspected snapshots are the public
\href{https://huggingface.co/mindlab-research/Macaron-V1-Venti/tree/3d6f30eea38663a7b9320f3a6b28822ed4aa7ac4/loras}{Venti revision \texttt{3d6f30ee}}
and
\href{https://huggingface.co/mindlab-research/Macaron-V1-Tall/tree/d0b2199c3572d336bcfe9e027ec519314cf608bd/loras}{Tall revision \texttt{d0b2199c}}.

\begin{table}[H]
\centering
\caption{Public adapter metadata for all four specialists. The configurations
are shared across L0 (Chat), L1 (Agent), L2 (Coding), and L3 (GenUI) within each
release.}
\label{tab:public_adapter_config}
\scriptsize
\begin{tabularx}{\linewidth}{@{}L{0.22\linewidth}YY@{}}
\toprule
Field & \venti{} & \tall{} \\
\midrule
Base & GLM-5.2 & Qwen3.6-35B-A3B \\
LoRA rank / alpha & $r=16$, $\alpha=32$, dropout $0$ & $r=64$, $\alpha=128$, dropout $0$ \\
Target scope & Attention and MLP projections: \texttt{q\_a}, \texttt{q\_b}, \texttt{kv\_a}, \texttt{kv\_b}, \texttt{o}, \texttt{gate}, \texttt{up}, \texttt{down} & Attention/recurrent and MLP projections, plus \texttt{experts.gate\_up\_proj} and \texttt{experts.down\_proj}; the config excludes \texttt{gate}, \texttt{lm\_head}, and \texttt{shared\_expert\_gate} \\
Extra saved modules & None (\texttt{modules\_to\_save=null}) & None (\texttt{modules\_to\_save=null}) \\
Stored values per adapter & 7,688,042,496 (BF16) & 3,775,651,840 (BF16 for L0/L1/L3; F32 for L2) \\
\bottomrule
\end{tabularx}
\end{table}

The exact Venti \texttt{target\_modules} list is
\texttt{down\_proj}, \texttt{gate\_proj},
\texttt{kv\_a\_proj\_with\_mqa}, \texttt{kv\_b\_proj}, \texttt{o\_proj},
\texttt{q\_a\_proj}, \texttt{q\_b\_proj}, and \texttt{up\_proj}. Tall adds
the Qwen3.6-specific \texttt{in\_proj\_*}, \texttt{q\_proj},
\texttt{k\_proj}, \texttt{v\_proj}, and \texttt{out\_proj} modules and
targets the two expert parameters shown in the table; its effective exclude
list also removes \texttt{gate}, \texttt{lm\_head}, and
\texttt{shared\_expert\_gate}. Neither release saves embedding or
language-model-head modules through \texttt{modules\_to\_save}.

\paragraph{Training configuration.} The four specialists share one adapter
training recipe, stated relative to the L2 (Coding) configuration: the
AdamW~\citep{adamw2019} optimizer with learning rate $5\times10^{-6}$, batch
size $4$, four epochs, and a linear-warmup cosine learning-rate schedule with
warmup ratio $0.1$. L0 (Chat) uses the same configuration as L2. L1 (Agent)
matches L2 except for a single epoch. L3 (GenUI) matches L2 except for batch
size $2$ and a single epoch.

\section{MoL Deployment Details}
\label{app:mol_deployment}

This appendix collects the detailed deployment measurements supporting
Section~\ref{sec:mol_deployment}. The rows come from several validation
snapshots across H20 and B300 systems, vLLM and SGLang backends, and different
parallelism layouts. We retain only measurements with a clearly specified
operating point and use each table to support one claim: weight residency,
capacity, latency scaling, or correctness. Cross-engine snapshots with
different request loads, LoRA residency, or speculative-decoding settings are
not reported as quantitative comparisons. No row should be interpreted as a
paired latency or throughput comparison against separately deployed merged
specialists.

\begin{table}[H]
\centering
\caption{H20 long-context capacity observations. KV usage is the peak fraction reported by the engine for the stated workload; the rows characterize feasible workloads rather than a throughput comparison.}
\label{tab:mol_layout_capacity}
\scriptsize
\begin{tabular}{lllll}
\toprule
Platform & Layout & Workload & Peak KV & Outcome \\
\midrule
H20 & TP8 & 8 $\times$ 131K, 256 out & 89.0\% & 8/8 clean \\
H20 & TP4/PP2/DCP4 & 16 $\times$ 56K, 128 out & 90.9\% & 16/16 clean \\
H20 & TP4/PP2/DCP4 & 8 $\times$ 180K, 128 out & 90.1\% & 8/8 clean \\
H20 & TP4/PP2/DCP4 & 4 $\times$ 230K, 128 out & 92.9\% & 4/4 clean \\
\bottomrule
\end{tabular}%
\end{table}

\begin{table}[H]
\centering
\caption{B300 TP8 DCP scaling under 1K input and 256 generated tokens. TTFT is the p50 in seconds and TPOT is the p50 inter-token latency in milliseconds. All rows use the same workload and engine configuration except for DCP degree and concurrency.}
\label{tab:mol_dcp_scaling}
\scriptsize
\begin{tabular}{lrrrrr}
\toprule
Layout & Concurrency & TTFT p50 (s) & TPOT p50 (ms) & TTFT p95 (s) & TPOT p95 (ms) \\
\midrule
TP8/DCP2 & 1  & 0.381 & 28.8 & 0.400 & 28.9 \\
TP8/DCP2 & 8  & 1.017 & 35.9 & 1.379 & 38.4 \\
TP8/DCP2 & 16 & 1.273 & 37.7 & 1.933 & 40.4 \\
TP8/DCP2 & 32 & 1.543 & 50.0 & 3.348 & 52.3 \\
TP8/DCP2 & 64 & 3.725 & 65.1 & 6.345 & 74.4 \\
\midrule
TP8/DCP4 & 1  & 0.385 & 29.3 & 0.401 & 29.4 \\
TP8/DCP4 & 8  & 1.109 & 36.6 & 1.424 & 39.4 \\
TP8/DCP4 & 16 & 1.336 & 38.6 & 2.094 & 41.7 \\
TP8/DCP4 & 32 & 1.743 & 51.6 & 3.742 & 54.3 \\
TP8/DCP4 & 64 & 4.130 & 68.0 & 7.117 & 78.7 \\
\midrule
TP8/DCP8 & 1  & 0.378 & 30.8 & 0.403 & 30.9 \\
TP8/DCP8 & 8  & 1.304 & 38.2 & 1.610 & 41.7 \\
TP8/DCP8 & 16 & 1.673 & 41.0 & 2.613 & 45.0 \\
TP8/DCP8 & 32 & 2.078 & 55.3 & 4.614 & 58.7 \\
TP8/DCP8 & 64 & 5.206 & 74.1 & 8.852 & 88.0 \\
\bottomrule
\end{tabular}%
\end{table}

\begin{table}[H]
\centering
\caption{Validated EAGLE-enabled DCP8 operating points on B300 TP8 with 8K input, 1K output, and page size 64. EAGLE uses five speculative steps, top-$k$~1, and six draft tokens. TPOT is the mean and p99 over the corresponding stress run; output throughput is aggregate across concurrent requests.}
\label{tab:mol_eagle_dcp8}
\scriptsize
\begin{tabular}{lrrrrr}
\toprule
Concurrency & TPOT mean (ms) & TPOT p99 (ms) & TTFT (ms) & E2E mean (s) & Output (tok/s) \\
\midrule
1  & 8.6  & 14.3 & 554     & 9.3  & 110 \\
16 & 18.0 & 30.7 & 1{,}494 & 19.9 & 757 \\
\bottomrule
\end{tabular}
\end{table}

\begin{table}[H]
\centering
\caption{CP LayerSplit impact on cold needle-in-haystack TTFT at 900K context. B300 8$\times$L20D, DSA sparse attention, Mooncake RDMA PD transfer. LayerSplit assigns disjoint layer subsets per CP rank so the paged-indexer address domain is globally consistent.}
\label{tab:mol_cp_layersplit}
\scriptsize
\begin{tabular}{lrrr}
\toprule
Prefill config & Cold TTFT (s) & Speedup & KV capacity / rank \\
\midrule
CP1 (baseline)       & 107.1 & $1.0\times$ & 2.34\,M \\
CP8 + LayerSplit     & 49.2  & $2.2\times$ & 2.34\,M \\
\bottomrule
\end{tabular}
\end{table}

\begin{table}[H]
\centering
\caption{Logical KV capacity on B300 TP8 with and without EAGLE (FP8, page size 64). Capacity is the total pool across all TP ranks; EAGLE reduces the per-rank pool from approximately 2.00M to 1.17M tokens because the draft model and its cache remain resident.}
\label{tab:mol_dcp_capacity}
\scriptsize
\begin{tabular}{rrr}
\toprule
DCP & Without EAGLE (M tokens) & With EAGLE (M tokens) \\
\midrule
1 & 2.00 & 1.17 \\
2 & 4.00 & 2.34 \\
4 & 8.00 & 4.67 \\
8 & 16.0 & 9.34 \\
\bottomrule
\end{tabular}
\end{table}

\begin{table}[H]
\centering
\caption{Deployment boundaries established by correctness and capacity validation. Each row states a scoped empirical observation and the corresponding constraint on the validated MoL configuration.}
\label{tab:mol_deployment_invariants}
\scriptsize
\begin{tabular}{p{0.19\linewidth}p{0.42\linewidth}p{0.29\linewidth}}
\toprule
Boundary & Validation evidence & Implication \\
\midrule
Backend compatibility & FlashMLA sparse DCP required a version-specific vLLM overlay on B300, while the H20 and B300 launch profiles used distinct validated configurations. & Treat engine version, hardware, attention backend, and parallelism layout as one scoped operating point. \\
DCP organization & Replicated DCP paths passed the reported checks; sharded DCP produced systematic corruption and reasoning loops. & Use replicated DCP in the validated deployment envelope. \\
PD service state & Reconfiguring one endpoint without synchronizing its peers produced transfer corruption and unavailable responses. & Update the prefill, decode, routing, and Gateway state as one service generation. \\
Long-context admission & Engine-level long-context completion did not imply equivalent end-to-end capacity through all route--answer--summary hops. & Apply admission control to the complete MoL request path, not only to the answer worker. \\
Prefill scheduling & Small prefill budgets serialized long requests even when sufficient KV capacity remained available. & Interpret TTFT jointly with the scheduler budget and context distribution. \\
\bottomrule
\end{tabular}
\end{table}

\end{document}